\documentclass[format=acmsmall,screen=true,review=false]{acmart_draft}

\usepackage{csquotes}
\usepackage[ngerman,english]{babel}
\usepackage[utf8]{inputenc}
\usepackage{pgfplots}
\usepackage{caption}
\usepackage{adjustbox}
\usepackage[super]{nth}
\usepackage{mathtools} 
\usepackage{amssymb} 
\usepackage{fontawesome}
\usepackage{tcolorbox}
\usepackage{leftidx}
\usepackage{tensor}
\usepackage{trimclip}
\usepackage{enumitem}
\usepackage{xcolor}
\usepackage{tcolorbox}
\usepackage{pdflscape}
\usepackage{multirow,tabularx}
\usepackage{booktabs}
\usepackage{collcell}
\usepackage{epstopdf}

\makeatletter
\DeclareRobustCommand{\cev}[1]{%
	\mathpalette\do@cev{#1}%
}
\newcommand{\do@cev}[2]{%
	\fix@cev{#1}{+}%
	\reflectbox{$\m@th#1\vec{\reflectbox{$\fix@cev{#1}{-}\m@th#1#2\fix@cev{#1}{+}$}}$}%
	\fix@cev{#1}{-}%
}
\newcommand{\fix@cev}[2]{%
	\ifx#1\displaystyle
	\mkern#23mu
	\else
	\ifx#1\textstyle
	\mkern#23mu
	\else
	\ifx#1\scriptstyle
	\mkern#22mu
	\else
	\mkern#22mu
	\fi
	\fi
	\fi
}

\makeatother

\makeatletter
\DeclareRobustCommand{\shortto}{%
	\mathrel{\mathpalette\short@to\relax}%
}

\newcommand{\short@to}[2]{%
	\mkern2mu
	\clipbox{{.70\width} 0 0 0}{$\m@th#1\vphantom{+}{\rightarrow}$}%
}
\makeatother

\makeatletter
\DeclareRobustCommand{\shortgets}{%
	\mathrel{\mathpalette\short@gets\relax}%
}

\newcommand{\short@gets}[2]{%
	\mkern2mu
	\clipbox{0 0 {.60\width} 0}{$\m@th#1\vphantom{+}{\leftarrow}$}%
}
\makeatother

\newcounter{autorownum}

\DeclareMathOperator{\adjheter}{\mathit{adj}_{\mathit{heter}}}
\DeclareMathOperator{\adjinter}{\mathit{adj}_{\mathit{inter}}}
\DeclareMathOperator{\adjintra}{\mathit{adj}_{\mathit{intra}}}

\DeclareMathOperator{\adjXi}{\mathit{adj}_{\mathit{mode}}}
\DeclareMathOperator{\Cbbpv}{C_{\mathit{bbpv}}}
\DeclareMathOperator{\Cbr}{C_{\mathit{br}}}
\DeclareMathOperator{\classify}{\theta}

\DeclareMathOperator{\Cws}{C_{\mathit{ws}}}

\DeclareMathOperator{\CwsX}{C_{\mathit{ws}_\mathit{mode}}}
\DeclareMathOperator{\Czh}{C_{\mathit{zh}}}

\DeclareMathOperator{\Distance}{\delta}

\DeclareMathOperator{\ged}{\mathit{ged}}
\DeclareMathOperator{\GED}{\operatorname{GED}}
\DeclareMathOperator{\gep}{\mathit{gep}}
\DeclareMathOperator{\ges}{\operatorname{GES}}
\DeclareMathOperator{\identity}{id}
\DeclareMathOperator{\scale}{scale}
\DeclareMathOperator{\oclassify}{\vartheta}
\DeclareMathOperator{\rclassify}{\theta^{\shortgets}}
\DeclareMathOperator{\roclassify}{\vartheta^{\shortgets}}

\DeclareMathOperator{\veo}{\operatorname{VEO}}

\DeclareMathOperator{\wal}{\operatorname{WAL}} 
\DeclareMathOperator{\wges}{\mathit{wges}}
\DeclareMathOperator{\myauthors}{\mathit{authors}}
\DeclareMathOperator{\articles}{\mathit{texts}}

\newcommand*{\Scale}[2][4]{\scalebox{#1}{$#2$}}

\newcommand{\activity}{\text{\faPencil}} 
\newcommand{\Author}[1]{#1}
\newcommand{\averageactivity}{\overline{|\activity(\cdot,\cdot)|}}
\newcommand{\classification}{\mathcal{C}}

\newcommand{\myto}{.} 
\newcommand{\Prime}[2]{#1_{#2}}
\newcommand{\BarPrime}[2]{\overline{#1_{#2}}}
\newcommand{\Stack}[3]{#1\to_{#3}#2}

\newcommand{\StackX}[3]{#1\stackrel{\Scale[0.65]{#3}}{\to}#2}
\newcommand{\StackXY}[4]{#1\overset{\Scale[0.65]{#3}}{\underset{\Scale[0.65]{#4}}{\leftrightarrow}}#2}

\newcommand{\topicnetwork}{T}
\newcommand{\ToSi}{\textit{ToSi}}
\newcommand{\Word}[1]{#1}
\newcommand{\XStackY}[4]{#1\,\leftidx{_{#4}}\!\!\leftrightarrow_{#3}#2}

\usepackage{xspace}
\newcommand{\Cities}{\textsc{Cities}\xspace}
\newcommand{\Regions}{\textsc{Regions}\xspace}
\newcommand{\Others}{\textsc{Others}\xspace}
\newcommand{\WPRegioOne}{\textsc{WP-Re\-gio-1}\xspace}
\newcommand{\WPRegioTwo}{\textsc{WP-Re\-gio-2}\xspace}
\newcommand{\WPOthersOne}{\textsc{WP-O\-the\-rs-1}\xspace}
\newcommand{\WPOthersTwo}{\textsc{WP-O\-th\-ers-2}\xspace}

\definecolor{SeminarBlau}{RGB}{0,154,224}
\definecolor{SeminarRot}{rgb}{0.75,0,0}
\definecolor{SeminarGruen}{RGB}{0,128,0}
\definecolor{SeminarOrange}{RGB}{237,167,45} 
\definecolor{SeminarGrau}{rgb}{0.32,0.3,0.38} 
\colorlet{SeminarHellBlau}{SeminarBlau!25!white}
\colorlet{SeminarHellGruen}{SeminarGruen!25!white}
\colorlet{SeminarHellRot}{SeminarRot!25!white}
\colorlet{SeminarHellGrau}{SeminarGrau!35!white}
\colorlet{SeminarSehrHellRot}{SeminarRot!15!white}
\colorlet{SeminarSehrHellGrau}{SeminarGrau!15!white}
\colorlet{SeminarSehrHellGruen}{SeminarGruen!15!white}
\colorlet{SeminarSehrHellBlau}{SeminarBlau!15!white}
\colorlet{SeminarSehrSehrHellGrau}{SeminarGrau!5!white}

\usepackage{tikz} 
\usetikzlibrary{matrix}

\theoremstyle{definition}
\newtheorem{exmp}{Example}[section]
\newtheorem{hypothesis}{Hypothesis}

\usepackage[autolanguage]{numprint}

\begin{document}

\title{From Topic Networks to Distributed Cognitive Maps}
\subtitle{Zipfian Topic Universes in the Area of Volunteered Geographic Information} 

\author{Alexander Mehler}
\affiliation{%
 \institution{Goethe-University Frankfurt}
 \streetaddress{}
 \city{Frankfurt}
}
\email{mehler@em.uni-frankfurt.de}

\author{Rüdiger Gleim}
\affiliation{%
	\institution{Goethe-University Frankfurt}
	\streetaddress{}
	\city{Frankfurt}
}
\email{gleim@em.uni-frankfurt.de}

\author{Regina Gaitsch}
\affiliation{%
	\institution{Network Caring Economy}
	\streetaddress{}
	\city{Bad Nauheim}
}
\email{r_gaitsch@yahoo.de}

\author{Wahed Hemati}
\affiliation{%
	\institution{Goethe-University Frankfurt}
	\streetaddress{}
	\city{Frankfurt}
}
\email{hemati@em.uni-frankfurt.de}

\author{Tolga Uslu}
\affiliation{%
	\institution{Goethe-University Frankfurt}
	\streetaddress{}
	\city{Frankfurt}
}
\email{uslu@em.uni-frankfurt.de}

\renewcommand{\shortauthors}{}

\begin{abstract}
Are nearby places (e.g.\ cities) described by related words?
In this article we transfer this research question in the field of lexical encoding of geographic information onto the level of intertextuality.
To this end, we explore \textit{Volunteered Geographic Information} (VGI) to model texts addressing places at the level of cities or regions with the help of so-called topic networks.
This is done to examine how language encodes and networks geographic information on the aboutness level of texts.
Our hypothesis is that the networked thematizations of places are similar -- regardless of their distances and the underlying communities of authors.
To investigate this we introduce \textit{Multiplex Topic Network}s (MTN), which we automatically derive from \textit{Linguistic Multilayer Network}s (LMN) as a novel model, especially of thematic networking in text corpora.
Our study shows a Zipfian organization of the thematic universe in which geographical places (especially cities) are located in online communication.
We interpret this finding in the context of \textit{cognitive maps}, a notion which we extend by so-called \textit{thematic maps}.
According to our interpretation of this finding, the organization of thematic maps as part of cognitive maps results from a tendency of authors to generate shareable content that ensures the continued existence of the underlying media.
We test our hypothesis by example of special wikis and extracts of Wikipedia.
In this way we come to the conclusion:
Places, whether close to each other or not, are located in neighboring places that span similar subnetworks in the topic universe.
\\
\textbf{Keywords:} Volunteered Geographic Information, Cognitive Maps, Multiplex Topic Networks, Linguistic Multilayer Networks
\end{abstract}

\maketitle

\section{Introduction}\label{sec:Introduction}

In this article, we explore crowd-sourced resources for automatically characterizing geographical places with the help of so-called \textit{topic networks}.
Our goal is to model the thematic structure of corpora of natural language texts that are about certain places seen as thematic frames.
This is done in order to automatically compare the thematic structures of corpora of texts about these places, which will be represented as topic networks.
In this way we want to investigate the regularity or systematicity according to which geographical objects (i.e.\ cities and regions) are dealt with, especially in online communication.

Our work relates to what is described by Crooks et al.\ \cite{Crooks:Pfoser:Jenkins:Croitoru:Stefanidis:Smith:Karagiorgou:Efentakis:Lamprianidis:2015} as a novel paradigm of modeling \enquote{\textit{urban morphologies}}.
We not only add special wikis such as regional and city wikis as candidates to the resources listed in \cite{Crooks:Pfoser:Jenkins:Croitoru:Stefanidis:Smith:Karagiorgou:Efentakis:Lamprianidis:2015}. 
Rather, we also introduce a novel method for modeling their content. 
This concerns local media of collaborative writing about places \cite[cf.][]{Crang:Graham:2007} which contain \textit{everyday place descriptions} \cite{Chen:Vasardani:Winter:Tomko:2018} authored and networked according to the wiki principle.
The corresponding wikis and the subgraphs of Wikipedia that we additionally analyze manifest \textit{Volunteered Geographic Information} (VGI) \cite{Goodchild:2007,Goodchild:Li:2012,Hardy:Frew:Goodchild:2012} and thus relate to what is called the wikification of \textit{Geographical Information Systems} (GIS) \cite{Sui:2008}.
VGI is \enquote{\textit{completing traditional authoritative geographic information}} \cite{Jiang:Thill:2015}, an information source which is still \enquote{\textit{underutilized}} in geography \cite{Salvini:Fabrikant:2016} as a source of big textual data \cite{Jiang:Thill:2015} making natural language processing an indispensable prerequisite for its analysis. 
According to Hardy et al.\ \cite{Hardy:Frew:Goodchild:2012} authoring VGI has a \textit{spatial} component in the sense that people likely write about \textit{local} content though this also holds for Wikipedia for a minor degree \cite{Hecht:Gergle:2010:b}.
This spatial component can be accompanied by a lack of quality assurance, which makes VGI susceptible to deficiencies and to a distorted resource of still unknown extent \cite{Goodchild:Li:2012}.
In any event, the biased coverage of VGI is a characteristic of resources like Wikipedia so that the same region can be displayed very differently in its various language editions \cite{Graham:Hogan:Straumann:Medhat:2014}, a sort of biasing which is typical for user generated content.
Nevertheless, Hahmann \& Burghardt \cite{Hahmann:Burghardt:2013} show that more than 50\% of the articles in the German Wikipedia contain geo-referenced data (at least indirectly via links to other articles), so that such media can be regarded as rich resources of VGI.
Moreover, Goodchild \& Li \cite{Goodchild:Li:2012} point to the fact that crowd-sourcing or, more precisely, crowd-curation \cite{Jenkins:Croitoru:Crooks:Stefanidis:2016}, as enabled by wikis, is a means of quality assurance.

We follow this concept and assume that geographic data, as manifested linguistically in online media, are a valuable resource to investigate how communities form a common sense for addressing places of common interest.
In line with Davies \cite[][41]{Davies:2009} we additionally assume that \enquote{\textit{[a]s people communicate more about a place, social consensus will create increased similarity between and within people's judgments of it}.}
However, we also assume that the latter similarity can affect communications of different communities about different places.
In this way, we assume a kind of horizontal self-similarity \cite{Mehler:Gleim:Luecking:Uslu:Stegbauer:2018} of the thematic structure of online media, which is more or less independent of the underlying theme and the community.
That is, our hypothesis on the theming of places is as follows:
\begin{hypothesis}\label{hypo:Hypothesis 1}
Thematizations of different places at a certain level of thematic abstraction tend to be similar among each other (rather than being dissimilar) in the sense that 
(1) they focus on similar topics,
(2) the way these topics are networked and
(3) with respect to the skewness of this focus, 
regardless of whether the underlying media are generated by different communities and whether these communities address related or unrelated places at near or distant spaces.
\end{hypothesis}
\begin{figure}[t]
	\includegraphics[width=1.00\textwidth]{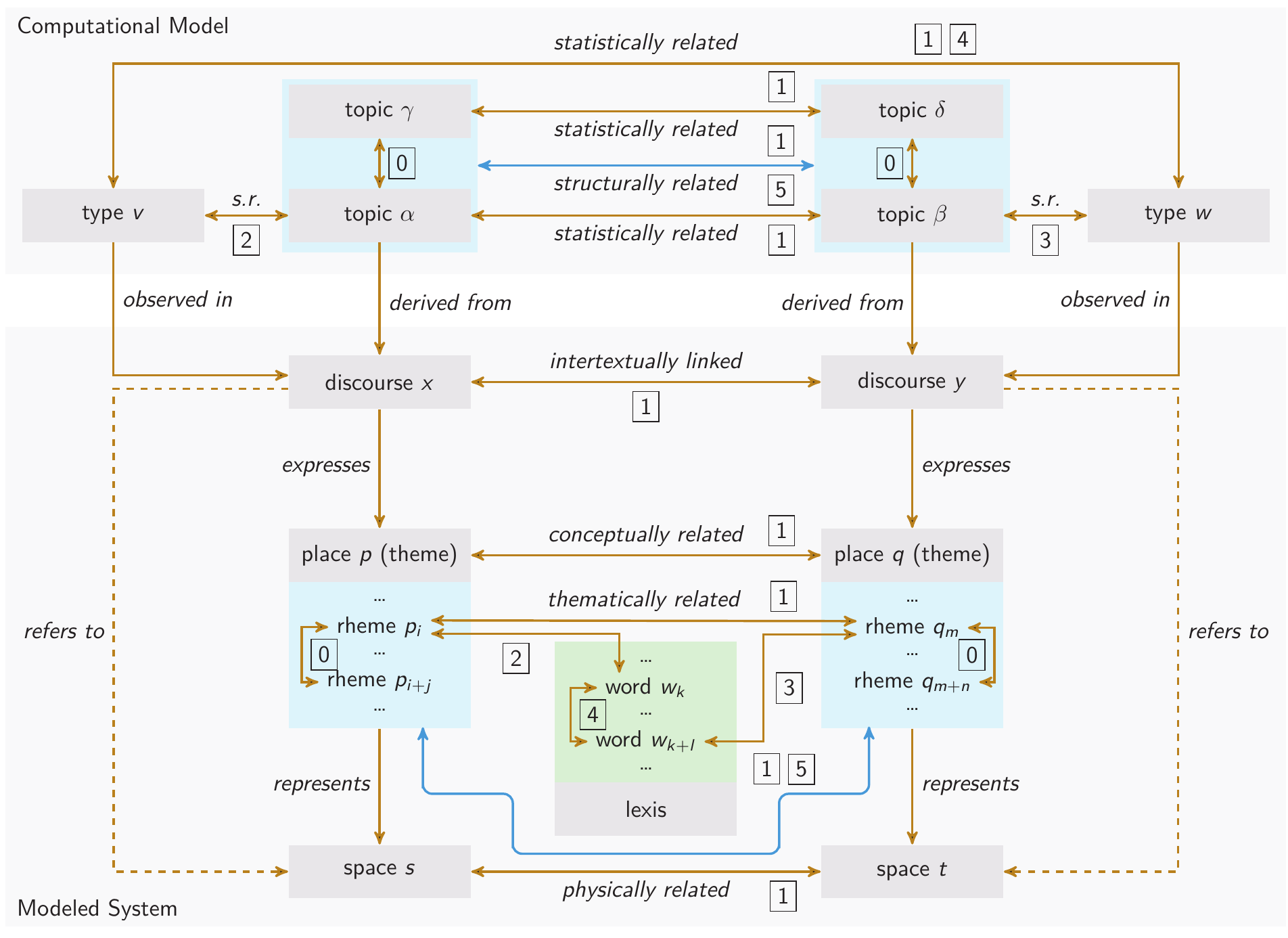}	
	\caption{Schematic depiction of a generalization of a hypothesis of Louwerse \& Zwaan \cite{Louwerse:Zwaan:2009} saying that language encodes geographical information: 
	the places $p,q$ are expressed in the discourses $x,y$, from which the topic representations $\alpha,\beta$ are computationally derived.
	Places are structured into systems of networked rhemes or subtopics. 
	The conceptual relatedness of $p$ and $q$ is grounded in the relatedness of the rhemes ${p_i}$ and ${q_m}$ and modeled by the relatedness of the derived topics $\alpha$ and $\beta$ modeling these rhemes.
	According to the semiotic triangle, we assume that the relation of signs (here: texts) to their referents (here: spaces) is mediated by sign processes. 
	We use dashed arcs to express the indirect relation of the former to the latter.
	In lexical variants of this approach, $p$ and $q$ are preferably denoted or described by some words $w_k, w_{k+l}$ of the underlying lexis, which are syntagmatically or paradigmatically associated and modeled by some types $v,w$.
	Framed numbers indicate relations that potentially parallelize each other. \textit{s.r.} means \textit{statistically related}.
	}
	\label{fig:TFL}
\end{figure}

The intuition behind Hypothesis 1 is that thematizations of places in web-based communication are seemingly somehow thematically redundant: 
In reporting, for example, on the cities in which people live, they may aim to emphasize the special character of these places. 
It seems, however, as if a thematic trend is breaking ground that ultimately makes such reports appear thematically very similar. 
Whether or not this intuition is actually a trend that can be observed specifically in the field of wiki-based media is something this study is intended to clarify.
From this point of view, it is obvious that Hypothesis 1 is only a starting point which in itself needs further clarification in order to be testable:
similarity, for example, is a highly context sensitive attribute \cite{Medin:Goldstone:Gentner:1993} that needs further definitional specifications in order to be computable.
Likewise the concept of thematization (theme or topic) -- a concept which according to \cite{Adamzik:2016} has so far found comparatively less attention in linguistics -- is not yet specified in Hypothesis 1.
Thus, an appropriate elaboration and concretization of Hypothesis 1 is one of the main tasks of the present paper.
To this end, it is developing a \textit{generic topic network model} in conjunction with a measurement procedure which will specify both the notion of \textit{similarity} (which will be defined in terms of the graph similarity of topic networks) and of the \textit{thematization of places} (which will be defined in terms of topic labeling and topic networking).
This topic network model will allow Hypothesis 1 to be reformulated and concretized in the form of variants (i.e., hypotheses 2, 3 and 4), which will be presented in the third part of the paper (in Section \ref{sec:Module 8 -- Machine Learning -- and Module 9 -- Classification Analysis}) and whose formulations presuppose the topic network model that this paper develops in the preceding sections.

The skewness that is mentioned by Hypothesis 1 reminds one of a Zipfian process, according to which a few topics dominate, while the majority of candidate topics is underrepresented or disregarded.
Therefore, we speak of \textit{Zipfian thematic universes}, which are spanned by the thematization of the same places in online media such as special wikis of the sort studied here.
By the term \textit{topic} we refer to the notion of aboutness of texts \cite{Adamzik:2016,Yablo:2014}.
From a linguistic point of view, the terminology of Hypothesis \ref{hypo:Hypothesis 1} seems to be confusing when referring to places as \textit{what is given} and with topic to \textit{what is said about these places}.
The reason is that linguistics distinguishes between what is given (\textit{theme} or \textit{topic}) and what is said about it (\textit{rheme}, \textit{comment} or \textit{focus}) in a given piece of text \cite{Adamzik:2016,Brinker:1992,Danes:1995,Hoffmann:2000:a}:
a mention of a city like \textit{Vienna}, for example, can be connected with certain subtopics (e.g.\ \textit{classical music}), which characterize this place rhematically by providing new information about it.
The latter distinction is meant when we relate subtopics in the role of rhemes to places in the role of topics in the linguistic sense.
Thus, when talking about topics as part of a computational model, we will use the term \textit{topic} (topic$_2$), while when talking about places as topics in the linguistic sense (topic$_1$), we will use the term \textit{theme} and speak about its rhemes as its subtopics modeled by topics (topic$_2$) as units of our model.
This scenario and its relation to Hypothesis \ref{hypo:Hypothesis 1} is depicted in Figure \ref{fig:TFL}.
It shows a generalization of a hypothesis of Louwerse \& Zwaan \cite{Louwerse:Zwaan:2009} according to which language encodes geographical information: 
the places $p$ and $q$, which are understood as conceptual units (i.e.\ mental models), are described by or expressed in two discourse units (texts, dialogs etc.) $x$ and $y$.
From the latter units, the topic representations $\alpha$ and $\beta$ are derived by means of a computational model (e.g., \textit{Latent Dirichlet Allocation} (LDA) \cite{Blei:Ng:Jordan:2003} or the topic network model introduced in Section \ref{sec:A Topic Network Model}).
While such derived topics are part of the computational model, the underlying discourses belong to the modeled system.
We assume that the conceptual unit $p$ ($q$) is structured into a system of networked rhemes or subtopics ${p_i}$ (${q_m}$). 
Ideally, the derived topic $\alpha$ in Figure \ref{fig:TFL} is a valid model of one of the rhemes of place $p$ (e.g.\ ${p_i}$) and $\beta$ of one of the rhemes of place $q$ (e.g.\ ${q_m}$).
If we assume now that $p$ and $q$ are conceptually related (e.g.\ similar) to each other, then the linguistic encoding hypothesis implies that this is possibly reflected by a relatedness (e.g.\ similarity) relation among some rhemes of these places (e.g.\ by the relatedness of ${p_i}$ and ${q_m}$).
From the point of view of modeling, this relation is ideally mapped by the relatedness (e.g.\ similarity) of the derived topics $\alpha$ and $\beta$.
We assume that conceptual relations between places can be parallelized by relations of physical proximity or distance between spaces that are mentally modeled by these places.
If one additionally assumes that proximity in space correlates with relatedness in conceptual space (the less distant, the more similar, for example), one obtains a linguistic variant of Tobler's so-called first law (see Section \ref{sec:Related Work}).
If we look at the literature (see Section \ref{sec:Related Work}), we find that the approaches in this area differ in terms of the linguistic level at which they observe the linguistic encoding of platial \cite{Jenkins:Croitoru:Crooks:Stefanidis:2016} relations: 
for example, at the level of intertextually linked texts, at the level of the topics these texts are about, or at the level of lexical elements used by these and other texts to deal with the latter topics.
In lexical variants of this approach, the places $p$ and $q$, for which we assume that they are conceptually related, are preferably referred to or described by means of lexical items $w_k, w_{k+l}$ (see Figure \ref{fig:TFL}) of the underlying lexis that are syntagmatically or paradigmatically associated.
From the point of view of modeling, we have then to assume two types $v,w$ (as models of the words $w_k, w_{k+l}$) for which we automatically detect, for example, their (paradigmatic) closeness in semantic space \cite[cf.][]{Rieger:2003,Davies:2013} or the similarity of their (syntagmatic) co-occurrence statistics \cite[cf.][]{Louwerse:Benesh:2012}.

From this analysis we obtain a series of reference points or means for encoding geographical information about conceptual relations (see [1] in Figure \ref{fig:TFL}) of places. 
This concerns more precisely a series of possible parallelizations of such relations, which may ultimately be parallelized by relations between the spaces designated by these places (for the numbers in brackets see Figure \ref{fig:TFL}):
at the level of the modeled system, this refers to thematically linked rhemes, intertextually linked discourse units (e.g.\ texts) and to syntagmatically or paradigmatically linked words ([1]).
From a modeling point of view, we distinguish the statistical relatedness of types or of topics as candidate parallelizations ([1]).
Beyond that we find the parallelization of the relatedness of rhemes and words on the one hand and of types and topics on the other ([2], [3]) as well as that of the relatedness of words on the one hand and of types on the other ([4]).
The parallelization of the relatedness of rhemes of the same place ([0]) by the relatedness of the rhemes of another place concerns the core of our network approach.
Such relations among rhemes constitute rhematic networks or networks of rhemes on both sides of the affected places.
Our main assumption is now that any such rhematic network, which manifests the thematic structure of a place, can be related \textit{as a whole} to that of another place.
In doing so it is, from a modeling point of view, ideally parallelized \textit{by the structural relatedness (e.g.\ similarity or complementarity) of topic networks}, which are derived from corpora of texts, each of which describes one of these places ([5]).
This type of parallelization affects entire networks of linguistic objects, and yet offers a means of encoding the conceptual relationship of places ([1]) or the proximity of spaces, respectively.
In the present paper we explore relations of Type [5] in order to learn about the encoding of geographical information in natural language texts, that is, about relations of Type [1].
To this end, we develop, instantiate and empirically test a formal model of multiplex topic networks derived from so-called linguistic multilayer networks as a model of relations of Type [5].

From this point of view, Hypothesis \ref{hypo:Hypothesis 1} means that certain rhemes of places and the structure they span resemble each other, regardless of how far the quantified distances of the spaces represented by these places are and regardless of the fact that the texts in which these rhemes are described are written by different communities.
To test this hypothesis, we introduce topic networks to make the networking of topics a research object according to the scenario described in Figure \ref{fig:TFL}, that is, in relation to the hypothesis of linguistic encoding of geographical information. 
The contributions of this article are of theoretical, methodical and empirical nature: 
\begin{enumerate}
\item \textit{Formal modeling:} 
We develop a generic, extensible formalism for the representation of topic networks that covers a wide range of informational sources for spanning and weighting topic links.
To this end, we introduce the notion of \textit{multiplex topic networks} derived from so-called \textit{multilayer linguistic networks}.
In this way we enable the same place to be represented by a family of thematic networks that offer different perspectives on the networking of its rhemes.
We exemplify this model by means of two perspectives provided by so-called \textit{Text Topic Networks} (TTN) and their corresponding \textit{Author Topic Networks} (ATN).
\item \textit{Procedural modeling:} 
we develop a measurement procedure for instantiating our formal model. 
To this end, we introduce novel measures of the similarity of labeled graphs that are sensitive to their links \textit{and} to their nodes.
\item \textit{Experimentation:} 
We further develop the range of baseline statistics in network theory in order to better assess the quality of our measurements. 
To this end, we test our model by means of a threefold classification experiment that compares a set of TTNs with each other, a set of corresponding ATNs with each other and the former TTNs with the latter ATNs.
\item \textit{Theory formation:} 
We interpret our findings in the context of cognitive maps, thus building a bridge between our network-theoretical approach and approaches to the cognitive representation of geographical information.
We show how to integrate the analysis of entire networks into the research about the linguistic encoding of geographical information (see Figure \ref{fig:TFL}).
\end{enumerate}

The paper is organized as follows:
Section \ref{sec:Related Work} discusses related work.
Section \ref{sec:A Topic Network Model} introduces our formal model of linguistic multilayer networks and of the multiplex topic networks derived from them.
Section \ref{sec:Experimentation} describes our experiments in detail and Section \ref{sec:Discussion} discusses our findings.
Finally, Section \ref{sec:Conclusion} concludes and gives an outlook on future work.

\section{Related Work}\label{sec:Related Work}

Our work is related to linguistic research on Tobler's \cite{Tobler:1970} first law (TFL) which says that \enquote{[\ldots] \textit{everything is related to everything else, but near things are more related than distant things}.} \cite[][236]{Tobler:1970}.
Due to its underspecification, this so-called law raised many questions about what it means to be \textit{related} or \textit{distant} \cite{Miller:2004}.
Accordingly, a range of approaches exist that make different proposals to interpret relatedness also in terms of \textit{semantic relatedness}.
In the context of information visualization, Montello et al.\ \cite{Montello:Fabrikant:Ruocco:Middleton:2003} test a variant of TFL called the first law of \textit{cognitive} geography which says that \enquote{\textit{people believe closer things to be more similar than distant things}} \cite[][317]{Montello:Fabrikant:Ruocco:Middleton:2003} where spatial distance is referred to for judging the similarity of information objects.
This approach is contrasted with a study by Hecht \& Moxley \cite{Hecht:Moxley:2009} who model relations of Wikipedia articles as a function of the probability of being linked in the web graph and find that this probability is related to the geographical distance of toponyms described in the articles.
Hecht \& Moxley relate their finding to the transitivity of networks by stating that the smaller the geographical distance of nodes, the higher their clustering coefficient \cite[][101]{Hecht:Moxley:2009}.
This work is extended by Li et al.\ \cite{Li:Sen:Hecht:2014}, who calculate semantic relationships of articles instead of hyperlinks and show that TFL holds independently of the geographical domain up to a certain distance threshold.
A lexical variant of TFL is mentioned by Yang et al.\ \cite{Yang:Chen:Lyu:King:2011}, according to which geographically close words tend to be clustered into the same geographical topics.
This phenomenon has earlier been studied by Louwerse et al.\ \cite[cf.\ the review in][]{Louwerse:Benesh:2012} who reformulate Firth's famous dictum by saying that \enquote{[\ldots] \textit{you shall know the physical distance between locations by the lexical company they keep.}}\ \cite[][1557]{Louwerse:Benesh:2012}.
This means that the distance of places correlates with syntagmatic associations between the lexical items used to describe them.
That is, language encodes geographical information \cite{Louwerse:Zwaan:2009} at least regarding the distances of semantically related places. 
From this perspective, TFL appears to be reformulated as a candidate for a geolinguistic law that is compatible with the more general 
\textit{Symbol Interdependency Hypothesis} (SIH) \cite{Louwerse:2011}. 
According to SIH, linguistic information encodes perceptual information so that the former serves as a shortcut to the latter \cite{Louwerse:2011}.
Finally, a rather text-linguistic variant of TFL is proposed by Adams \& McKenzie \cite{Adams:McKenzie:2013}, which states that near places are each described by texts whose topics are more similar than in the case of texts about distant places.

In contrast to these approaches, we hypothesize that places, no matter how far apart, have similar topic distributions when their descriptions are transmitted by media such as city and region wikis.
If we find evidence for this hypothesis, there are various candidates for explaining it:
firstly, such a finding could indicate a trivial meaning of TFL \cite[cf.][]{Miller:2004} in relation to the topics modeled by us, implying that everything, distant or not, is highly related.
Secondly, it could indicate the (in-)effectiveness of distances and similarities at different scales: 
at the level of local, specific topics (within the scope of TFL) and at the level of global, more general topics (outside the scope of TFL).
Thirdly, such a finding could indicate a hidden similarity of processes of collaboratively writing wikis about different places, even if the wikis are written by different communities (see Hypothesis \ref{hypo:Hypothesis 1}).
In order to decide between these alternatives, we need a new topic model that derives networks of thematic structures at different scales from texts in online media about the same places.
This should at least include the networking of topics along relations of intertextuality and co-authorship in order to allow for revealing similarities of the underlying processes of collaborative writing.
To this end, we will develop multiplex networks that integrate text- and author-driven topic networks.

So far, most approaches to thematic aspects of places use topic modeling based on \textit{Latent Dirichlet Allocation} (LDA) to associate topics and texts about geographical units, where topics are represented as sets of thematically related words.
An early approach in this regard is described by Mei et al.\ \cite{Mei:et:al:2006} who model \textit{spatio-temporal theme patterns} to identify dominant topics in texts that are connected to places. 
A related approach is proposed by Hao et al.\ \cite{Hao:et:al:2010}, who aim to detect topics that are \enquote{localized} in places. 
This is done to ground their similarities in relations of their thematic representations -- a scenario that is omnipresent in linguistically motivated work in the context of TFL (cf.\ Figure \ref{fig:TFL}).
Likewise, Adams \& McKenzie \cite{Adams:McKenzie:2013} extract topic models from travel blogs to detect topics as groups of semantically related words associated to places, so that relations among places can be identified by shared topics. 
Another example is proposed by Bahrehdar \& Purves \cite{Bahrehdar:Purves:2018}: 
instead of documents written by individual authors, they analyze tagging data extracted from image descriptions in Flickr.
A hybrid model of topic modeling comes from Yin et al.\ \cite{Yin:Cao:Han:Zhai:Huang:2011}, in which representations of regions are used instead of documents to link topics to places.
A related \textit{region-topic model} that uses regions as topics to map words, sentences and texts to distributions of regions or to ground them semantically \cite[cf.][]{Recchia:Louwerse:2014}, is proposed by Speriosu et al.\ \cite{Speriosu:Brown:Moon:Baldridge:Erk:2010}. 
A promising extension is developed by Gao et al.\ \cite{Gao:Janowicz:Couclelis:2017} who aim at detecting higher-level functional regions as semantically coherent areas of interest. 
To this end, they analyze co-occurrence relations between topics to describe many-to-may relations of locations and urban functions.
Another direction is pursued by Lansley \& Longley \cite{Lansley:Longley:2016}, who investigate the location- and time-based distribution of topics in Twitter, setting a number of twenty topics as a target for LDA.
See also Jenkins et al.\ \cite{Jenkins:Croitoru:Crooks:Stefanidis:2016} who utilize a list of six high-level topic categories. 
One of the largest studies in this context is the one of Gao et al.\ \cite{Gao:Janowicz:Montello:Hu:Yang:McKenzie:Ju:Gong:Adams:Yan:2017} who present an integrative approach to modeling texts from a range of different media such as Wikipedia, Twitter, Flickr etc.\ to demarcate cognitive regions \cite{Montello:2003}.
All these approaches start from topic modeling to map natural language texts onto distributions of topics in order to relate the places thematized by these texts (cf.\ Figure \ref{fig:TFL}).

A prominent precursor of topic models \cite{Leopold:2007} is given by \textit{Latent Semantic Analysis} (LSA) \cite{Landauer:Dumais:1997}. 
Consequently, there are studies in the context of TFL based on this predecessor.
Davies \cite{Davies:2013}, for example, interprets the associations of place names computed by LSA from place descriptions as a model of the cognitive representation of the corresponding spaces \cite[cf.][]{Davies:Tenbrink:2018}. 
This approach opens up a perspective for measuring biased cognitive representations of spatial systems: 
according to Davies, her approach provides representations of cognitive geographies that are explored by the associations of semantically close place names in accordance or not with the underlying geographical relations, that is, in accordance or not with TFL \cite[cf.][]{Recchia:Louwerse:2014}.
These and related studies produce interesting results about the localization of topics or vice versa about the thematization of places in texts.
However, they mostly disregard topic networking, not to mention the networking of topics viewed from different angles.
Although it is easy to derive a network approach from binary relations of topic similarity, relationships that cannot be traced back to sharing similar words are hardly mapped by topic models of the sort considered so far.
By generating topic distributions per location, for example, we know nothing about the dynamics of the co-authorship of the underlying texts:
in the extreme case one observes (dis-)similarities, which result from the activity of a small number of authors or even only one author -- in contrast to the assumed collaboration density of online media such as Wikipedia.
Therefore, it is our goal to develop a model of topic networks that simultaneously addresses the dynamics of the co-authorship of the underlying texts.
A subtask will be to develop a formal model of thematic networking that is generic enough to integrate a wide range of sources of networking -- at least theoretically.

While most of the approaches considered so far ignore aspects of networking, a second branch of research tends to follow the paradigm of network theory.
Hu et al.\ \cite{Hu:Ye:Shaw:2017}, for example, measure the semantic relatedness of cities as nodes of a city network \cite{Salvini:Fabrikant:2016} depending on the co-occurrences of city names in news articles.
This approach is related to Liu et al.\ \cite{Liu:Wang:Kang:Gao:Lu:2014}, who explore co-occurrences of toponyms to induce city networks that can be used to test predictions associated with TFL.
Hu et al.\ \cite{Hu:Ye:Shaw:2017} further develop this approach to networking cities by reference to topics of articles in which the corresponding toponyms are observed.
They use {Labeled LDA} \cite{Ramage:Hall:Nallapati:Manning:2009} to learn to extract topics $\alpha$ from texts to finally determine the $\alpha$-relative similarity of cities based on the co-occurrences of their names in texts about $\alpha$. 
Another approach to city networks using Wikipedia as a data source, is proposed by Salvini \& Fabrikant \cite{Salvini:Fabrikant:2016}:
they link cities as a function of the number of articles \enquote{co-siting} \cite{Bjoerneborn:2004} their Wikipedia articles.
A comprehensive perspective on modeling spatial information is developed by Luo et al.\ \cite{Luo:Wang:Liu:Gao:2019}, who propose a three-part network model that integrates representations of spatial, social, and semantic networks. 
In this conceptual model, semantics plays the role of interpreting behavior in spatial and social space and thus of bridging them.
Although we share this hybridization of the network perspective on spatial information, we strive for a more concrete model that can be empirically tested.

Any such study has to face various aspects of the vagueness \cite{Montello:2003,Agarwal:2005} or informational uncertainty \cite{Goodchild:Li:2012} of concepts of regions \cite{Montello:2003} and places \cite{Jenkins:Croitoru:Crooks:Stefanidis:2016} and especially of the names of such entities \cite{Gao:Janowicz:Montello:Hu:Yang:McKenzie:Ju:Gong:Adams:Yan:2017}.
According to Winter \& Freksa \cite{Winter:Freksa:2012} this includes \textit{semantic ambiguity}, \textit{indeterminacy of spatial extent} or \textit{boundary vagueness} \cite{Gao:Janowicz:Montello:Hu:Yang:McKenzie:Ju:Gong:Adams:Yan:2017}, preference-oriented \textit{re-scaling of extent} and the \textit{dynamics of salience} affected by various dimensions of contrast.
Beyond boundary vagueness, Gao et al.\ \cite{Gao:Janowicz:Montello:Hu:Yang:McKenzie:Ju:Gong:Adams:Yan:2017} speak of the shape and location vagueness by example of cognitive regions.
Furthermore, Jenkins et al.\ \cite{Jenkins:Croitoru:Crooks:Stefanidis:2016} refer to the temporal dynamics of places as evolving concepts as a source of uncertainty.
From a methodological point of view, this multi-faceted uncertainty has two implications: 
in relation to the model, which should be flexible enough to map these facets, and in relation to the object itself, which could complicate its modeling by unsystematically distorting it.

In accordance with Hu \cite{Hu:2017} we assume that the thematic perspective complements the spatial and temporal perspective of the study of places.
A rheme can be understood as the \enquote{content} of a geographical region that expands its dimensionality \cite{Montello:2003}.
This content may be further specified in terms of affordances, functions or shared conceptual representations associated by members of a community with the corresponding place so that different places can be related by being associated with similar content.
This thematic perspective will be at the core of our article.
To this end, we follow the approach of Jenkins et al.\ \cite{Jenkins:Croitoru:Crooks:Stefanidis:2016}, according to which places are connected with meanings generated by collaborators of crowd-sourcing media such as Wikipedia: 
their collaboration creates what Jenkins et al.\ call \textit{platial themes}, namely themes that are characteristic for certain places.
As shared meanings, these platial themes ultimately create a \enquote{\textit{collective sense of place}}, as it is perceived by the corresponding community.
In this context, Jenkins et al.\ \cite{Jenkins:Croitoru:Crooks:Stefanidis:2016} propose to study \textit{politics}, \textit{business}, \textit{education}, \textit{recreation}, \textit{sports}, and \textit{entertainment} as six high-level topics of places.
However, by reference to the \textit{Dewey Decimal Classification} (DDC) we will instead deal with more than six hundred hierarchically organized topics, each of which is manifested by a range of Wikipedia articles. 
In any event, we have to consider that thematic aspects may distort the conceptualization and perception of spatial objects \cite{Gao:Janowicz:Montello:Hu:Yang:McKenzie:Ju:Gong:Adams:Yan:2017}.
A central question then concerns the regularity or systematicity of this distortion in the sense of asking to what extent thematic representations of different places show similar aspects of being biased.
This question will be at the core of this article.

\section{Multiplex Topic Networks: a Novel Approach to Topic Modeling}\label{sec:A Topic Network Model}

\begin{figure}[t]
	\includegraphics[width=0.75\textwidth]{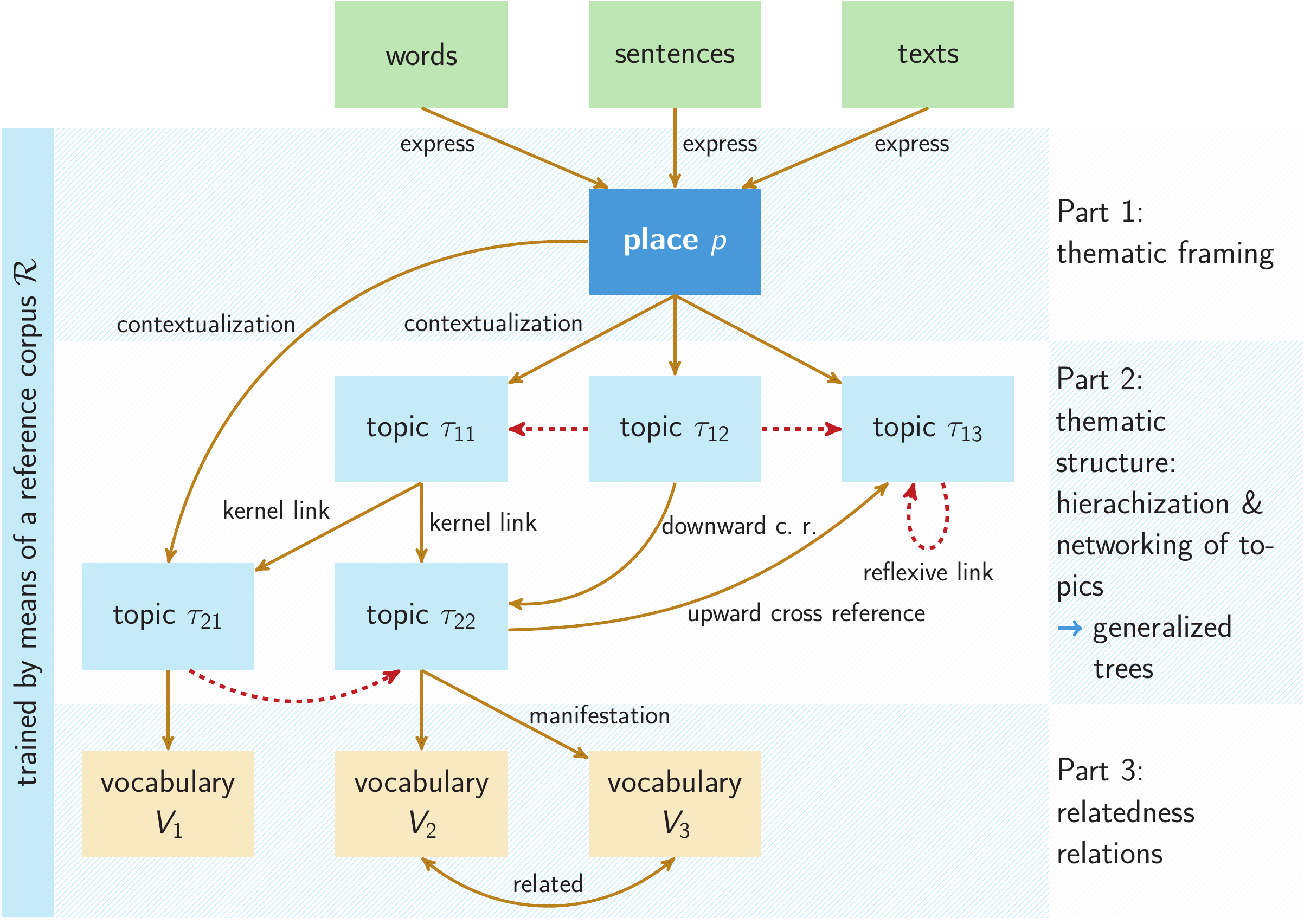}
	\caption{Schema of mapping texts onto hierarchically organized topic networks:
	words, sentences and texts describing a certain thematic frame (e.g.\ a place as the central topic of a city wiki) are mapped onto a topic hierarchy as an example of a so-called generalized tree \cite{Dehmer:Mehler:Emmert-Streib:2007:a,Mehler:2008:a}.
	Based on kernel links of thematic spe\-cia\-li\-za\-tion, the topics are organized hierarchically, whereby this organization is superimposed by up- and downward cross references.
	Dashed links are inferred as a result of modeling the thematic networking of input words, sentences or texts.
	As we assume that the underlying topic model has been trained by means of a reference corpus $\mathcal{R}$ (see Definition \ref{def:Definitional Setting}), each topic is associated with a distribution of lexical elements of $\mathcal{R}$ that are preferably used to \textit{manifest} this topic (see the types $v,w$ in relation to the topics $\alpha, \beta$ in Figure \ref{fig:TFL}).
	This preference relation may be extended to higher-level units such as sentences etc.
	}
	\label{fig:Topic Models Architecture}
\end{figure}

In order to study relations of thematic preference in VGI as a manifestation of distributed cognition, we introduce \textit{Topic Network}s (TN) as an alternative to \textit{Topic Model}s (TM) \cite{Blei:Ng:Jordan:2003,Blei:2012,Steyvers:Griffiths:2007}.
TMs are based on the idea that texts manifest probabilistic distributions of topics which are represented as probability distributions over the lexical constituents of these texts, where these distributions may be affected by style, the underlying genre or any other (syntactic, semantic or pragmatic) criterion of text production \cite{Hsu:Glass:2006,Rosen-Zvi:et:al:2010,Heinrich:2012}.
Regardless of its success, this model is unsuitable for modeling TNs as manifestations of distributed cognitive maps because of the following problems:
\begin{enumerate}[label=P\arabic*]
\item\label{enumerate:R1} \textit{Corpus specificity:}
the corpus specificity of TMs impairs comparability and transferability to ever new corpora, since the topic distributions are learned from the input corpora whose topics are to be modeled. 
This approach apparently cannot use a transferable topic model as a basis for representing the topics of a large number of different corpora.

\item\label{enumerate:R2} \textit{Topic labeling:}
the corpus-specific derivation of topic labels from the input corpora makes it difficult to compare their topic distributions.
As reviewed by Herzog et al.\ \cite{Herzog:John:Mikhaylov:2018}, external resources can be used for this task.
However, there are hardly any such resources for all possible topic combinations -- unless one wants to explore an overarching system such as Wikidata making such a project considerably more difficult due to its size.
The labeling problem can be addressed using, for example, Labeled LDA \cite{Ramage:Hall:Nallapati:Manning:2009}, an approach that leads us into the area of supervised classification, which is also followed here.

\item\label{enumerate:R3} \textit{Scalability:}
instead of dealing with corpora of equally large texts, online communication often leads to sparse, tiny texts that sometimes consist of a single sentence, a single phrase or a single word.
Regardless of the size of the text, we need a procedure that determines its topic distributions so that texts of different size can be compared using topic models of comparable size.
Even if small texts are \textit{post}-processed (after topic modeling) in such a way that their topic distributions are derived from their lexical constituents, such an approach would nevertheless mean to exclude text snippets from the training process. 

\item\label{enumerate:R5} \textit{Rare topics:}
one reason to prefer training by means of corpora as large as Wikipedia is to allow for detecting topics even if they form a kind of \textit{thematic hapax legomenon} in the corpora to be analyzed.
If we try to identify rare topics directly from these corpora, we will probably not detect them, since by definition these corpora do not provide enough information to identify such topics.
In any event, the rarity of evidence about a topic should not be an impediment to identifying its occurrences even at the level of single sentences.

\item\label{enumerate:R7} \textit{Methodical closeness:}
instead of deriving all distributions of all dependent and independent variables as part of the same topic model, one possibly wants to include different information sources that are computed by different methods based on diverse computational paradigms (e.g., ontological approaches to measuring sentence similarities, approaches to word embeddings based on neural networks, topic models, etc.).
In order to enable this, we look for a methodologically open topic model that allows such different resources to be easily integrated.
\end{enumerate}

In a nutshell: 
We are looking for an approach that (i) allows thematic comparisons of previously unforeseen text corpora using an underlying reference corpus, (ii) offers a generic solution to the problem of topic labeling, (iii) is highly scalable and can therefore map even the smallest text snippets to topic distributions, (iv) simultaneously takes rare topics into account and (vii) is methodologically open and expandable.
Such a topic network model is now developed in two steps:
in Section \ref{sec:A Formal Model of Topic Networks} we introduce the underlying formal apparatus. 
This is done by deriving multiplex topic networks from linguistic multilayer networks.
Section \ref{sec:A Procedural Model of Topic Network Analysis} describes a method by which this model is instantiated as a prerequisite for its empirical testing.

\subsection{From Linguistic Multilayer Networks to Multiplex Topic Networks}\label{sec:A Formal Model of Topic Networks}

In this section, we introduce multiplex topic networks. 
This is a type of network that is based on the idea of deriving the networking of topics of textual units by evaluating evidence from different sources of information such as text vocabulary, higher-level text components, distributed authorship or readership, genre, register or medium.
Since these sources of evidence can be explored in different compositions, this can lead to different perspectives on the salience and networking of the topics addressed by the same texts. 
Topic networks are multiplex in precisely this respect: 
the different evidence-providing perspectives may lead to different topic networks that allow comparisons to be made through which differences in the linguistic, social or otherwise contextual embedding of thematizations become visible.
This concept of a multiplex topic network is now being generically formalized.

To introduce multiplex topic networks, we start with defining \textit{linguistic multilayer networks} (Definition \ref{def:Linguistic Multilayer Networks}) whose layeredness allows for distinguishing several (non-)linguistic information sources of topic networking.
We refer to supervised topic classifiers trained by means of large reference corpora to tackle the challenges \ref{enumerate:R1}, \ref{enumerate:R2}, \ref{enumerate:R3} and \ref{enumerate:R5}.
Based thereon, we introduce so-called \textit{text topic networks} (Definition \ref{def:Text Topic Network}), which evaluate intra- and intertextual relations for the purpose of topic networking.
Then, we introduce \textit{two-level topic networks} (Definition \ref{def:Two-layer Topic Network}) and exemplify them by \textit{author} (Definition \ref{def:Author Topic Network}) and \textit{word topic networks} (Definition \ref{def:Word Topic Network}) which explore relations of (co-)authorship and lexical relatedness, respectively, as sources of topic networking.
These notions are generalized to arrive at \textit{$n$-level topic networks} (Definition \ref{def:n-Layer Topic Network}) which are based on $n > 1$ informational sources of topic networking (cf.\ challenge \ref{enumerate:R7}).
Finally, \textit{multiplex topic networks} are defined as families of $n$-level topic networks (Definition \ref{def:Multiplex Topic Network}) representing the networking of the same set of topics from different informational perspectives and, thus allowing for mapping the thematic dynamics, for example, of descriptions of the same place. 

\begin{definition}\label{def:Linguistic Multilayer Networks}
Let $X = \{x_1,\ldots, x_n\}$ be a corpus of texts. 
A \textbf{\textit{Linguistic Multilayer Network}} (LMN)\footnote{Mehler \cite{Mehler:2008:a} speaks of \textit{multilevel graphs}. 
See Boccaletti et al.\ \cite{Boccaletti:et:al:2014} for a comprehensive overview of related notions whose formalism is used here. See Stella et al.\ \cite{Stella:et:al:2018} for an example of a multiplex network of lexical systems.} 
\begin{alignat}{2}\label{math:multilayer network}
\mathcal{L}(X,l) &= 
(\mathbb{L}, \mathbb{C})\\ 
\mathbb{L} &= \{ L_{i} = (\Prime{V}{i}, \Prime{A}{i}, \Prime{\mu}{i}, \Prime{\nu}{i}, \Prime{\lambda}{i}, \Prime{\kappa}{i}) \mid i = 1..l\} \\
\mathbb{C} &= \{ C_{i\myto j} = (\Prime{V}{i\myto j}, \Prime{A}{i\myto j}, \Prime{\mu}{i\myto j}, \Prime{\nu}{i\myto j}, \Prime{\lambda}{i\myto j}, \Prime{\kappa}{i\myto j}) \mid i,j = 1..l\!: i\not=j\}
\end{alignat}
is a tuple of two sets of directed graphs such that the set of \textit{kernel layers} $\mathbb{L}$ consists of a pivotal text layer and several derivative layers, that is, a coauthoring layer, a language-systematic word layer and possibly several layers modeling the networking of constituents of the pivotal texts: 
\begin{enumerate}
	\item the \textit{pivotal text layer} 
	$L_1 = (\Prime{V}{1}, \Prime{A}{1}, \Prime{\mu}{1}, \Prime{\nu}{1}, \Prime{\lambda}{1}, \Prime{\kappa}{1})$, also called \textit{text network}, is spanned by texts of the corpus $\Prime{V}{1} = X$ such that $\Prime{A}{1}$ is manifesting intra- (as in the case of reflexive arcs) or intertextual relations, 
	\item the \textit{author layer:} 
	$L_2 = (\Prime{V}{2}, \Prime{A}{2}, \Prime{\mu}{2}, \Prime{\nu}{2}, \Prime{\lambda}{2}, \Prime{\kappa}{2})$, also called \textit{agent network}, is spanned by the network of agents (co-)authoring the texts in $\Prime{V}{1}$ and their social relations,
	\item the \textit{lexicon layer}
	$L_3 = (\Prime{V}{3}, \Prime{A}{3}, \Prime{\mu}{3}, \Prime{\nu}{3}, \Prime{\lambda}{3}, \Prime{\kappa}{3})$, also called \textit{word network}, is spanned by the lang\-u\-a\-ge-sy\-s\-te\-ma\-tic lexical signs (i.e., lexemes and related units) used by agents of $\Prime{V}{2}$ as part of their agent lexica to author the texts in $\Prime{V}{1}$, 
	\item for $3 < i \le l' < l$, $L_i = (\Prime{V}{i}, \Prime{A}{i}, \Prime{\mu}{i}, \Prime{\nu}{i}, \Prime{\lambda}{i}, \Prime{\kappa}{i})$ is called a \textit{constituent layer} modeling the networking of (e.g., lexical, phrasal, sentential etc.) constituents of texts $x\in \Prime{V}{1}$ such that $\Prime{A}{i}$ maps intra- (e.g., anaphoric) or intertextual (e.g., sentence similarity) relations,
	\item for $l' < i \le l$, $L_i = (\Prime{V}{i}, \Prime{A}{i}, \Prime{\mu}{i}, \Prime{\nu}{i}, \Prime{\lambda}{i}, \Prime{\kappa}{i})$ is called a \textit{contextual layer} modeling the networking of units (e.g., media, genres, registers \cite{Halliday:Hasan:1989} etc.) of the contextual embedding of texts $x\in \Prime{V}{1}$ such that $\Prime{A}{i}$ maps for example relations of the switching, merging or embedding \cite{Ventola:1987,Clarke:1996} of these contextual units,
	\item for each $i,j\in\{1,\ldots, l\}$, $i\not= j$, $C_{i\myto j}\in \mathbb{C}$, $|\mathbb{C}| = l(l-1)$, is called a \textit{margin layer} where 
	$\Prime{V}{i\myto j} = \Prime{V}{i}\cup \Prime{V}{j}$, 
	$\Prime{A}{i\myto j} \subseteq \Prime{V}{i}\times \Prime{V}{j}$, 
	$\Prime{\mu}{i\myto j} = \Prime{\mu}{i}\cup \Prime{\mu}{j}$ and 
	$\Prime{\lambda}{i\myto j} = \Prime{\lambda}{i}\cup \Prime{\lambda}{j}$.
\end{enumerate}
For $i,j = 1..l, i\not=j$, $\Prime{\mu}{i}$, $\Prime{\mu}{i\myto j}$ are vertex weighting functions,
$\Prime{\nu}{i}$, $\Prime{\nu}{i\myto j}$ are arc weighting functions,
$\Prime{\lambda}{i}$, $\Prime{\lambda}{i\myto j}$ are vertex labeling functions
and $\Prime{\kappa}{i},\Prime{\kappa}{i\myto j}$ arc labeling functions.
We say that the linguistic multilayer network $\mathcal{L}(X,l)$ is \textit{spanned over the text corpus $X$} and \textit{layered into $l$ layers}.
\hfill $\Box$
\end{definition}

\begin{exmp}\label{ex:LMN}
To illustrate our definitions, we construct a minimized example.
Suppose a corpus of four texts $V_1 = X = \{x_1, x_2, x_3, x_4\}$, each containing three lexemes $x_1 = \{w_{1}, w_2, w_3\}$, $x_2 = \{w_{1}, w_2, w_4\}$, $x_3 = \{w_{5}, w_6, w_7\}$, $x_4 = \{w_{4}, w_8, w_{9}\}$ (for reasons of simplicity we exemplify texts as bag-of-words), that is, $V_3 = \{w_1,\ldots,w_9\}$, $V_{3.1} = \{w_1,\ldots, w_9, x_1,\ldots, x_4\}$ and $A_{3.1} = \{
(w_1,x_1),(w_2,x_1),(w_3,x_1), \ldots, (w_4,x_4),(w_8,x_4),(w_9,x_4) \}$.
Further, we assume four authors $V_2 = \{a_1, a_2, a_3, a_4\}$ such that $a_1$ and $a_2$ co-authored $x_1$ and $x_2$, while $a_3$ and $a_4$ co-authored $x_3$ and $x_4$, that is, $V_{2.1} = \{a_1,\ldots, a_4, x_1,\ldots, x_4\}$ and $A_{2.1} = \{
(a_1,x_1),$ $(a_2,x_1),$ $(a_1,x_2),$ $(a_2,x_2),$
$ (a_3,x_3),$ $(a_4,x_3),$ $(a_3,x_4),$ $(a_4,x_4)
\}$.
Further, we assume that the texts $x_1, x_2$ are linked by some intertextual coherence relation (e.g.\ by a rhetorical relation, an argument relation or by some hyperlink) as are the texts $x_3, x_4$ so that $A_1 = \{(x_1, x_2), (x_3,x_4)\}$.
Note that additional arcs of the layers $L_1, L_2, L_3$ will be generated according to the subsequent definitions.
For simplicity reasons we assume all weighting functions to be limited to the set $\{0,1\}$ of vertex/arc weights.
Since we assume no additional constituent layer we get $l = 3$. 
Thus, any linguistic multilayer network $\mathcal{L}(X,3)$ based on this setting is layered into three layers.
\end{exmp}

Throughout this paper, we use the following simplifying notation:
for any graph $G = (V, A, \lambda)$ of order $|G| = |V|$, arc set $A\subseteq V^2$ of size $|A|$ and vertex labeling function $\lambda$ and any vertex $v\in V$, we write $\dot{v} = \Prime{\lambda}{}(v)$. 
Thus, for any two graphs $G_{i}, G_{j}$ with vertex labeling functions $\Prime{\lambda}{i}$ and $\Prime{\lambda}{j}$, for which $\Prime{\lambda}{i}(v) = \Prime{\lambda}{j}(w)$, $v\in \Prime{V}{i}, w\in \Prime{V}{j}$, we can write $\dot{v} = \dot{w}$.
Further, for any function $f\!: X\times Y\to Z$, for which $f(x,y) = z$, we use the following alternative notations: 
\begin{equation}
f(x,y) = z \Leftrightarrow \Stack{x}{y}{f} = z \Leftrightarrow \StackX{x}{y}{f} = z \Leftrightarrow f_y(x) = z
\end{equation}
Finally, for any function $f\!: Z^n\to Z$ we introduce the following notation based on square brackets:
\begin{equation}
f(\ldots, \StackX{x}{y}{f}, \StackX{y}{x}{g}\ldots) = z
\Leftrightarrow f[\ldots, \StackXY{x}{y}{f}{g}\ldots] = z 
\Leftrightarrow f[\ldots, \XStackY{x}{y}{f}{g}\ldots] = z 
\end{equation}
To leave no room for ambiguity, we assume that expressions of the sort $\Stack{x}{y}{f}, \Stack{y}{x}{g}$ are replaced from left to right into expressions of the sort $\XStackY{x}{y}{f}{g}$.
Henceforth, a structure such as $\Stack{x}{y}{f}$ will be called \textit{information link}. 
Based on Definition \ref{def:Linguistic Multilayer Networks} we start now with introducing text topic networks using the following auxiliary notion:

\begin{definition}\label{def:Definitional Setting}
Let $\classification = (V_{\classification}, A_{\classification})$ be a directed \textit{Generalized Tree} (GT) according to \cite{Mehler:2009:b,Mehler:2009:c} representing a hierarchical topic structure, henceforth called \textit{Reference Classification System} (RCS), that is spanned by kernel arcs which are possibly superimposed by upward, downward, lateral, sequential, external or reflexive arcs.\footnote{See Figure \ref{fig:Topic Models Architecture} for an example of a GT. This notion is required since we may decide for using, for example, the category system of Wikipedia as an RCS, which spans a GT \cite{Mehler:2009:c}.}
That is, vertices $t\in V_{\classification}$ represent topics, while kernel arcs $(t,u)\in A_{\theta}$ represent subordination relations according to which $u$ is a thematic specialization of $t$.
Let further $\classify$ be a \textit{hierarchical text classifier} \cite{Sebastiani:2002} taking values in $V_{\classification}$ that has been trained, validated and tested by means of a reference corpus $\mathcal{R}$. 
Let now $\mathcal{L}(X,l) = (\mathbb{L}, \mathbb{C})$ be a LMN spanned over the text corpus $X$ and layered into $l$ layers.
We call the structure 
\begin{equation}
\mathcal{S} = (\classification, \classify, \mathcal{L}(X,l))
\end{equation}
a \textbf{\textit{Definitional Setting}} for defining topic networks. 
\end{definition}

\begin{exmp}\label{ex:Definitional Setting}
Given the LMN of Example \ref{ex:LMN}, the \textit{Dewey Decimal Classification} (see Section \ref{sec:A Procedural Model of Topic Network Analysis} and Figure \ref{fig:DDC Subtree}) and the topic classifier $\theta$ of \cite{Uslu:Mehler:Niekler:Baumartz:2018}, which uses the DDC as its reference classification system $\classification$, a definitional setting is exemplified by $(\textit{DDC}, \theta, \mathcal{L}(X,3))$. 
More specifically, by $t_1, t_2, t_3$ we will denote three topic labels of the third level of the DDC so that $V_{\classification} = \{\ldots, t_1, t_2, t_3,\ldots \}$.
Note that by using the DDC as a reference classification, the generalized tree of Definition \ref{def:Definitional Setting} is reduced to a tree (see Section \ref{sec:A Procedural Model of Topic Network Analysis} for more details).
\end{exmp}
	
\begin{figure}[t]
	\includegraphics[width=0.65\textwidth]{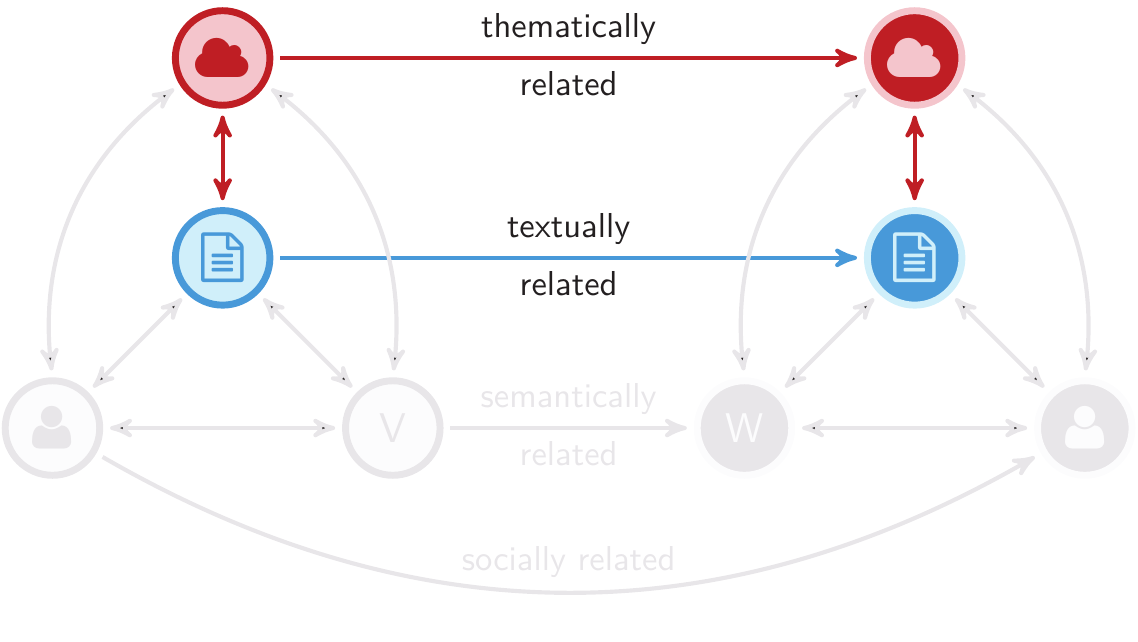}
	\caption{Schematic depiction of the informational sources of linking topics (red vertices) in \textit{text topic networks} as a function of the textual relatedness of two texts (blue vertices) (belonging to layer $L_1$ of a corresponding LMN -- see Definition \ref{def:Linguistic Multilayer Networks}. 
	Bidirectional red arcs denote arcs of the corresponding margin layers:
	in the present case, this concerns the relation between texts and topics (see below).
	Relations of thematic relatedness are inferred in this example (see Definition \ref{def:Text Topic Network}). 
	Gray nodes and arcs indicate unconsidered sources of evidence.
	}
	\label{fig:TTN:ground:scheme}
\end{figure}

\begin{figure}[t]
	\includegraphics[width=1.00\textwidth]{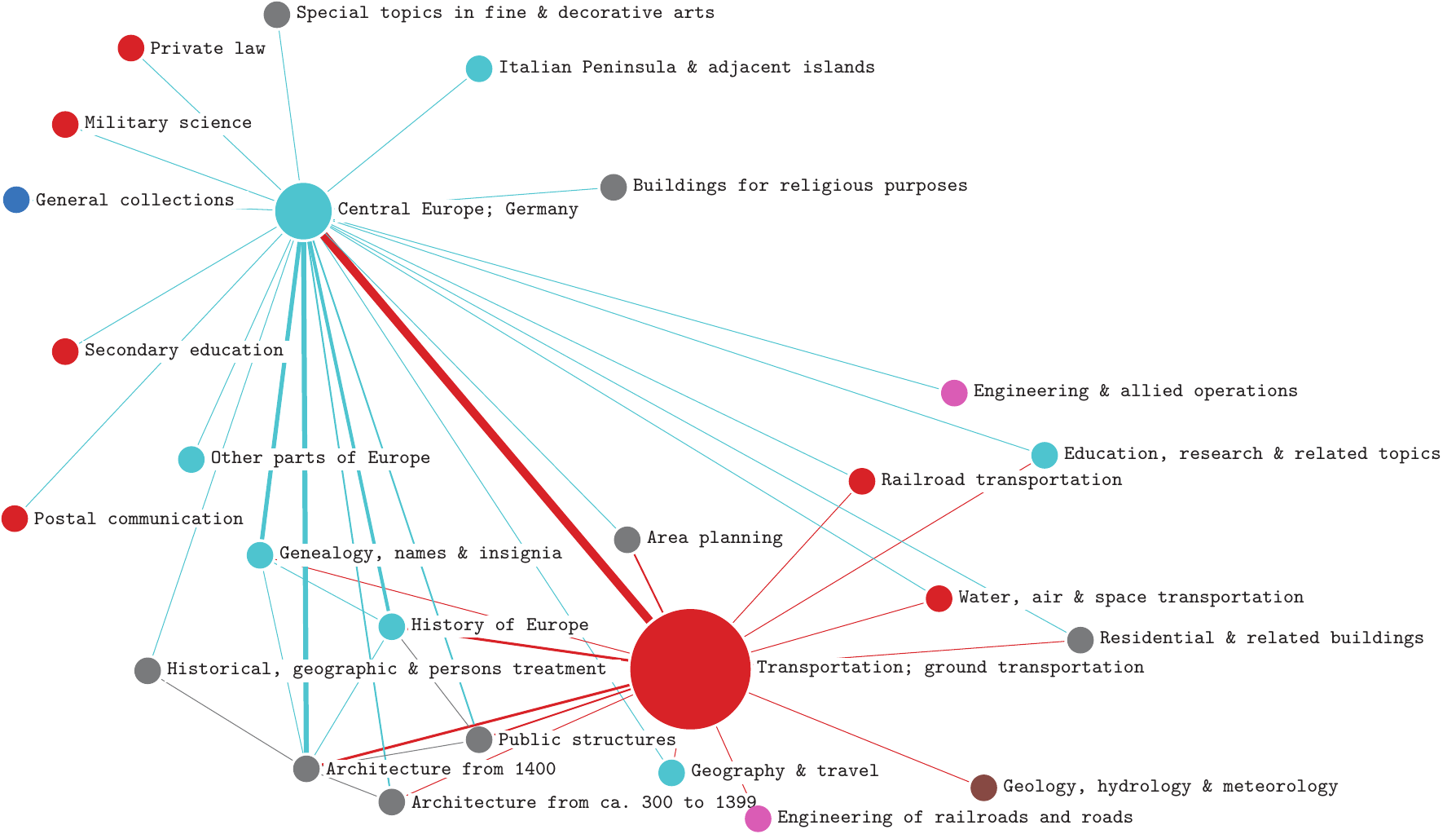}
	\caption{Visualization of a segment of the TTN of the city wiki Dresden (\url{http://www.stadtwikidd.de/wiki/Hauptseite}) using the 3rd level of the DDC as the underlying RCS for the definition of topics according to Section \ref{sec:A Procedural Model of Topic Network Analysis}.
	The segment shows the highest weighted topics and their (undirected) links.
	Edges have been colored to show the two centers of this graph.
	}
	\label{fig:Dresden city wiki}
\end{figure}
	
\begin{definition}\label{def:Text Topic Network}
Given a definitional setting $\mathcal{S} = (\classification, \classify, \mathcal{L}(X,l))$ according to Definition \ref{def:Definitional Setting}, a \textbf{\textit{Text Topic Network}} (TTN) is a vertex- and arc-weighted simple directed graph 
\begin{equation}\label{math:Cotextual Topic Network}
\topicnetwork(L_1) = \topicnetwork(L_1, \{\}) = (\Prime{V}{}, \Prime{A}{}, \Prime{\mu}{}, \Prime{\nu}{}, \Prime{\lambda}{}, \Prime{\kappa}{})
\end{equation}
with vertex set $\Prime{V}{}$ and arc set $\Prime{A}{}\subseteq \Prime{V}{}^2$ which is said to be \textit{derived from} $\mathcal{S}$ and \textit{inferred from} $L_1$ by means of the optional classifier $\rclassify$ and the monotonically increasing functions $\alpha, \beta, \gamma, \delta\!: \mathbb{R}^+_0\to \mathbb{R}^+_0$ iff $\forall v\in \Prime{V}{}$ and $\forall a = (v,w)\in \Prime{A}{}$:
\begin{alignat}{2}
\Prime{\mu}{}(v) &= 
\alpha\Bigl(
	\sum_{x\in \Prime{V}{1}} \beta(\classify(x, \Prime{\lambda}{}(v)), \rclassify(\Prime{\lambda}{}(v), x)) \Bigr) 
	= 
	\alpha\Bigl(\sum_{x\in \Prime{V}{1}} 
		\beta(\classify_x(\dot{v}), 
		\theta_{\dot{v}}^{\shortgets}(x))
	\Bigr) 
	\label{math:Cotextual Topic Network:1} \\
&= 
\alpha\Bigl(
	\sum_{x\in \Prime{V}{1}} 
	\beta(
		\StackX{x}{\dot{v}}{\classify}, 
		\StackX{\dot{v}}{x}{\rclassify}
	)
\Bigr) 
= 
\alpha\Bigl(
	\sum_{x\in \Prime{V}{1}} 
	\beta(
		\StackXY{x}{\dot{v}}{\classify}{\rclassify} 
	)
\Bigr)  > 0 \label{math:Cotextual Topic Network:2} \\
\nu_{}(a) 
&= 
\gamma\Bigl(
	\sum_{x,y\in \Prime{V}{1}} 
	\delta(
		\classify(x, \Prime{\lambda}{}(v)), 
		\rclassify(\Prime{\lambda}{}(v), x), 
		\classify(y, \Prime{\lambda}{}(w)), 
		\rclassify(\Prime{\lambda}{}(w), y), 
		\Prime{\nu}{1}(x,y)
	)
\Bigr) \label{math:Cotextual Topic Network:3} \\
&= 
\gamma\Bigl(
	\sum_{x,y\in \Prime{V}{1}} 
	\delta[
		\StackXY{x}{\dot{v}}{\classify}{\rclassify}, 
		\StackXY{y}{\dot{w}}{\classify}{\rclassify}, 
		\StackX{x}{y}{\Prime{\nu}{1}}
	]
\Bigr) > 0
\label{math:Cotextual Topic Network:4}
\end{alignat} 
$\Prime{\mu}{}\!:\Prime{V}{}\to \mathbb{R^+}$ is a vertex weighting function, $\Prime{\nu}{}\!:\Prime{A}{}\to \mathbb{R^+}$ an arc weighting function, $\Prime{\lambda}{}\!: \Prime{V}{}\to V_{\classification}$ an injective vertex labeling function, $V_{\classification}(V) = \{\lambda(v)\,|\, v\in V\}\subseteq V_{\classification}$, and $\Prime{\kappa}{}$ an injective arc labeling function.
$\topicnetwork(L_1)$ is called a \textit{one-layer topic network} that is generated by the \textit{generating layer} $L_1$.
\hfill $\Box$
\end{definition}

Formulas \ref{math:Cotextual Topic Network:2} and \ref{math:Cotextual Topic Network:4} require that the weighting values for nodes and arcs are greater than $0$: otherwise, the candidate vertices and arcs do not exist in the TTN.
$\rclassify$ is a classifier mapping pairs $(t,x)$ of topics $t\in V_{\classification}$ and texts $x$ onto real numbers indicating the extent to which $x$ is a \enquote{prototypical} instance of $t$.\footnote{Obviously, the textual arguments of the functions $\classify$ and $\rclassify$ are not restricted to elements of $X$.}

\begin{exmp}\label{ex:TTNs}
Given Example \ref{ex:Definitional Setting}, we assume that $\lambda(v_1) = t_1, \lambda(v_2) = t_2, \lambda(v_3) = t_3$ and 
$\theta(x_1, t_1) = 1$, $\theta(x_2, t_2) = 1$, $\theta(x_3, t_3) = \theta(x_4, t_3) = 1$ so that $V = \{v_1, v_2, v_3\}$. 
In our example, we disregard $\rclassify$. 
Further, we assume that the functions $\alpha, \beta, \gamma, \delta$ are identify functions.
Thus, $\mu(v_1) = \mu(v_2) = 1$ and $\mu(v_3) = 2$.
Now, we can generate a topic link between $v_1$ and $v_2$ by exploring the intertextual relation $(x_1,x_2)\in A_1$: 
To this end, we assume that 
\[
\delta[
	\StackXY{x}{\dot{v}}{\classify}{\rclassify}, 
	\StackXY{y}{\dot{w}}{\classify}{\rclassify}, 
	\StackX{x}{y}{\Prime{\nu}{1}}
] \gets 
\delta[
	\StackX{x}{y}{\Prime{\nu}{1}}
] \gets \text{id}(\StackX{x}{y}{\Prime{\nu}{1}}) = \StackX{x}{y}{\Prime{\nu}{1}}
\]
so that $\nu((v_1,v_2)) = 1$.
By analogy to this case, we link topic $v_3$ by means of a reflexive link so that $A = \{(v_1,v_2), (v_3,v_3)\}$.
Note that these simplifications are made for simplicity's sake only:
Section \ref{sec:A Procedural Model of Topic Network Analysis} will elaborate a realistic weighting scenario.
However, the function of the latter illustration is to show that by the intertextual linkage of both texts, we get evidence about the linkage of the topics instantiated by these texts.
TTNs always operate according to this premise: they network topics as a function of the networking of an underlying set of texts.
Figure \ref{fig:TTN:ground:scheme} gives a schematic depiction of this scenario, which is varied subsequently to illustrate the other types of topic networks developed in this paper.
\end{exmp}

\begin{figure}[t]
	\includegraphics[width=0.45\textwidth]{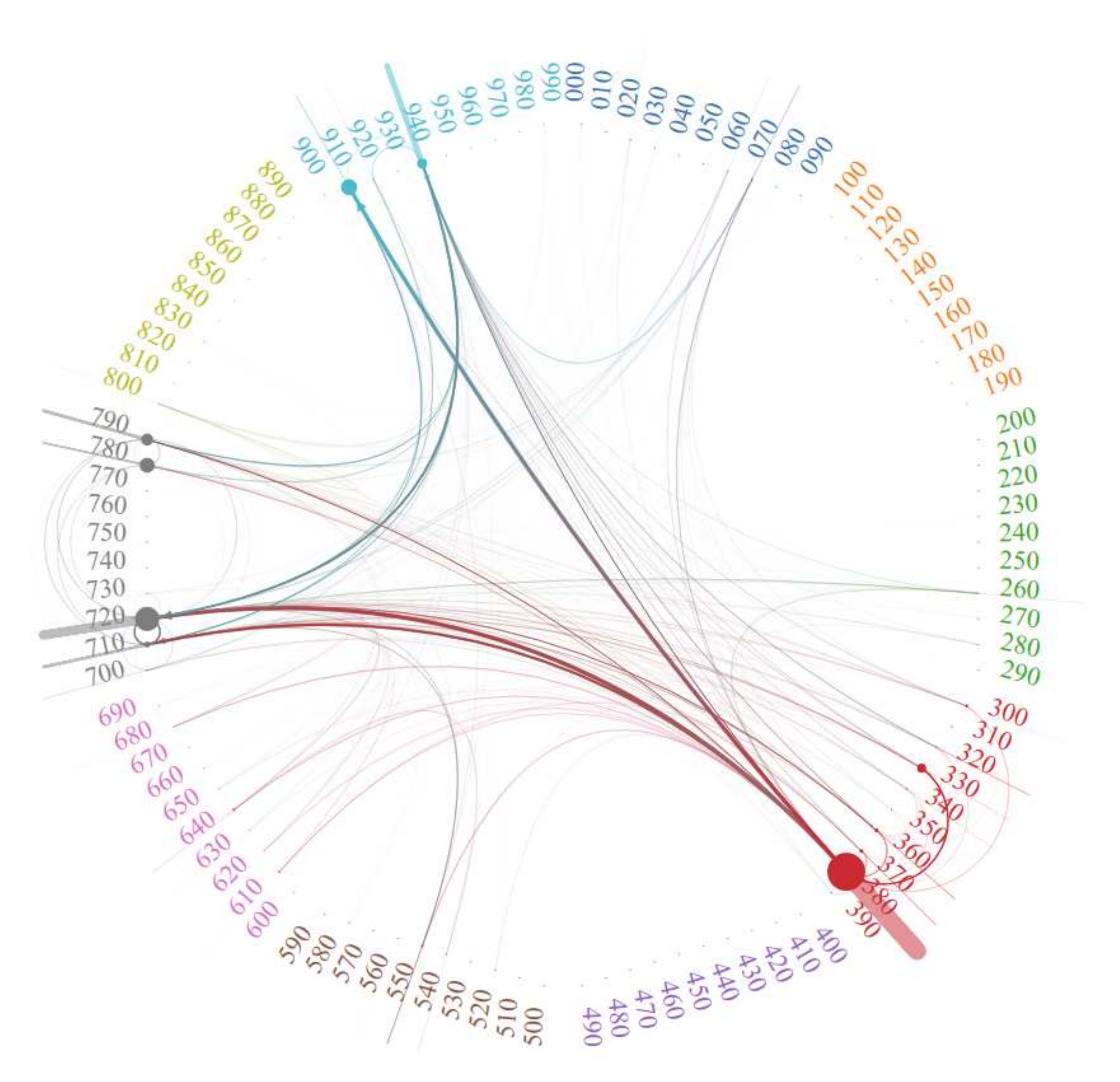}
	\includegraphics[width=0.45\textwidth]{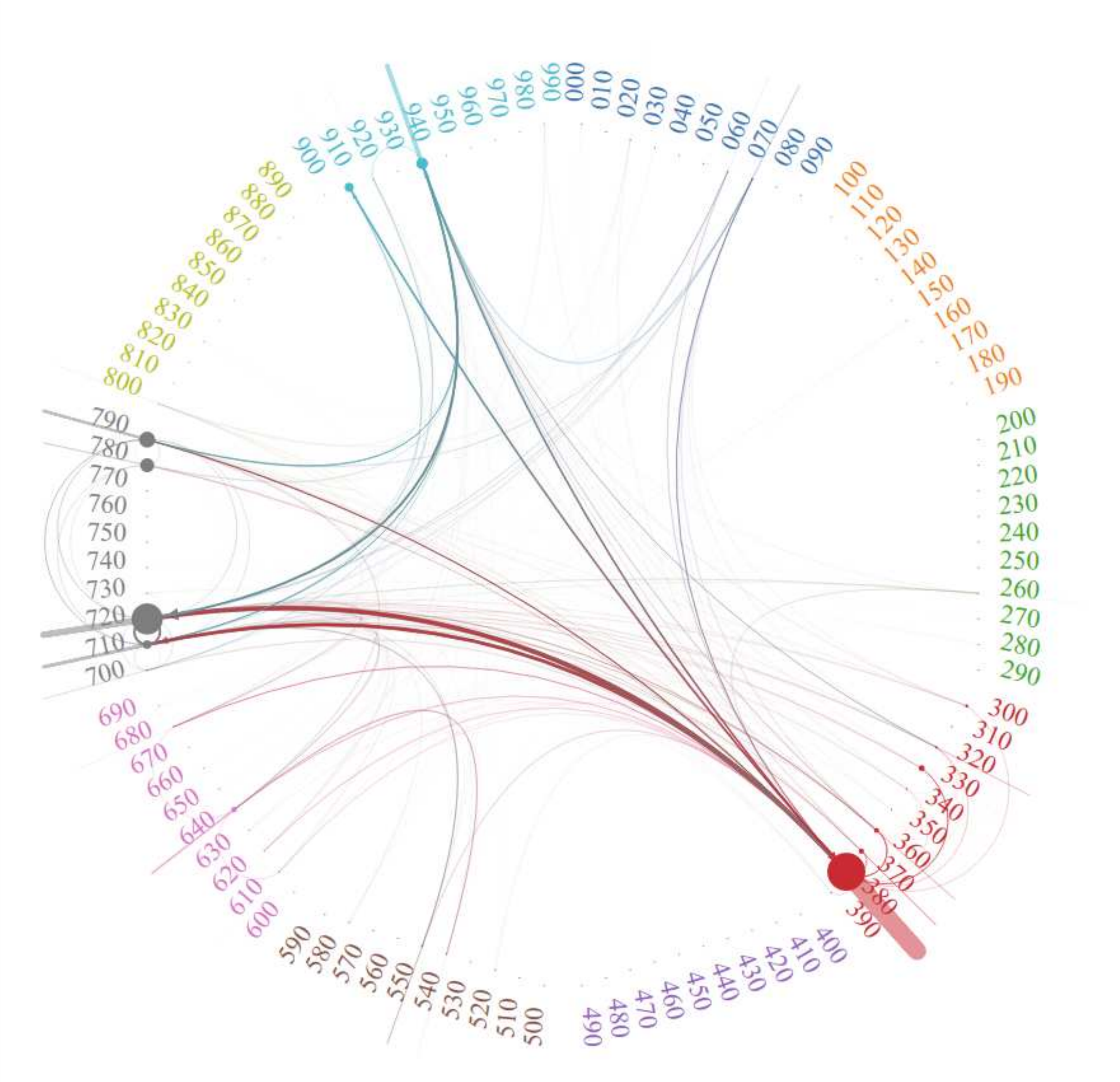}
	\includegraphics[width=0.45\textwidth]{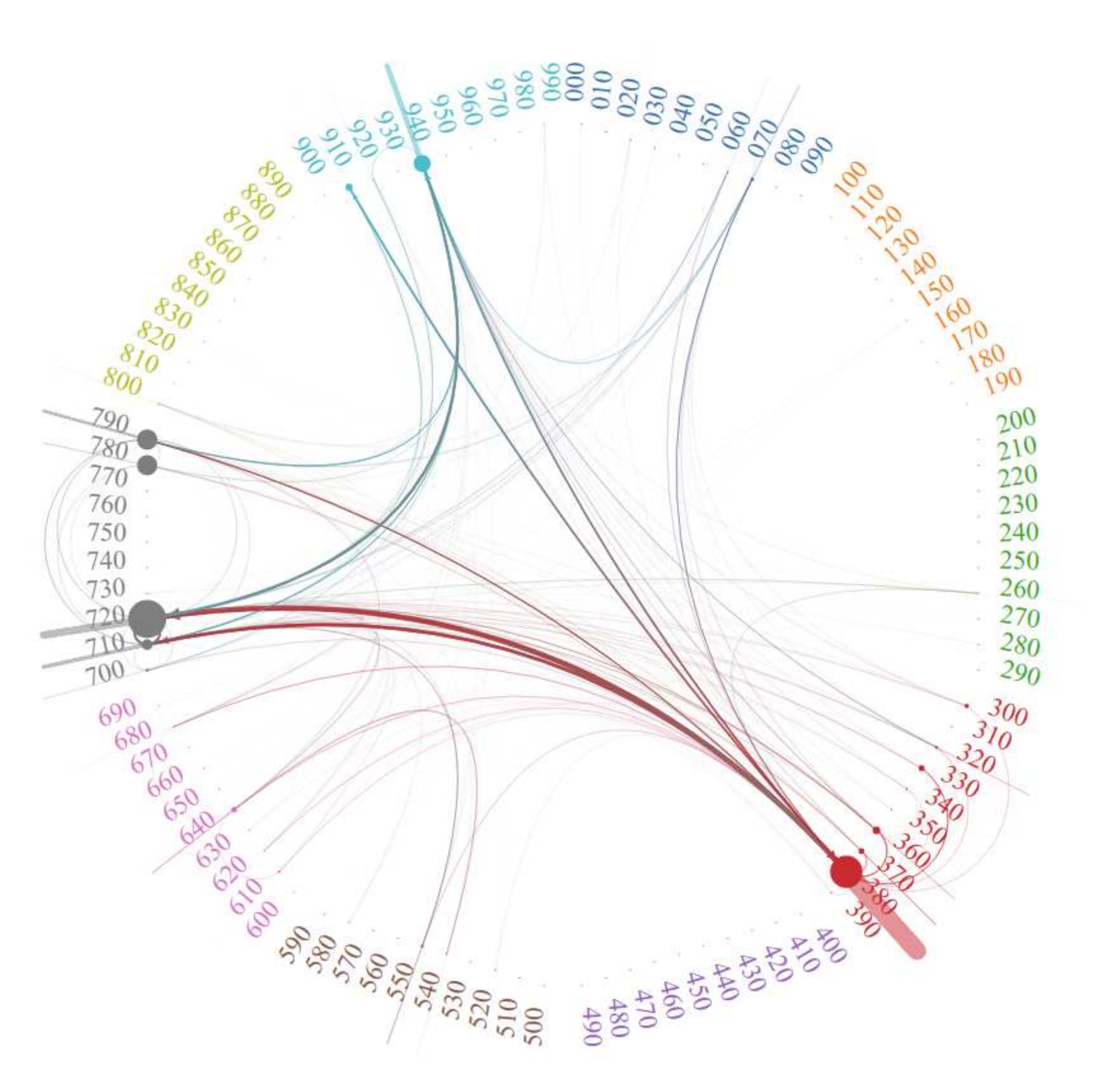}
	\caption{Visualizations of a TTN (top left) and two corresponding ATNs.
	The TNs are derived from the city wiki \textsc{München} (\url{https://www.muenchenwiki.de/wiki/Hauptseite}) (see Section \ref{sec:Experimentation} and Table \ref{table:general_wiki_stats} for statistics about this wiki) using the procedural model of Section \ref{sec:A Procedural Model of Topic Network Analysis}.
	Top right shows the ATN for which (co-)authorship activities are estimated by means of Wikipedia (see Section \ref{sec:Module 3: Network Induction}).
	The ATN for which these activities are estimated via the wiki itself is displayed below.
	The visualizations are carried out by means of PolyViz \cite{Uslu:Mehler:2018} regarding the \nth{2} level of the DDC:
	nodes are labeled (with numbers denoting the respective \nth{2}-level class) and colored to encode their membership to one of the top 10 DDC classes (see appendix).
	The higher the weight of a topic, the larger the node, and the higher the weight of an arc, the thicker the line.
	Node and line sizes are defined relative to the maximum vertex and arc weights of the underlying network.
}
\label{fig:TTN and ATNs of Muenchen City Wiki}
\end{figure}

A concrete example of a TTN that is derived from the articles of the so-called Dresden wiki (see Section \ref{sec:resources}) is depicted in Figure \ref{fig:Dresden city wiki}.
It shows the highest weighted topics addressed by these articles and their (undirected) links.
The TTN has been computed by means of the procedural model of Section \ref{sec:A Procedural Model of Topic Network Analysis}.
Evidently, the topic \textit{Transportation; ground transportation} is most prominent in this wiki followed by the topic \textit{Central Europe; Germany}.
Most topics belong to the areas \textit{transportation} (red), \textit{geography and history} (turquoise) and \textit{architecture} (gray) (for the color code see the appendix).
More examples of TTNs can be found in Figures \ref{fig:TTN and ATNs of Muenchen City Wiki}, \ref{fig:First and Second Orbit of Integralrechnung} and \ref{fig:First and Second Orbit of Kernkraftwerk}.

Arguments of the sort $\Stack{x}{\dot{v}}{\classify}$ can be used to quantify evidence about text $x$ as an instance of topic $\dot{v}$:
the more evidence of this sort, the higher possibly the impact of $x$ in Formula \ref{math:Cotextual Topic Network:2}, the higher possibly the final weight of $v$.
The adverb \textit{possibly} refers to what is licensed by the parameters $\gamma, \delta$.
Arguments of the sort $\Stack{x}{y}{\Prime{\nu}{1}}$, where $x\not=y$, can be used to quantify evidence that text $x$ is intertextually linked to text $y$:
the more evidence of this sort, the higher possibly the weight of the link from $x$ to $y$, the higher possibly the influence of this link onto the weight of the link from topic $v$ to topic $w$ in Formula \ref{math:Cotextual Topic Network:4}.\footnote{In cases in which there is no explicit information about intertextual links, one can use functions of aggregated word embeddings of the lexical constituents of texts to calculate their intertextual similarity.}
In this and related definitions, we do not fully specify the functions $\classify, \rclassify, \alpha, \beta, \gamma, \delta$ to leave enough space for different instances of topic networks.

\begin{figure}[t]
	\includegraphics[width=0.75\textwidth]{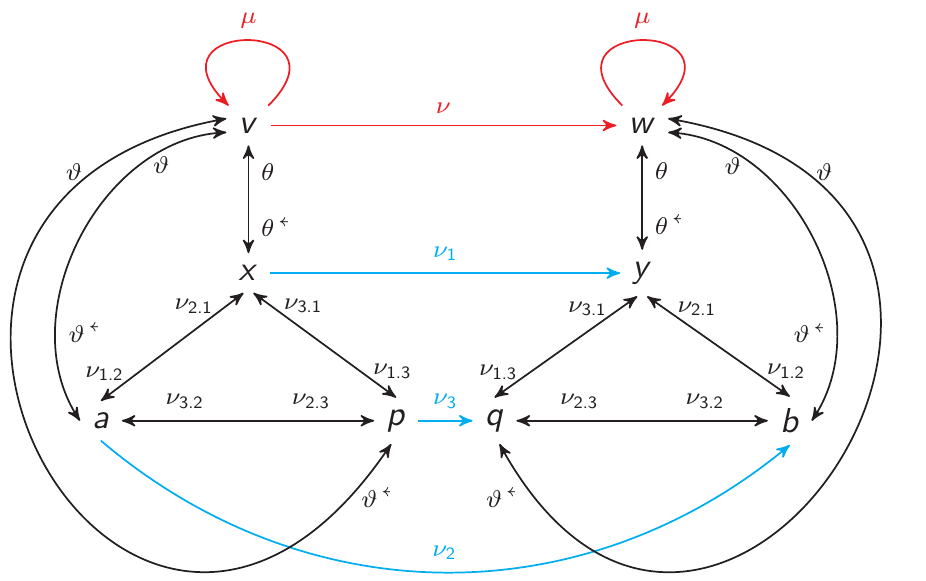}
	\caption{A diagrammatic depiction of inferred arcs (red) in topic networks, inferred by means of various arcs (black and blue) of an underlying LMN. 
	Orientation of inferred arcs is provided by three types of input arcs (blue).
	$x,y\in \Prime{V}{1}$ denote two texts, $\Author{a},\Author{b}\in \Prime{V}{2}$ denote two authors working on $x$ and $y$, respectively, $\Word{p},\Word{q}\in \Prime{V}{3}$ denote two lexical units occurring in $x$ and $y$, respectively. 
	Inferred weights of vertices are denoted by means of (red) reflexive arcs.}\label{fig:Diagrammatic Depiction of Inferring Topic Networks}
\end{figure}

Definition \ref{def:Text Topic Network} relies on the pivotal text layer for deriving topic networks.
To integrate further layers into the process of inferring topic networks, we introduce the following generalized schema:

\begin{definition}\label{def:Two-layer Topic Network}
Given a definitional setting $\mathcal{S} = (\classification, \classify, \mathcal{L}(X,l))$ according to Definition \ref{def:Definitional Setting}, an \textbf{$(L_1,\mathbb{L}')$-\textit{Topic Network}}, $\mathbb{L}'\in \{\emptyset\}\cup \{ \{L_{i}\}\mid i\in \{2,\ldots, l\}\}$, is a vertex- and arc-weighted simple directed graph 
\begin{equation}\label{math:Two-layer Topic Network}
\topicnetwork(L_1, \mathbb{L}') = 
(\Prime{V}{}, \Prime{A}{}, \Prime{\mu}{}, \Prime{\nu}{}, \Prime{\lambda}{}, \Prime{\kappa}{}) 
\end{equation}
which is said to be \textit{derived from} $\mathcal{S}$ and 
\textit{inferred from} $L_1$ and the elements of $\mathbb{L}'$ by means of the optional classifiers $\rclassify, \oclassify\!: \Prime{V}{i}\times V_{\classification}\to \mathbb{R}_0^+, \roclassify\!: V_{\classification} \times \Prime{V}{i} \to \mathbb{R}_0^+$ and monotonically increasing functions ${\alpha}, {\beta}, {\gamma}, {\delta}\!: \mathbb{R}^+_0\to \mathbb{R}^+_0$ iff $\forall v\in \Prime{V}{}$ and $\forall a = (v,w)\in \Prime{A}{}$:
\begin{alignat}{1}
\label{math:two-layer topic network:vertices}
\Prime{\mu}{}(v) 
&= 
{\alpha}\Bigl(
	\sum_{x\in \Prime{V}{1}, r\in \Prime{V}{i}} 
	{\beta}[ 
		\StackXY{x}{\dot{v}}{\classify}{\rclassify}, 
		\StackXY{r}{\dot{v}}{\oclassify}{\roclassify}, 
		\StackXY{r}{x}{\Prime{\nu}{i\myto 1}}{\Prime{\nu}{1\myto i}}
	]
\Bigr) > 0 \\ 
\label{math:two-layer topic network:arcs}
\nu_{}(a) 
&= 
{\gamma}\Bigl( 
	\sum_{x,y\in \Prime{V}{1}, r,s\in \Prime{V}{i}} 
	{\delta}[
	 \StackXY{x}{\dot{v}}{\classify}{\rclassify},
	 \StackXY{y}{\dot{w}}{\classify}{\rclassify},
	 \StackXY{r}{\dot{v}}{\vartheta}{\roclassify},
	 \StackXY{s}{\dot{w}}{\vartheta}{\roclassify},
	 \StackXY{r}{x}{\Prime{\nu}{i\myto 1}}{\Prime{\nu}{1\myto i}}, 
	 \StackXY{s}{y}{\Prime{\nu}{i\myto 1}}{\Prime{\nu}{1\myto i}}, 
	 \StackX{r}{s}{\Prime{\nu}{i}},
	 \StackX{x}{y}{\Prime{\nu}{1}}
	]
\Bigr) 
> 0 
\end{alignat}
where $\mathbb{L}' = \{L_{i}\}$.
$\Prime{\mu}{}\!:\Prime{V}{}\to \mathbb{R^+}$ is a vertex weighting function, $\Prime{\nu}{}\!:\Prime{A}{}\to \mathbb{R^+}$ an arc weighting function, $\Prime{\lambda}{}\!: \Prime{V}{}\to V_{\classification}$ an injective vertex labeling function, $V_{\classification}(V) = \{\lambda(v)\,|\, v\in V\}\subseteq V_{\classification}$, and $\Prime{\kappa}{}$ an injective arc labeling function.
For $\mathbb{L}' = \{L_i\}$, we say that $\topicnetwork(L_1, \mathbb{L}')$ \textit{is a two-level topic network} that is generated by the \textit{generating layers} $L_1$ and $L_i$.
If $\mathbb{L}' = \emptyset$, then Formula $\ref{math:two-layer topic network:vertices}$ changes to Formula \ref{math:Cotextual Topic Network:2} and Formula \ref{math:two-layer topic network:arcs} to Formula \ref{math:Cotextual Topic Network:4}.
By omitting any optional classifier $g\in \{\rclassify, \roclassify\}$, expressions of the sort $\XStackY{r}{\dot{v}}{f}{g}$ change to $\Stack{r}{\dot{v}}{f}$.
$\oclassify$ is treated analogously.
\hfill $\Box$
\end{definition}

To understand Formula \ref{math:two-layer topic network:vertices} look at Figure \ref{fig:Diagrammatic Depiction of Inferring Topic Networks}:
among other things, Formula \ref{math:two-layer topic network:vertices} collects the triangle spanned by $v$, $x$ and $a$ supposed that the {two-level topic network} is based on text and authorship links.
Obviously, Definition \ref{def:Two-layer Topic Network} generalizes Definition \ref{def:Text Topic Network}.
Now it should be clear why we speak of the text network of an LMN as its pivotal level: 
it is the reference layer of any additional layer that is integrated into a two-level topic network according to Definition \ref{def:Two-layer Topic Network}.
This role is maintained below when we generalize this definition to capture $n$ layers, $n > 2$.
With the help of Definition \ref{def:Two-layer Topic Network}, we can immediately derive so-called \textit{author topic networks}:

\begin{definition}\label{def:Author Topic Network}
An \textbf{\textit{Author Topic Network}} (ATN) is a directed graph 
\begin{equation*}\label{math:Coauthorial Topic Network}
\topicnetwork(L_1, \mathbb{L}') = 
(\Prime{V}{}, \Prime{A}{}, \Prime{\mu}{}, \Prime{\nu}{}, \Prime{\lambda}{}, \Prime{\kappa}{})
\end{equation*}
according to Definition \ref{def:Two-layer Topic Network} such that $\mathbb{L}' = \{L_2\}$.
\hfill $\Box$\end{definition}

The relational arguments of this definition can be motivated as follows -- assuming that they are instantiated appropriately:

\begin{enumerate}
\item $\StackX{x}{\dot{v}}{\classify}$ can be used to represent evidence that text $x$ is about topic $\dot{v}$ possibly in relation to other topics of $V_{\classification}$.
\item $\StackX{\dot{v}}{x}{\rclassify}$ can be used to represent evidence that text $x$ is a prototypical instance of topic $\dot{v}$ possibly in relation to other texts in $\Prime{V}{1}$. 
\item $\StackX{r}{\dot{v}}{\oclassify}$ can be used to represent the extent to which agent $r$ tends to write about topic $\dot{v}$ possibly in relation to other topics of $V_{\classification}$. 
\item $\StackX{\dot{v}}{r}{\roclassify}$ represents evidence that agent $r$ is a prototypical author writing about topic $\dot{v}$ possibly in relation to other agents in $\Prime{V}{2}$. 
\item For $x\not= y$, $\StackX{x}{y}{\Prime{\nu}{1}}$ can be calculated to represent evidence about text $x$ to be intertextually linked to text $y$ (e.g.\ in the sense of linking contributions of different authors).
Otherwise, if $x=y$, $\Stack{x}{y}{\Prime{\nu}{1}}$ can be used to quantify evidence about $x$ being intratextually structured.
\item $\StackX{r}{x}{\Prime{\nu}{2\myto 1}}$ can be used to quantify evidence about the role of agent $r$ as an author of text $x$ possibly in relation to other texts authored by $r$. 
Typically, $\Prime{\nu}{2\myto 1}$ is a function of the number of edit actions performed by $r$ on $x$ \cite{Brandes:Kenis:Lerner:vanRaaij:2009}.
\item $\StackX{x}{r}{\Prime{\nu}{1\myto 2}}$ can be used to quantify evidence about the role of agent $r$ as a prototypical author of text $x$ possibly in relation to other authors of $x$. 
In the simplest case, $\Prime{\nu}{2\myto 1}$ is symmetric making $\Prime{\nu}{1\myto 2}$ obsolete.
\item $\StackX{r}{s}{\Prime{\nu}{2}}$ represents evidence that agent $r$ is a coauthor of or interacting with $s$. 
For instantiating $\Prime{\nu}{2}$, the literature knows a wide range of alternatives \cite{Newman:2004:a,Brandes:Kenis:Lerner:vanRaaij:2009} (which mostly concern symmetric measures of co-authorship).
Note that we do not require that $r\not= s$.
\end{enumerate}

\begin{figure}[t]
	\includegraphics[width=0.65\textwidth]{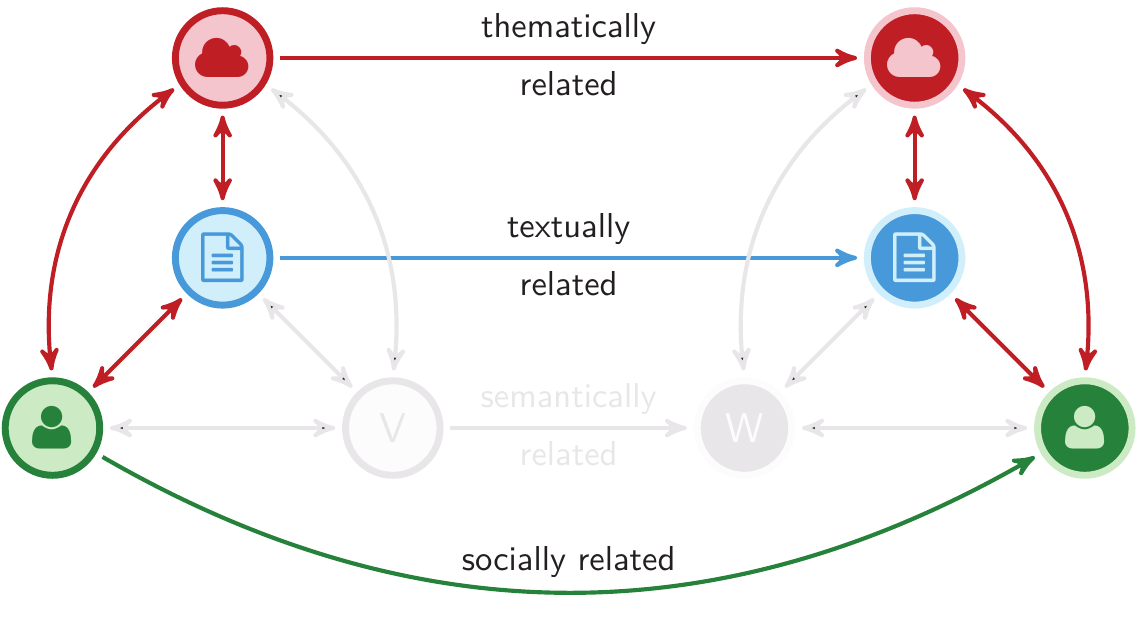}
	\caption{Schematic depiction of the informational sources of linking topics (red vertices) in \textit{author topic networks} as a function of the textual relatedness of two texts (blue vertices) (that belong to layer $L_1$ of a corresponding LMN -- see Definition \ref{def:Linguistic Multilayer Networks}) and the social relatedness of corresponding authors (green vertices) (that belong to layer $L_2$ of a corresponding LMN).
	Bidirectional red arcs denote arcs of the corresponding margin layers in Definition \ref{def:Linguistic Multilayer Networks}.
	}
	\label{fig:ATN:ground:scheme}
\end{figure}

\begin{exmp}\label{ex:ATNs}
Starting from Example \ref{ex:TTNs} to exemplify arcs between topics in \textit{author topic networks}, we can now additionally explore the evidence, that text $x_1$ and $x_2$ are both co-authored by the agents $a_1, a_2$. 
That is, we can assume a co-authorship link $(a_1,a_2)\in A_2$ ($A_2$ is the arc set of the author layer in Definition \ref{def:Linguistic Multilayer Networks}) of weight $\nu(a_1,a_2) = 1$.
Let us now assume the following simplification of the function $\delta$ in Definition \ref{def:Two-layer Topic Network}, for which we assume that it simply multiplies and adds up its argument values in the following way:
\begin{eqnarray*}
\delta[
\StackXY{x}{\dot{v}}{\classify}{\rclassify},
\StackXY{y}{\dot{w}}{\classify}{\rclassify},
\StackXY{r}{\dot{v}}{\vartheta}{\roclassify},
\StackXY{s}{\dot{w}}{\vartheta}{\roclassify},
\StackXY{r}{x}{\Prime{\nu}{2\myto 1}}{\Prime{\nu}{1\myto 2}}, 
\StackXY{s}{y}{\Prime{\nu}{2\myto 1}}{\Prime{\nu}{1\myto 2}}, 
\StackX{r}{s}{\Prime{\nu}{2}},
\StackX{x}{y}{\Prime{\nu}{1}}
]
& \gets & \\
\delta[
\StackX{x}{\dot{v}}{\classify},
\StackX{y}{\dot{w}}{\classify},
\StackX{r}{x}{\Prime{\nu}{2\myto 1}}, 
\StackX{s}{y}{\Prime{\nu}{2\myto 1}}, 
\StackX{r}{s}{\Prime{\nu}{2}},
\StackX{x}{y}{\Prime{\nu}{1}}
] 
& \gets & \\
(\StackX{x}{\dot{v}}{\classify}) \cdot 
(\StackX{y}{\dot{w}}{\classify}) \cdot
(\StackX{r}{x}{\Prime{\nu}{2\myto 1}}) \cdot
(\StackX{s}{y}{\Prime{\nu}{2\myto 1}}) \cdot
(\StackX{r}{s}{\Prime{\nu}{2}} + \StackX{x}{y}{\Prime{\nu}{1}}) & = & (1\cdot 1 \cdot 1\cdot 1) (1+ 1) \\ & = & 2
\end{eqnarray*}
In our example, we get $\dot{v} = t_1 = \lambda(v_1)$, $\dot{w} = t_2 = \lambda(v_2)$, $x= x_1$, $y= x_2$, $r=a_1$ and $s=a_2$.
Since there is no other interlinked pair of texts (see Example \ref{ex:LMN}), instantiating the topics $v_1, v_2$, we get $\nu((v_1,v_2)) = 2$ as the weight of this topic link in the corresponding ATN.
By this simplified example of an ATN, we get the information that the link of topic $v_1$ to topic $v_2$ is additionally supported by the co-authorship of agents $a_1,a_2$: 
this information extends the evidence about the topic link as provided by the underlying TTN of Example \ref{ex:TTNs}.
Likewise, the reflexive link of topic $v_3$ is augmented by 1 compared to the underlying TTN, while there is no other topic link to be considered in this example of an ATN.
By analogy to Figure \ref{fig:TTN:ground:scheme}, Figure \ref{fig:ATN:ground:scheme} gives a schematic depiction of this scenario.
Note that in our example, the weight of the link between authors $a_1,a_2$ (cf.\ $\StackX{r}{s}{\Prime{\nu}{2}}$) is a function of their co-authorship: this is only one alternative to weight the social relatedness of both agents, actually one that can be measured by exploring (special) wikis. 
However, any other social relatedness might be explored to weight the interaction of agents.
\end{exmp}

By comparing a text topic network $\topicnetwork(L_1) = (\Prime{V}{l+1}, \Prime{A}{l+1}, \Prime{\mu}{l+1}, \Prime{\nu}{l+1}, \Prime{\lambda}{l+1}, \Prime{\kappa}{l+1})$ with an author topic network $\topicnetwork(L_1, \{L_{2}\}) = (\Prime{V}{l+2}, \Prime{A}{l+2}, \Prime{\mu}{l+2}, \Prime{\nu}{l+2}, \Prime{\lambda}{l+2}, \Prime{\kappa}{l+2})$ derived from the same LMN $\mathcal{L}(X,l)$, we can learn how the topics of $V_{\classification}$ are manifested in the texts of corpus $X$ in the form of a concomitance or a disparity of intertextual and co-authorship-based networking.
Consider, for example, two vertices $v\in \Prime{V}{l+1}, w\in \Prime{V}{l+2}$ such that $\dot{v}=\dot{w}$; let further $\bot$ and $\top$ denote the minimum and maximum that the vertex weighting functions of both graphs can assume.
Then we can distinguish four extremal cases:

\begin{enumerate}
\item Cases of the sort
\begin{equation}\label{math:case:mu:1}
\bot \ll \Prime{\mu}{l+1}(v) \approx \Prime{\mu}{l+2}(w) \approx \top
\end{equation} 
provide information on prominent topics that tend to be addressed by many texts which are coauthored by many authors.
\item Situations like
\begin{equation}\label{math:case:mu:2}
\top \gg \Prime{\mu}{l+1}(v) \approx \Prime{\mu}{l+2}(w) \approx \bot
\end{equation}
probably apply to the majority of the topics in $V_{\classification}$, which are hardly or even not at all addressed by texts in $\Prime{V}{1} = X$ due to the narrow thematic focus of these texts.
\item Cases like
\begin{equation}\label{math:case:mu:3}
\top \approx \Prime{\mu}{l+1}(v) \gg \Prime{\mu}{l+2}(w) \approx \bot
\end{equation} 
suggests a Zipfian topic effect, according to which a prominent topic is addressed by a small group of agents or even by a single author.
\item Finally, situations of the sort
\begin{equation}\label{math:case:mu:4}
\bot \approx \Prime{\mu}{l+1}(v) \ll \Prime{\mu}{l+2}(w) \approx \top
\end{equation} 
refer to rarely manifested topics addressed by a few but highly coauthored texts.
In conjunction with many cases of the sort described by Formula \ref{math:case:mu:3}, situations of this kind indicate a Zipfian coauthoring effect, according to which many authors write only a few texts, while many texts are written by a few authors without encountering many (relevant) coauthors.
\end{enumerate}

Formulas \ref{math:case:mu:1}--\ref{math:case:mu:4} compare the node weighting functions of a TTN with those of a related ATN.
The same can be done regarding their arc weighting functions. 
That is, for two arcs $a = (r,s) \in \Prime{A}{l+1}$ and $b = (v,w) \in \Prime{A}{l+2}$, for which $\dot{r}=\dot{v} \wedge\dot{s}=\dot{w}$, we distinguish again four cases ($\bot$ and $\top$ now denote the minimum and maximum the arc weighting functions of both graphs can assume):
\begin{enumerate}
\item In the case of
\begin{equation}\label{math:case:nu:1}
\bot \ll \Prime{\nu}{l+1}(a) \approx \Prime{\nu}{l+2}(b) \approx \top
\end{equation} 
topic $\dot{v}$ is intertextually linked more strongly to topic $\dot{w}$ and authors of its text instances tend to cooperate with those of instances of topic $\dot{w}$ likewise to a greater extent.
\item In the case of
\begin{equation}\label{math:case:nu:2}
\top \gg \Prime{\nu}{l+1}(a) \approx \Prime{\nu}{l+2}(b) \approx \bot
\end{equation} 
topic $\dot{v}$ is intertextually less strongly linked to topic $\dot{w}$ and the few authors of its textual instances tend to cooperate with authors of instances of topic $\dot{w}$ likewise to a lesser extent.
\item In the case of
\begin{equation}\label{math:case:nu:3}
\top \approx \Prime{\nu}{l+1}(a) \gg \Prime{\nu}{l+2}(b) \approx \bot
\end{equation} 
topic $\dot{v}$ is intertextually more strongly connected with topic $\dot{w}$, while authors of its text instances tend to cooperate with those of instances of topic $\dot{w}$ to a lesser extent, if at all.
\item Finally, in the case of 
\begin{equation}\label{math:case:nu:4}
\bot \approx \Prime{\nu}{l+1}(a) \ll \Prime{\nu}{l+2}(b) \approx \top
\end{equation} 
topic $\dot{v}$ is intertextually less strongly linked to topic $\dot{w}$, while the numerous authors of its text instances tend to cooperate with those of instances of topic $\dot{w}$ to a much greater extent.
\end{enumerate}

Our central question regarding the relationship between TTNs and ATNs \textit{derived from the same LMN} is whether these networks are similar or not. 
If they are similar, we expect that cases of the sort described by formulas \ref{math:case:mu:1}, \ref{math:case:mu:2}, \ref{math:case:nu:1}, \ref{math:case:nu:2} predominate so that cases matched by Formula \ref{math:case:mu:1} are parallelized by those considered by Formula \ref{math:case:nu:1} and where cases according to Formula \ref{math:case:mu:2} are concurrent to those described by Formula \ref{math:case:nu:2}.
An opposite situation would be that two topic nodes in the TTN are highly weighted but weakly linked, while they are weakly weighted but strongly linked in the corresponding ATN.
In this case, a few or even only a single author is responsible for the thematic focus of the TTN. 
Note that this scenario reminds again of a Zipfian effect regarding the relation of TTNs and ATNs.
By characterizing TTNs in relation to ATNs along these and related scenarios, we want to investigate laws of the interdependence of both types of networks, which may consist, for example, in the simultaneity of dense or sparse intertextuality-based networking on the one hand and dense or sparse co-authorship-based networking on the other.
We may expect, for example, that the more related two topics, the more likely the authors of their textual instances cooperate. 
However, not so much is known about such scenarios in the area of VGI especially with regard to Hypothesis \ref{hypo:Hypothesis 1}.
Thus, we address this gap -- at least by introducing a novel theoretical model which may help filling it.

Figure \ref{fig:TTN and ATNs of Muenchen City Wiki} exemplifies two ATNs in relation to a corresponding TTN (T1) which were computed using the apparatus of Section \ref{sec:A Procedural Model of Topic Network Analysis} to instantiate the formal model of this Section. 
The upper right ATN (A1) is computed by globally weighting co-authorship activities based on Wikipedia (as explained in Section \ref{sec:Module 3: Network Induction}); 
the ATN (A2) below is calculated by weighting of these activities relative to the city wiki itself.
Figure \ref{fig:TTN and ATNs of Muenchen City Wiki} shows that the topic with DDC number 720 (\textit{Architecture}) is weighted higher in A1 than in T1. 
This is all the more pronounced in A2, where 720 becomes the most prominent topic and consequently displaces the top subject from T1, that is, topic 380 (\textit{Commerce, communications \& transportation}).
That is, although topic 380 is most frequently addressed in this wiki's texts, topic 720 is not only almost as salient, but also attracts many more activities among its interacting coauthors.
Similar observations concern the switch of the roles of the topics 910 (\textit{Geography \& travel}) and 940 (\textit{History of Europe}) from T1 to A1 and A2.

Regardless of the answer to this and related questions, we will also ask whether the shape of an ATN can be predicted if one knows the shape of the corresponding TTN and vice versa.
To answer this question, we will consider LMNs of different text genres: 
of city wikis and regional wikis on the one hand and extracts of encyclopedic wikis on the other.
We expect that LMNs spanned over corpora of the same genre exhibit a pattern of collaboration- and intertextuality-based networking that makes TTNs and ATNs derived from them mutually recognizable or predictable, whereas for LMNs generated from corpora of different genres this does not apply.

For reasons of formal variety we now consider an alternative to author topic networks, namely so-called word topic networks, which in turn are derived from Definition \ref{def:Two-layer Topic Network}:

\begin{definition}\label{def:Word Topic Network}
A \textbf{Word Topic Network} (WTN) is a directed graph 
\begin{equation*}\label{math:Word Topic Network}
\topicnetwork(L_1, \mathbb{L}') = (\Prime{V}{}, \Prime{A}{}, \Prime{\mu}{}, \Prime{\nu}{}, \Prime{\lambda}{}, \Prime{\kappa}{})
\end{equation*}
according to Definition \ref{def:Two-layer Topic Network} such that $\mathbb{L}' = \{L_3\}$.
\hfill $\Box$\end{definition}

This definition departs by five new relational arguments from Definition \ref{def:Author Topic Network}, which -- if being instantiated appropriately -- can be motivated as follows:

\begin{enumerate}
	\item $\StackX{a}{x}{\Prime{\nu}{3\myto 1}}$ quantifies evidence about the role of word $a$ as a lexical constituent of text $x$ possibly in relation to all other texts in which $a$ occurs. 
	Typically, $\Prime{\nu}{3\myto 1}$ is implemented by a global term weighting function 	\cite{Salton:Buckley:1988} or by a neural network-based feature selection function. 
	\item $\StackX{x}{a}{\Prime{\nu}{1\myto 3}}$ quantifies evidence about the role of the word $a$ as a lexical constituent of the text $x$ possibly in relation to other lexical constituents of $x$. 
	Typically, $\Prime{\nu}{1\myto 3}$ is a local term weighting function, such as normalized term frequency \cite{Salton:Buckley:1988}, or a topic model-based function. 
	\item $\StackX{a}{\dot{v}}{\oclassify}$ represents evidence about the word $a$ to be associated with the topic $\dot{v}$ possibly in relation to all other topics of $V_{\classification}$.
	\item $\StackX{\dot{v}}{a}{\roclassify}$ calculates evidence about the extent to which the topic $\dot{v}$ is prototypically labeled by the word $a$, possibly in relation to all other words in $\Prime{V}{3}$. 
	\item $\StackX{a}{b}{\Prime{\nu}{3}}$ quantifies evidence about the extent to which the word $a$ associates the word $b$.
	Typically, $\Prime{\nu}{3}$ is computed by means of word embeddings \cite{Mikolov:Yih:Zweig:2013}.
\end{enumerate}

Based on this list we better understand what topic networks offer in contrast to TMs.
This concerns the flexibility with which we can include informational resources computed by different methods (e.g.\ based on neural networks, topic models, LSA, etc.) to generate topic networks (cf.\ challenge \ref{enumerate:R7} on page \pageref{enumerate:R7}).
Different relational arguments $\Stack{X}{Y}{Z}$ can be quantified using different methods, which in turn can belong to a wide range of computational paradigms.
Table \ref{tab:genericity of the model} gives an account of the generality of our approach by hinting at candidate procedures for computing the different relations of Figure \ref{fig:Diagrammatic Depiction of Inferring Topic Networks}.

\begin{table}[t]
	\resizebox{0.95\textwidth}{!}{
		\begin{tikzpicture}[
 C1/.style={minimum height=0.8cm, text width=1.50cm, align=right, fill=SeminarSehrSehrHellGrau, anchor=north},
 C2/.style={rectangle, fill=SeminarSehrSehrHellGrau, text=black, minimum height=0.8cm, text width=9.00cm, align=left, anchor=north},
 ]
 
 \matrix [matrix of nodes, row sep=0.1em, column sep=0.1em,
 column 1/.style={nodes={C1}},
 column 2/.style={nodes={C1,align=center,text width=1.50cm}},
 column 3/.style={nodes={C1,align=left}},
 column 4/.style={nodes={C2}},
 row 1/.append style={nodes={fill=SeminarSehrHellBlau}},
 %
 ] (A) {
 \textbf{Source\strut} & \textbf{Relation\strut} & \textbf{Target\strut} & \textbf{Candidate Procedure\strut} \\
 $\text{text}$ & $\stackrel{\classify}{\to}$ & $\text{topic}$ & text2ddc \cite{Uslu:Mehler:Niekler:Baumartz:2018} \\
 $\text{topic}$ & $\stackrel{\rclassify}{\to}$ & $\text{text}$ & {text2ddc$^{-1}$}\\
 $\text{text}$ & $\stackrel{\Prime{\nu}{1}}{\to}$ & $\text{text}$ & measures of sentence/text similarity, text embeddings \cite{Harispe:et:al:2015} \\
 $\text{agent}$ & $\stackrel{\oclassify}{\to}$ & $\text{topic}$ & {topic models \cite{Steyvers:Griffiths:2007}}\\
 $\text{topic}$ & $\stackrel{\roclassify}{\to}$ & $\text{agent}$ & {topic models \cite{Steyvers:Griffiths:2007}}\\
 $\text{agent}$ & $\stackrel{\Prime{\nu}{2\myto 1}}{\to}$ & $\text{text}$ & {edit networks \cite{Brandes:Kenis:Lerner:vanRaaij:2009}}\\
 $\text{text}$ & $\stackrel{\Prime{\nu}{1\myto 2}}{\to}$ & $\text{agent}$ & {edit networks \cite{Brandes:Kenis:Lerner:vanRaaij:2009}}\\
 $\text{agent}$ & $\stackrel{\Prime{\nu}{2}}{\to}$ & $\text{agent}$ & {co-authorship \cite{Brandes:Kenis:Lerner:vanRaaij:2009,Newman:2004:b}}\\
 $\text{word}$ & $\stackrel{\oclassify}{\to}$ & $\text{topic}$ & {text2ddc \cite{Uslu:Mehler:Niekler:Baumartz:2018}, topic models \cite{Steyvers:Griffiths:2007}}\\
 $\text{topic}$ & $\stackrel{\roclassify}{\to}$ & $\text{word}$ & {text2ddc$^{-1}$, topic models \cite{Steyvers:Griffiths:2007}}\\
 $\text{word}$ & $\stackrel{\Prime{\nu}{3\myto 1}}{\to}$ & $\text{text}$ & {fastText, topic models \cite{Steyvers:Griffiths:2007}}\\
 $\text{text}$ & $\stackrel{\Prime{\nu}{1\myto 3}}{\to}$ & $\text{word}$ & {fastText, topic models \cite{Steyvers:Griffiths:2007}}\\
 $\text{word}$ & $\stackrel{\Prime{\nu}{3}}{\to}$ & $\text{word}$ & {word embeddings \cite{Mikolov:Yih:Zweig:2013,Levy:Goldberg:2014,Ling:Dyer:Black:Trancoso:2015,Komninos:Manandhar:2016}}\\
 $\ldots$ & $\ldots$ & $\ldots$ & $\ldots$\\
 };
 \end{tikzpicture}
 
	}
	\caption{Building blocks of topic networks (texts, topics, words, agents etc.), their relations according to Figure \ref{fig:Diagrammatic Depiction of Inferring Topic Networks} and candidate procedures for weighting the corresponding arcs (last column).}
	\label{tab:genericity of the model}
\end{table}

\begin{figure}[t]
	\includegraphics[width=0.65\textwidth]{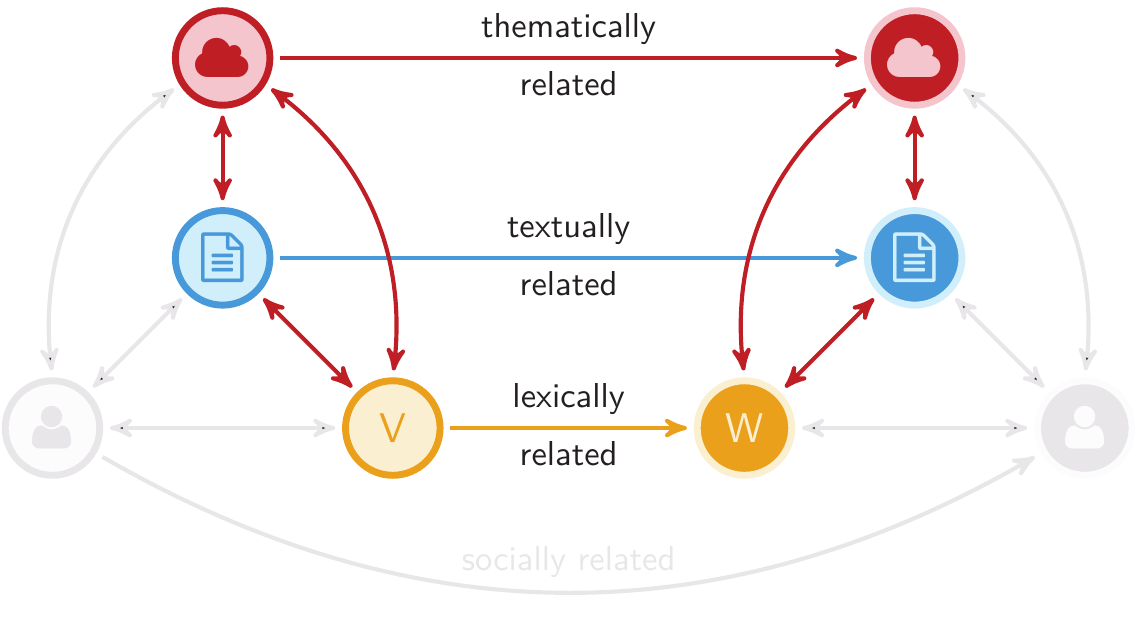}
	\caption{Schematic depiction of the informational sources of linking topics (red vertices) in \textit{word topic networks} as a function of the textual relatedness of two texts (blue vertices) (that belong to layer $L_1$ of a corresponding LMN -- see Definition \ref{def:Linguistic Multilayer Networks}) and the lexical relatedness of corresponding words (orange vertices) (that belong to layer $L_3$ of a corresponding LMN).
	Bidirectional red arcs denote arcs of the corresponding margin layers in Definition \ref{def:Linguistic Multilayer Networks}.
	}
	\label{fig:WTN:ground:scheme}
\end{figure}

\begin{exmp}\label{ex:WTNs}
Starting from Example \ref{ex:TTNs} to exemplify arcs between topics in \textit{word topic networks}, we have to additionally explore evidence regarding the lexical relatedness of the vocabularies of the texts $x_1$ and $x_2$. 
In Example \ref{ex:LMN}, we assumed that the intersection of both texts (represented as bags-of-words) is given by the set $\{w_1,w_2\}$.
By analogy to Example \ref{ex:ATNs}, we assume now the following simplification of the function $\delta$ of Definition \ref{def:Two-layer Topic Network}:
\begin{eqnarray*}
	\delta[
	\StackXY{x}{\dot{v}}{\classify}{\rclassify},
	\StackXY{y}{\dot{w}}{\classify}{\rclassify},
	\StackXY{r}{\dot{v}}{\vartheta}{\roclassify},
	\StackXY{s}{\dot{w}}{\vartheta}{\roclassify},
	\StackXY{r}{x}{\Prime{\nu}{2\myto 1}}{\Prime{\nu}{1\myto 2}}, 
	\StackXY{s}{y}{\Prime{\nu}{2\myto 1}}{\Prime{\nu}{1\myto 2}}, 
	\StackX{r}{s}{\Prime{\nu}{2}},
	\StackX{x}{y}{\Prime{\nu}{1}}
	]
	& \gets & \\
	(\StackX{x}{\dot{v}}{\classify}) \cdot 
	(\StackX{y}{\dot{w}}{\classify}) \cdot
	(\StackX{r}{x}{\Prime{\nu}{2\myto 1}}) \cdot
	(\StackX{s}{y}{\Prime{\nu}{2\myto 1}}) \cdot
	(\StackX{r}{s}{\Prime{\nu}{2}} + \StackX{x}{y}{\Prime{\nu}{1}}) 
\end{eqnarray*}
In this scenario, we have to instantiate Definition \ref{def:Two-layer Topic Network} as follows: 
$\dot{v} \gets t_1 = \lambda(v_1)$, $\dot{w} \gets t_2 = \lambda(v_2)$, $x= x_1$, $y= x_2$, $r=w_1$ and $s=w_1$ for one summand and -- everything else being constant -- $r= w_2$ and $s=w_2$ for a second summand (for $w_3$ ($w_4$) we do not assume a lexical relatedness w.r.t.\ the words of text $w_4$ ($w_3$)).
Note that under this regime, we assume that \textit{relatedness of lexical constituents} only concerns shared usages of identical words -- of course, this is a simplifying example.
By analogy to the setting of Example \ref{ex:ATNs} we have thus to conclude that $\nu((v_1,v_2)) = 4$ as the weight of the topic link from $v_1$ to $v_2$ in the corresponding WTN.
For texts $x_3, x_4$ we may alternatively assume that lexical relatedness does not only concern shared lexical items but also relatedness that is measured, for example, by means of a terminological ontology \cite{Budanitsky:Hirst:2006} or by means of word embeddings \cite{Mikolov:Yih:Zweig:2013}.
In this way, we may additionally arrive at a topic link between $v_2$ and $v_3$.
In order to allow for a comparison of a WTN with its corresponding TTN, a more realistic weighting scheme is needed that also reflects above and below average lexical relatednesses of the lexical constituents of interlinked texts -- in Section \ref{sec:A Procedural Model of Topic Network Analysis} we elaborate such a model regarding ATNs in relation to TTNs.
Figure \ref{fig:WTN:ground:scheme} gives a schematic depiction of the scenario of WTNs as elaborated so far.
\end{exmp}

It is worth emphasizing that instead of the (language-systematic) lexicon layer $L_3$, we may use a constituent layer $L_k, k > 3$, to infer a two-level topic network.
For example, we can use the layer spanned by the sentences of the pivotal texts to obtain a sort of \textit{sentence topic network}.
In this case, $\Stack{a}{b}{\Prime{\nu}{k}}$ may quantify evidence about the extent to which the sentence $a$ entails the sentence $b$ or the extent to which the sentence $a$ is similar to the sentence $b$ etc., while $\Stack{x}{a}{\Prime{\nu}{1\myto k}}$ may quantify evidence about the extent to which the sentence $a$ is thematically central for the text $x$ etc.
In sentence topic networks, topic linkage is a function of sentence linkage: 
prominent topics emerge from being addressed by many sentences, while prominent topic links arise from the relatedness of many underlying sentences.
Another example of inferring two-level topic networks is to link topics as a function of places mentioned (by means of toponyms) within the texts of the underlying corpus $X$ where geospatial relations of these places can be explored to infer concurrent topic relations: 
if place $p$ is mentioned in text $x$ about topic $\dot{v}$ and place $q$ in text $y$ about topic $\dot{w}$, where the platial relation $R(p, q)$ relates $p$ and $q$, this information can be used to link the topic nodes $v, w$ in the corresponding topic network.
As a result, we obtain networks manifesting the networking of topics as a function of parallelized geographical relations.

Obviously, any other relationship (e.g., entailment among sentences, sentiment polarities shared by linked texts, co-reference relations etc.) can be investigated to induce such two-level networks.
And even more, we can think of $n$-level networks in which several such relationships are explored at once to generate topic links.
We can ask, for example, which locations are linked by which geospatial relations while being addressed in which sentences about which topics where these sentences are related by which sentiment relations.
Another example is to ask which authors prefer to write about which topics while tending to use which vocabulary:
the higher the number of authors who use the same words more often to write about the same topic, and the higher the number of such words, the higher the weight of that topic.
In this case, topic weighting is a function of frequently observed pairs of linguistic (here: lexical) means \textit{and} authors.
On the other hand, the higher the degree of co-authorship of two authors contributing to different topics and the higher the degree of association of the words used by these authors to write about these topics, the higher the weight of the link between the topics.
This concept of a topic network induced by the text, the co-authorship and the lexicon layer of an LMN is addressed by the following generalization, which provides a generation scheme for topic networks:
\begin{figure}[t]
	\includegraphics[width=0.90\textwidth]{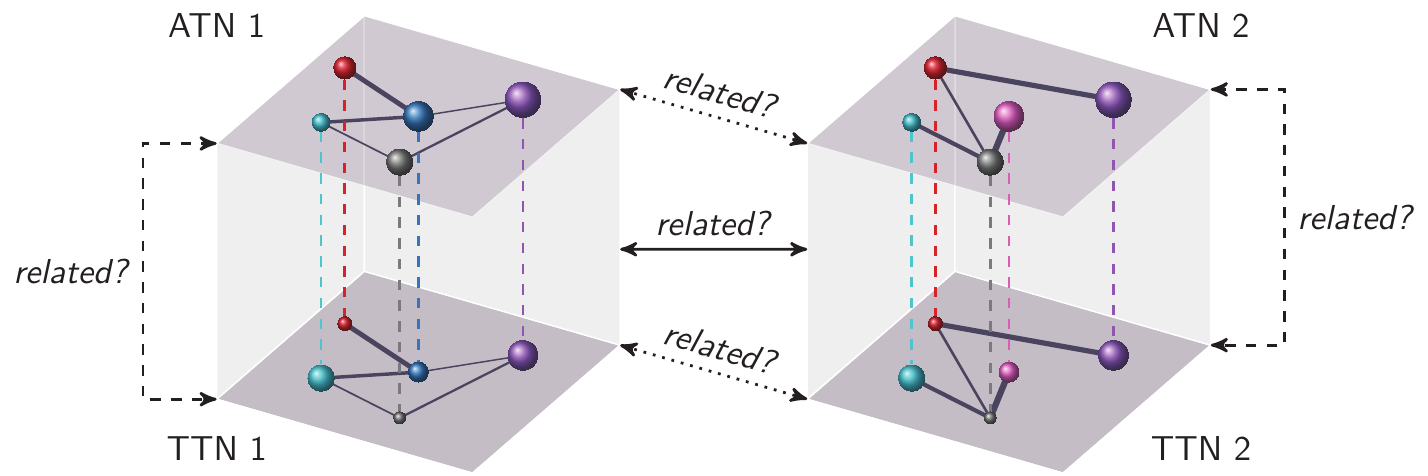}
	\caption{3D depiction of two MTNs (left and right) each consisting of two layers (including a TTN at the bottom and an ATN at the top of the respective cube). Shared colors of nodes and dashed vertical lines indicate identically labeled vertices. The depiction disregards the orientation of the arcs. In this example, all four layers span topic networks over the same set of topics (vertices). Any such two-layer MTN can be used to represent the intertextuality- and co-authorship-based networking of the topics derived from the same corpus of texts about the same place. In this way, we gain several perspectives for the analysis of such multiplex networks: by comparing the TTNs or the ATNs of different MTNs (dotted arcs), by comparing the TTNs of different networks with their corresponding ATNs (dashed arcs) or by comparing the different MTNs as a whole with each other (solid arc).}
	\label{fig:Comparing two Two-level Topic Networks}
\end{figure}
\begin{definition}\label{def:n-Layer Topic Network}
Given a definitional setting $\mathcal{S} = (\classification, \classify, \mathcal{L}(X,l))$ according to Definition \ref{def:Definitional Setting}, an \textbf{$(L_1, \mathbb{L}')$-\textit{Topic Network}}, for which 
\begin{equation}
\mathbb{L}' = \{L_{i_1}, \ldots, L_{i_n}\} \in 2^{\{L_2, \ldots, L_l\}},
\end{equation}
is a vertex- and arc-weighted simple directed graph 
\begin{equation}\label{math:Jointly induced Topic Network}
\topicnetwork(L_1, \mathbb{L}') = (\Prime{V}{}, \Prime{A}{}, \Prime{\mu}{}, \Prime{\nu}{}, \Prime{\lambda}{}, \Prime{\kappa}{})
\end{equation}
which is said to be \textit{derived from} $\mathcal{S}$ and \textit{inferred from} $L_1$ and the elements of $\mathbb{L}'$ by means of the optional classifiers 
$\rclassify, 
\forall i_j\in \{{i_1}, \ldots, {i_n}\}\!: 
\oclassify_{i_j}\!: \Prime{V}{i_j}\times V_{\classification}\to \mathbb{R}_0^+, 
\vartheta_{i_j}^{\shortgets}\!: V_{\classification} \times \Prime{V}{i_j} \to \mathbb{R}_0^+$ 
and monotonically increasing functions ${\alpha}, {\beta}, {\gamma}, {\delta}\!: \mathbb{R}^+_0\to \mathbb{R}^+_0$ iff $\forall v\in \Prime{V}{}$ and $\forall a = (v,w)\in \Prime{A}{}$:
\begin{alignat}{3}
\Prime{\mu}{}(v) 
&= 
{\alpha}\Bigr(
	\sum_{\substack{x\in \Prime{V}{1},\\ r_{i_1}\in \Prime{V}{i_1}, \ldots, r_{i_n}\in \Prime{V}{i_n}}} 
	&&{\beta}[ 
	 \StackXY{x}{\dot{v}}{\classify}{\rclassify}, \;
	 \StackXY{r_{i_1}}{\dot{v}}{\classify_{i_1}}{\theta_{i_1}^{\shortgets}}, 
	 \ldots, 
	 \StackXY{r_{i_n}}{\dot{v}}{\classify_{i_n}}{\theta_{i_n}^{\shortgets}}, \;
	 \StackXY{r_{i_1}}{x}{\Prime{\nu}{{i_1}\myto 1}}{\Prime{\nu}{1\myto {i_1}}},
	 \ldots,
	 \StackXY{r_{i_n}}{x}{\Prime{\nu}{{i_n}\myto 1}}{\Prime{\nu}{1\myto {i_n}}}
	]
\Bigr) > 0
\label{math:n-level topic networks - mu}\\ 
\Prime{\nu}{}(a) 
&= 
{\gamma}\Bigl( 
	\sum_{
		\substack{
			x,y\in \Prime{V}{1}, \\
			r_{i_1}\in \Prime{V}{i_1}, \ldots, r_{i_n}\in \Prime{V}{i_n}, \\
			s_{i_1}\in \Prime{V}{i_1}, \ldots, s_{i_n}\in \Prime{V}{i_n}
		}
	} 
	&&{\delta}[
		\StackXY{x}{\dot{v}}{\classify}{\rclassify}, 
		\StackXY{y}{\dot{w}}{\classify}{\rclassify}, \\ 
		&&&
		\StackXY{r_{i_1}}{\dot{v}}{\vartheta_{i_1}}{\vartheta_{i_1}^{\shortgets}}, \ldots, \StackXY{r_{i_n}}{\dot{v}}{\vartheta_{i_n}}{\vartheta_{i_n}^{\shortgets}}, 
		\;\; 
		\StackXY{s_{i_1}}{\dot{w}}{\vartheta_{i_1}}{\vartheta_{i_1}^{\shortgets}}, \ldots, 	\StackXY{s_{i_n}}{\dot{w}}{\vartheta_{i_n}}{\vartheta_{i_n}^{\shortgets}}, 
		\nonumber \\
		&&&
		\StackXY{r_{i_1}}{x}{\Prime{\nu}{{i_1}\myto 1}}{\Prime{\nu}{1\myto {i_1}}}, \ldots, \StackXY{r_{i_n}}{x}{\Prime{\nu}{{i_n}\myto 1}}{\Prime{\nu}{1\myto {i_n}}}, 
		\;\; 
		\StackXY{s_{i_1}}{y}{\Prime{\nu}{{i_1}\myto 1}}{\Prime{\nu}{1\myto {i_1}}}, \ldots, \StackXY{s_{i_n}}{y}{\Prime{\nu}{{i_n}\myto 1}}{\Prime{\nu}{1\myto {i_n}}}, 
		\nonumber \\
		&&&
		\StackX{r_{i_1}}{s_{i_1}}{\nu_{i_1}}, 
		\ldots, 
		\StackX{r_{i_n}}{s_{i_n}}{\nu_{i_n}}, 
		\nonumber \\
		&&&
		\StackXY{r_{i_1}}{s_{i_2}}{\Prime{\nu}{i_1\myto i_2}}{\Prime{\nu}{i_2\myto i_1}}
		\ldots, 
		\StackXY{r_{i_1}}{s_{i_n}}{\Prime{\nu}{i_1\myto i_n}}{\Prime{\nu}{i_n\myto i_1}}, 
		\ldots, 
		\StackXY{r_{i_n}}{s_{i_1}}{\Prime{\nu}{i_n\myto i_1}}{\Prime{\nu}{i_1\myto i_n}}
		\ldots, 
		\StackXY{r_{i_n}}{s_{i_{n-1}}}{\Prime{\nu}{i_n\myto i_{n-1}}}{\Prime{\nu}{i_{n-1}\myto i_n}}, 
		\nonumber \\
		&&&
		\StackX{x}{y}{\Prime{\nu}{1}} 
	]
\Bigr) > 0 \label{math:n-level topic networks - nu}
\end{alignat}
$\Prime{\mu}{}\!:\Prime{V}{}\to \mathbb{R^+}$ is a vertex weighting function, $\Prime{\nu}{}\!:\Prime{A}{}\to \mathbb{R^+}$ an arc weighting function, $\Prime{\lambda}{}\!: \Prime{V}{}\to V_{\classification}$ an injective vertex labeling function, $V_{\classification}(V) = \{\lambda(v)\,|\, v\in V\}\subseteq V_{\classification}$, and $\Prime{\kappa}{}$ an injective arc labeling function.
For $|\mathbb{L}'| = n$, we say that $\topicnetwork(L_1, \mathbb{L}')$ \textit{is an $m$-level, $m = n+1$, topic network} generated by the \textit{generating layers} $L_1$ and the elements of $\mathbb{L}'$.
If $\mathbb{L}' = \emptyset$, Formula $\ref{math:n-level topic networks - mu}$ changes to Formula \ref{math:Cotextual Topic Network:2} and Formula \ref{math:n-level topic networks - nu} to Formula \ref{math:Cotextual Topic Network:4}.
By omitting the optional classifier $g\in \{ \vartheta_{i_j}^{\shortgets}\mid j\in \{{1}, \ldots, {n}\} \}$, expressions of the sort $\XStackY{r}{\dot{v}}{f}{g}$ change to $\Stack{r}{\dot{v}}{f}$.
$\classify$ and $\vartheta_{i_j}$ are treated analogously.
In order to derive an \textit{undirected} $m$-level topic network
$\overline{\topicnetwork}(L_1, \mathbb{L}') = (\Prime{V}{}, \Prime{E}{}, \Prime{\mu}{}, \BarPrime{\nu}{}, \Prime{\lambda}{}, \BarPrime{\kappa}{})$ from $\topicnetwork(L_1, \mathbb{L}')$, we define: $\{v,w\}\in \Prime{E}{} \leftrightarrow (v,w)\in \Prime{A}{} \vee (w,v) \in \Prime{A}{}$ and
\begin{equation}
\BarPrime{\nu}{}(\{v,w\}) = 
\begin{cases}
\zeta_1(\Prime{\nu}{}((v,w)), \Prime{\nu}{}((w,v))) & (v,w)\in \Prime{A}{} \wedge (w,v)\in \Prime{A}{} \\
\zeta_2(\Prime{\nu}{}((v,w))) & (v,w)\in \Prime{A}{} \wedge (w,v)\not\in \Prime{A}{} \\
\end{cases}
\end{equation}
and where $\zeta_1, \zeta_2$ are monotonically increasing functions.
\hfill $\Box$\end{definition}

\begin{figure}[t]
	\includegraphics[width=0.65\textwidth]{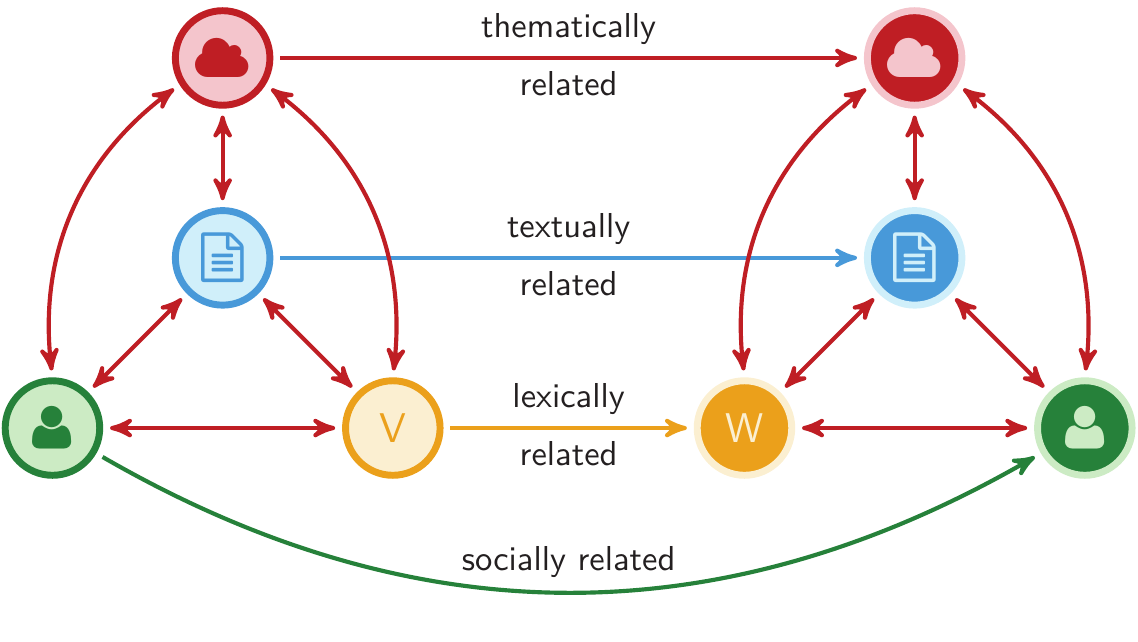}
	\caption{Schematic depiction of informational sources explored to link topics (red vertices) in a \textit{3-level topic network} as a function of the textual relatedness of texts (blue vertices) (belonging to layer $L_1$ of Definition \ref{def:Linguistic Multilayer Networks}), the social relatedness of corresponding authors (green vertices) (belonging to layer $L_2$ of Definition \ref{def:Linguistic Multilayer Networks}) and the lexical relatedness of corresponding words (orange vertices) (belonging to layer $L_3$ of Definition \ref{def:Linguistic Multilayer Networks}).
	In this scenario, thematic relatedness is the information to be inferred, while textual, lexical and social relations concern given information or evidence.
	Bidirectional red arcs denote arcs of corresponding margin layers of Definition \ref{def:Linguistic Multilayer Networks}.
	}
	\label{fig:N-Level-TN:ground:scheme}
\end{figure}

\begin{figure}[t]
	\includegraphics[width=0.85\textwidth]{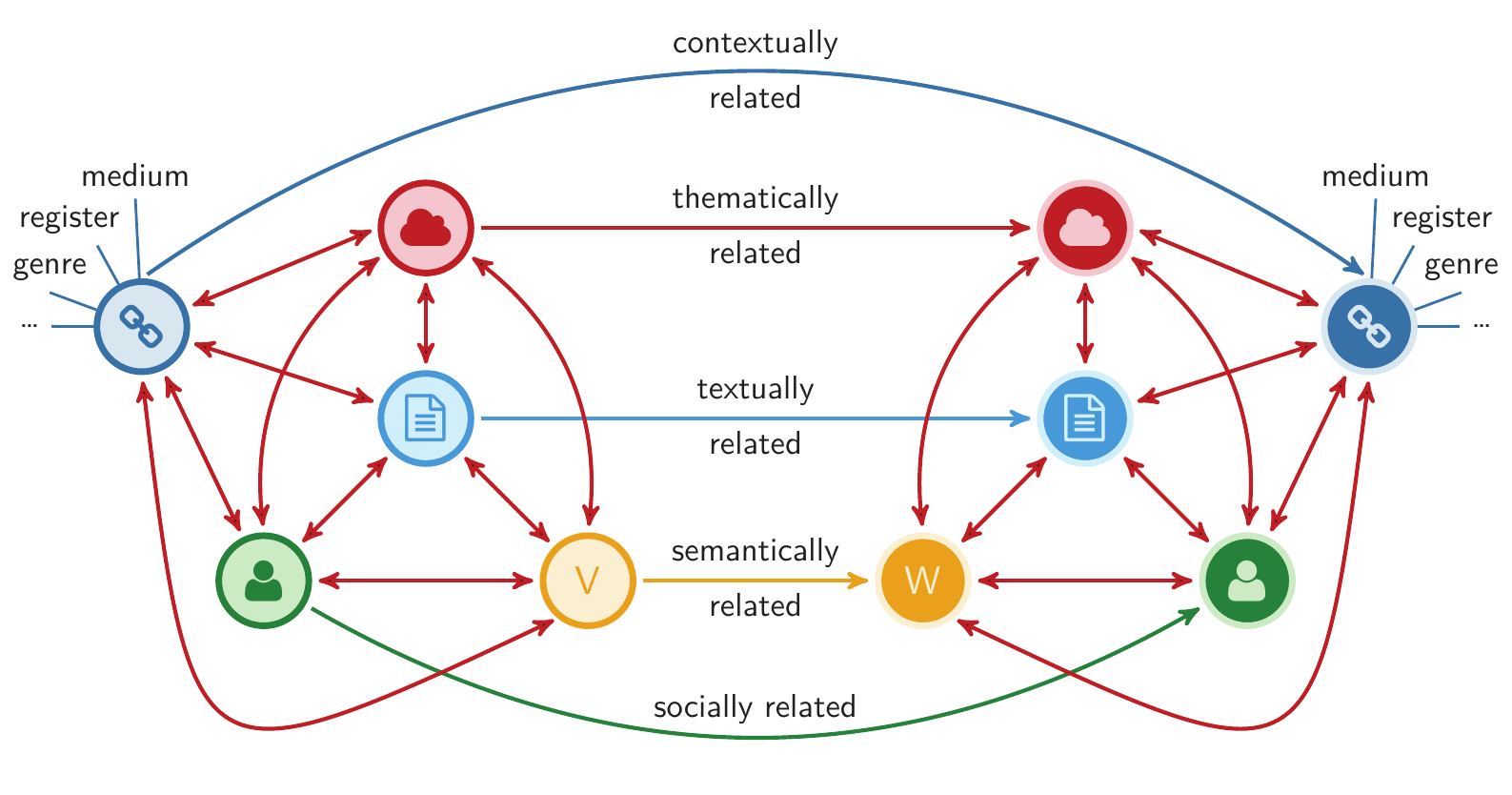}
	\caption{Schematic depiction of informational sources explored to link topics (red vertices) in an \textit{$n$-level topic network}, $n > 3$, as a function of the textual relatedness of texts (blue vertices) (belonging to layer $L_1$ of Definition \ref{def:Linguistic Multilayer Networks}), the social relatedness of corresponding authors (green vertices) (belonging to layer $L_2$ of Definition \ref{def:Linguistic Multilayer Networks}), the lexical relatedness of corresponding words (orange vertices) (belonging to layer $L_3$ of Definition \ref{def:Linguistic Multilayer Networks}) and additional layers of contextual patterns concerning, for example, the underlying medium, genre or register instantiated by the texts under consideration.
	}
	\label{fig:M-Level-TN:ground:scheme}
\end{figure}

Evidently, Definition \ref{def:n-Layer Topic Network} is a generalization of Definition \ref{def:Text Topic Network} by considering higher numbers of generating layers.
A schematic depiction of the scenario addressed by this definition is shown in Figure \ref{fig:N-Level-TN:ground:scheme} by example of a 3-level topic network that explores evidence about topic linking starting from the text, the author and the lexicon layer of Definition \ref{def:Linguistic Multilayer Networks}.
Likewise, Figure \ref{fig:M-Level-TN:ground:scheme} depicts an $n$-level topic network, $n > 3$, in which additional resources are explored beyond the word, author and text level.
Figure \ref{fig:Diagrammatic Depiction of Inferring Topic Networks} illustrates more formally the inference process underlying Definition \ref{def:n-Layer Topic Network}, and in particular of the arguments used.
It illustrates the inference of an arc that connects two topics by exploring the links of the text, author, and lexicon layers of an underlying LMN. 
In this example, the blue and black arcs are evaluated to determine the weights of red arcs connecting the focal topic nodes.
Blue arcs are used to orientate inferred arcs.
We will not develop this apparatus further, nor will we empirically examine $n+1$-layer topic networks for $n > 2$.
Rather, the apparatus developed so far serves to demonstrate the generality, flexibility and extensibility of our formal model.

Above we explained that one of the reasons for introducing a flexible and extensible formalism of topic networks is to compare topic networks derived from different layers (e.g.\ from the text layer on the one hand and the author layer on the other).
In order to systematize this approach, we finally introduce the concept of a \textit{multiplex topic network}, which is derived from the same or from different linguistic multi-layer networks:

\begin{figure}[t]
	\includegraphics[width=0.45\textwidth]{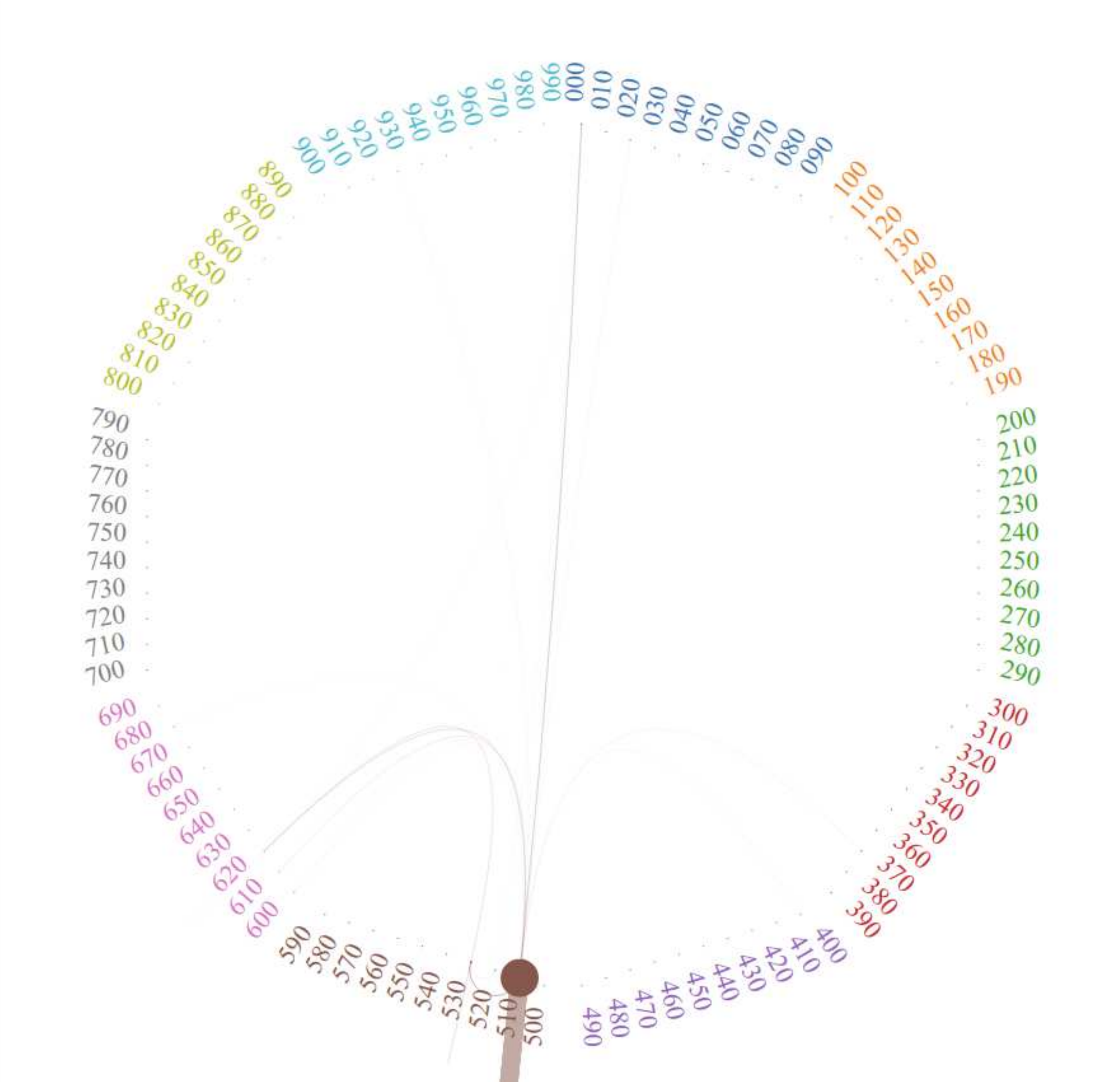} 
	\hfill
	\includegraphics[width=0.45\textwidth]{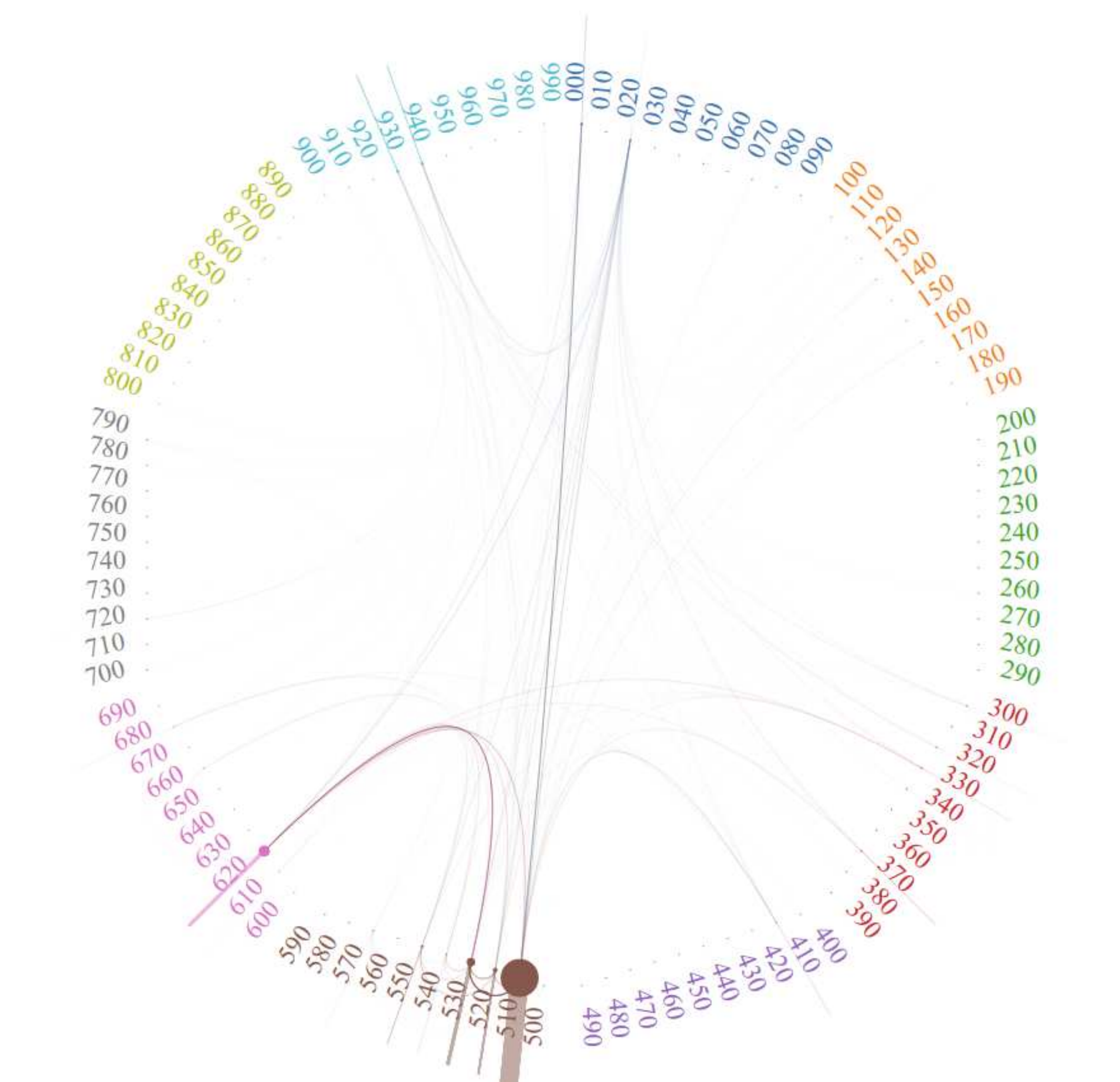}
	\caption{Visualization (by means of PolyViz \cite{Uslu:Mehler:2018}) of the TTN of the \nth{1} orbit (left) and of the \nth{2} orbit (right) of the German Wikipedia article \textit{Integralrechnung} (\textit{Integral}).
	The TTNs are derived from the corpora of articles in the \nth{1} and \nth{2} orbit (see Formula \ref{math:Orbit n-graph}) of this article (see Table \ref{table:general_wikisubgraph_stats} for the corresponding corpus statistics). 
	Obviously, the most prominent \nth{2}-level DDC class in both TTNs is 510 (\textit{Mathematics}). 
	}
	\label{fig:First and Second Orbit of Integralrechnung}
\end{figure}

\begin{figure}[t]
	\includegraphics[width=0.45\textwidth]{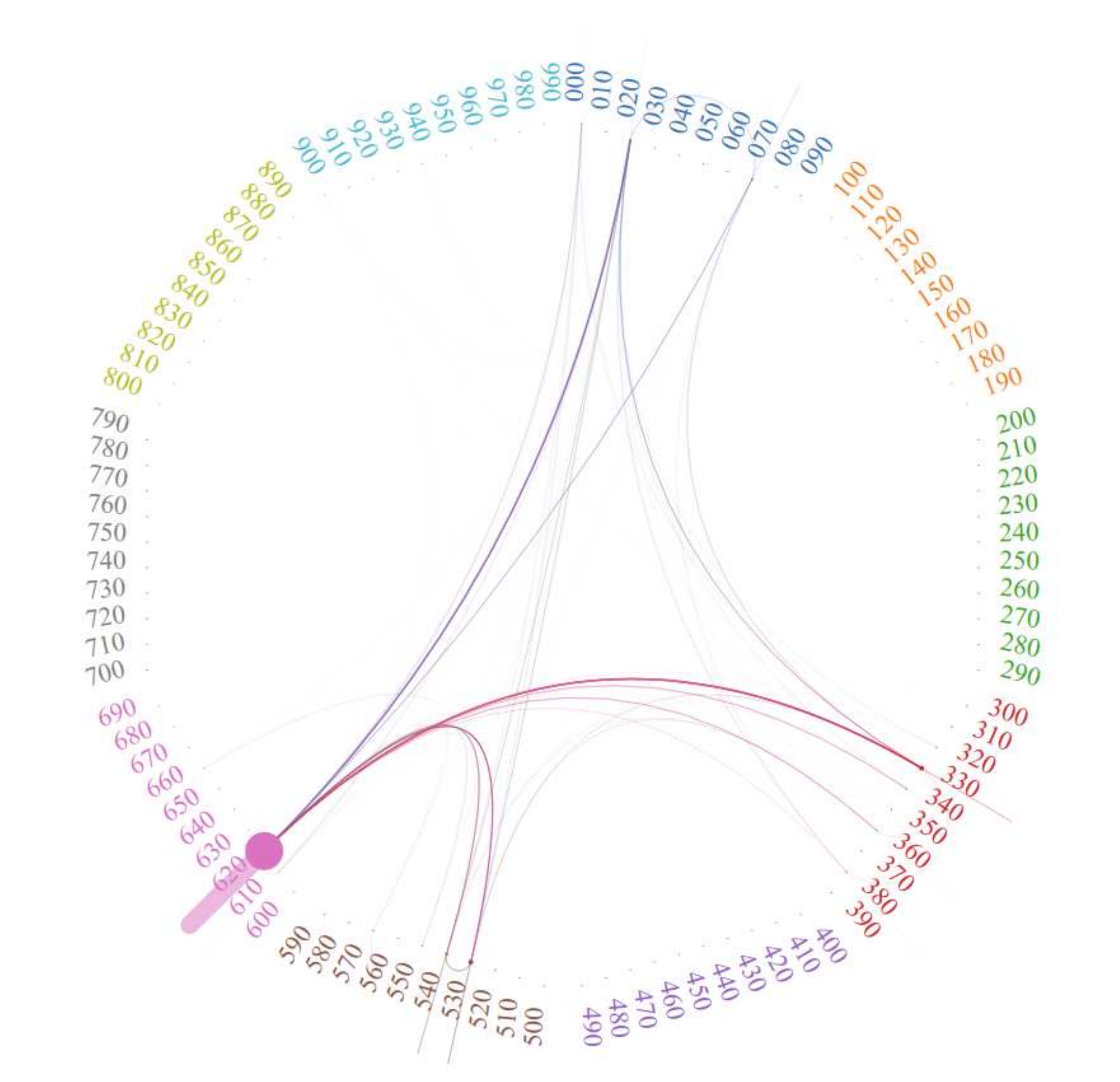}
	\hfill
	\includegraphics[width=0.45\textwidth]{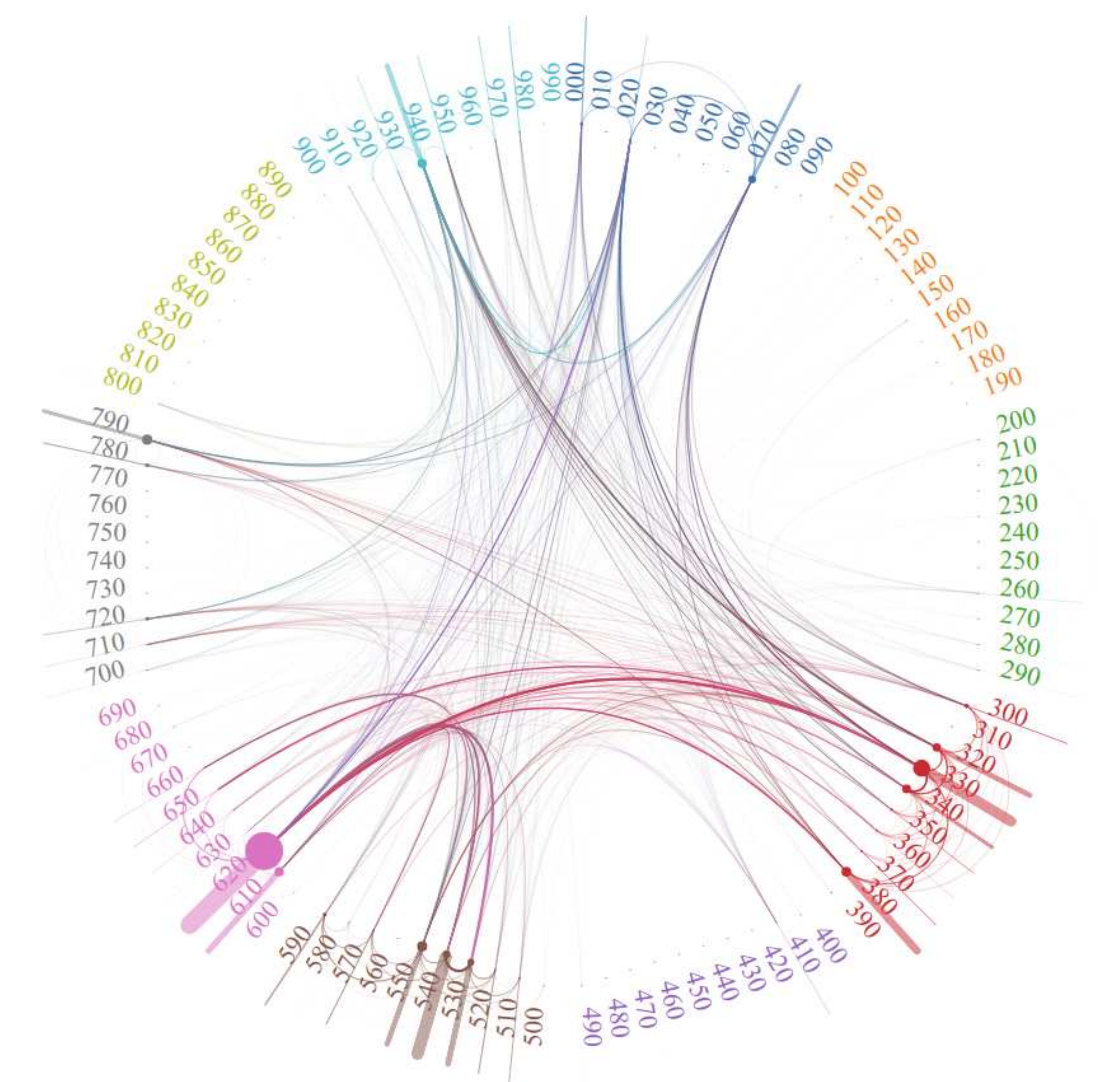}
	\caption{Visualization (by means of PolyViz \cite{Uslu:Mehler:2018}) of the TTN of the \nth{1} orbit (left) and of the \nth{2} orbit (right) of the German Wikipedia article \textit{Kernkraftwerk} (\textit{Nuclear power plant}).
	The TTNs are derived from the corpora of articles in the \nth{1} and \nth{2} orbit (see Formula \ref{math:Orbit n-graph}) of this article (see Table \ref{table:general_wikisubgraph_stats} for the corresponding corpus statistics). 
	Obviously, the most prominent \nth{2}-level DDC class in both TTNs is 620 (\textit{Engineering}). 
	Compared to the example in Figure \ref{fig:First and Second Orbit of Integralrechnung}, the \nth{2} orbit is now thematically much more diversified. 
	}
	\label{fig:First and Second Orbit of Kernkraftwerk}
\end{figure}

\begin{definition}\label{def:Multiplex Topic Network}
Given a definitional setting $\mathcal{S} = (\classification, \classify, \mathcal{L}(X,l))$ according to Definition \ref{def:Definitional Setting}, a \textbf{\textit{Multiplex Topic Network}} (MTN) is a $k$-layer network 
\begin{alignat}{2}\label{math:multiplex topic network}
\mathcal{M}(X,k) &=  
(\mathbb{M}, \mathbb{D}) \\ 
\mathbb{M} 
&= 
\{ 
	M_{i} = 
	(\Prime{V}{i}, \Prime{A}{i}, \Prime{\mu}{i}, \Prime{\nu}{i}, \Prime{\lambda}{i}, \Prime{\kappa}{i}) \mid i = 1..k
\} \\
\mathbb{D} 
&= 
\{ 
	D_{i\myto j} = 
	(\Prime{V}{i\myto j}, \Prime{A}{i\myto j}, \Prime{\mu}{i\myto j}, \Prime{\nu}{i\myto j}, \Prime{\lambda}{i\myto j}, \Prime{\kappa}{i\myto j}) \mid i,j = 1..k\!: i\not=j
\}
\end{alignat}
such that each $M_i$, $i \in\{1,\ldots, k\}$, is an $(L_1, \mathbb{L}_i')$-Topic Network derived from $\mathcal{S}$ according to Definition \ref{def:n-Layer Topic Network} and for each $i,j\in\{1,\ldots, l\}$, $i\not= j$, $D_{i\myto j}\in \mathbb{D}$, $|\mathbb{D}| = k(k-1)$, is called a \textit{margin layer} fulfilling the following requirements:
$\Prime{V}{i\myto j} = \Prime{V}{i}\cup \Prime{V}{j}$, 
$\Prime{A}{i\myto j} = \{(v,w)\in V_i\times V_j\mid \dot{v} = \dot{w} \}$, 
$\Prime{\mu}{i\myto j} = \Prime{\mu}{i}\cup \Prime{\mu}{j}$,
$\Prime{\lambda}{i\myto j} = \Prime{\lambda}{i}\cup \Prime{\lambda}{j}$.
\hfill $\Box$\end{definition}

See Figure \ref{fig:Comparing two Two-level Topic Networks} for a schematic depiction of the comparison of two MTNs.
Note that because of Definition \ref{def:n-Layer Topic Network}, it does not necessarily hold that $V_{\classification}(V_i) = V_{\classification}(V_j)$, but it always holds that $V_{\classification}(V_i) \subseteq V_{\classification} \supseteq V_{\classification}(V_j)$.
In this respect, we depart from \cite{Boccaletti:et:al:2014}, who instead require more strongly that $V_i = V_j$.
In the case of topic networks, this would be too restrictive, as different topic networks derived from the same definitional setting can focus on different subsets of topics, while ignoring the rest of the topics in the codomain $V_{\classification}$ of $\classify$.\footnote{A way to extend Definition \ref{def:Multiplex Topic Network} is to include the RCS $\classification = (V_{\classification}, A_{\classification})$ of Definition \ref{def:Definitional Setting} as an additional layer. This would allow for directly relating its constituent topic networks with the hierarchical classification system $\classification$.} 

In this paper, we quantify similarities of the different layers of MTNs to shed light on Hypothesis \ref{hypo:Hypothesis 1}.
More specifically: 
we generate an LMN for each corpus of a set of different text corpora in order to derive a separate two-layer MTN for each of these LMNs, each consisting of a TTN and an associated ATN. 
Then, among other things, we conduct a triadic classification experiment: 
firstly with respect to the subset of all TTNs derived from our corpus, secondly with respect to the subset of all corresponding ATNs and thirdly with respect to the subset of all TTNs in relation to the subset of the corresponding ATNs (see Figure \ref{fig:Classification Scenarios}).
In the next section, we explain the measurement procedure for carrying out this triadic classification experiment.

\subsection{A Procedural Model of Topic Network Analysis}\label{sec:A Procedural Model of Topic Network Analysis}

\begin{figure}[t]
	\includegraphics[width=1.00\textwidth]{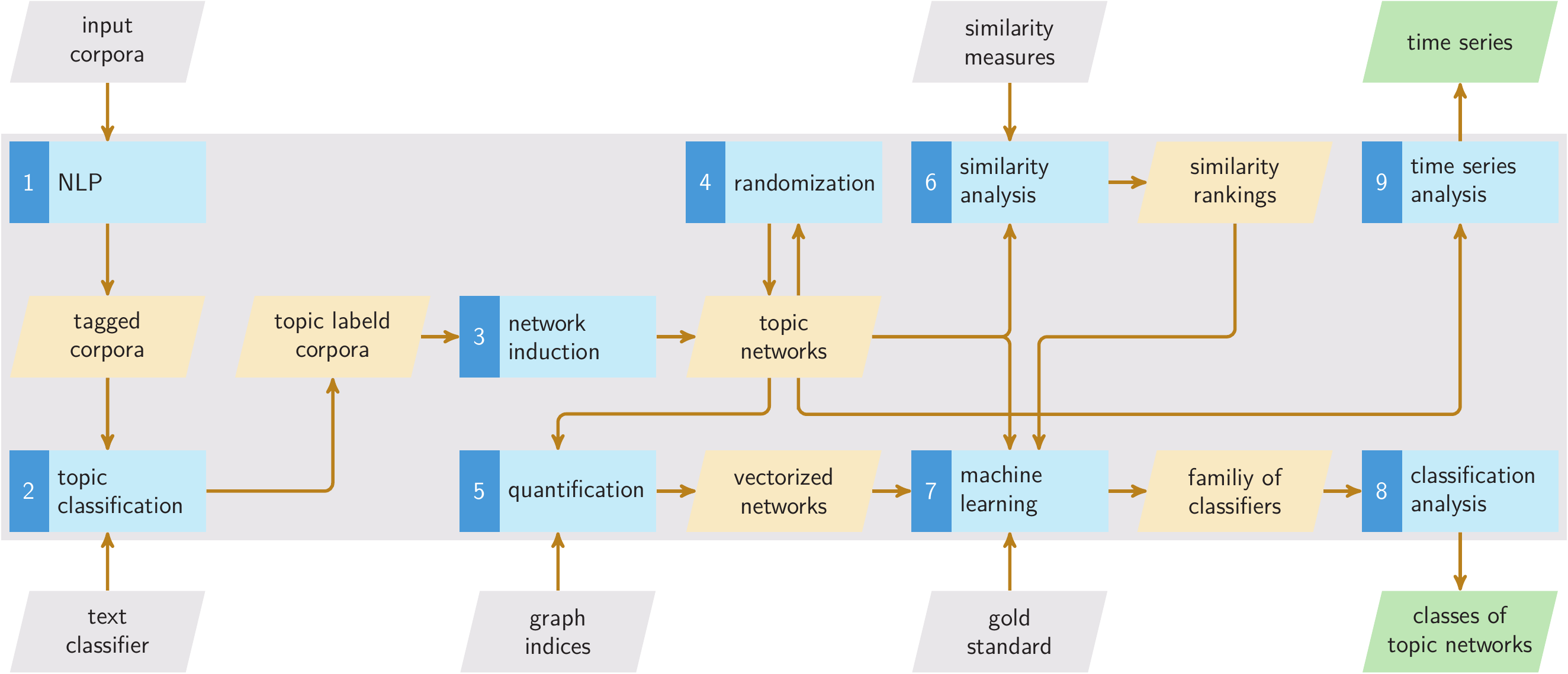}
	\caption{A procedural model of investigating LMNs and MTNs: 
	generating, randomizing and quantifying topic networks in 9 steps including \textit{Natural Language Processing} (NLP) (1), topic classification using a classifier $\theta$ according to Definition \ref{def:Definitional Setting} (2), topic network induction according to Definition \ref{def:Multiplex Topic Network} (3), network randomization according to Section \ref{sec:Module 4: Network Randomization} (4), network quantification (5) and network similarity analysis (6) both based on Section \ref{sec:Module 6: Graph Similarity Analysis}, machine learning of network classifiers (7) and classification analysis (8) both based on Section \ref{sec:Module 8 -- Machine Learning -- and Module 9 -- Classification Analysis} and, finally, time series analysis of topic networks (which will not be performed here) (9).}
	\label{fig:A Procedural Model of Topic Network Analysis}
\end{figure}

In order to instantiate topic networks as manifestations of the rhematic networking of places, we employ the procedure depicted in Figure \ref{fig:A Procedural Model of Topic Network Analysis}.
It combines nine modules for the induction, comparison and classification of topic networks.

\subsubsection{Module 1: Natural Language Processing}\label{sec:Module 1: Natural Language Processing}

Preparatory for all modules is the natural language processing of the input text corpora.
To this end, we utilize the NLP tool chain of \textit{TextImager} \cite{Hemati:Uslu:Mehler:2016} to carry out tokenization, sentence splitting, part of speech tagging, lemmatization, morphological tagging, named entity recognition, dependency parsing \cite{Bohnet:et:al:2013} and automatic disambiguation -- the latter by means of fastSense \cite{Uslu:et:al:2018:a}. 
For more details on these submodules see \cite{Eger:Mehler:Gleim:2016,Uslu:et:al:2018:a}.
As a result of Module 1, the topic classification can be fed with texts whose lexical components are disambiguated at the sense level.
As a sense model, we use the disambiguation pages of Wikipedia, currently the largest available model of lexical ambiguity.

\subsubsection{Module 2: Topic Classification}\label{sec:Module 2: Topic Classification}

\begin{figure}[t]
	\includegraphics[width=0.7\textwidth]{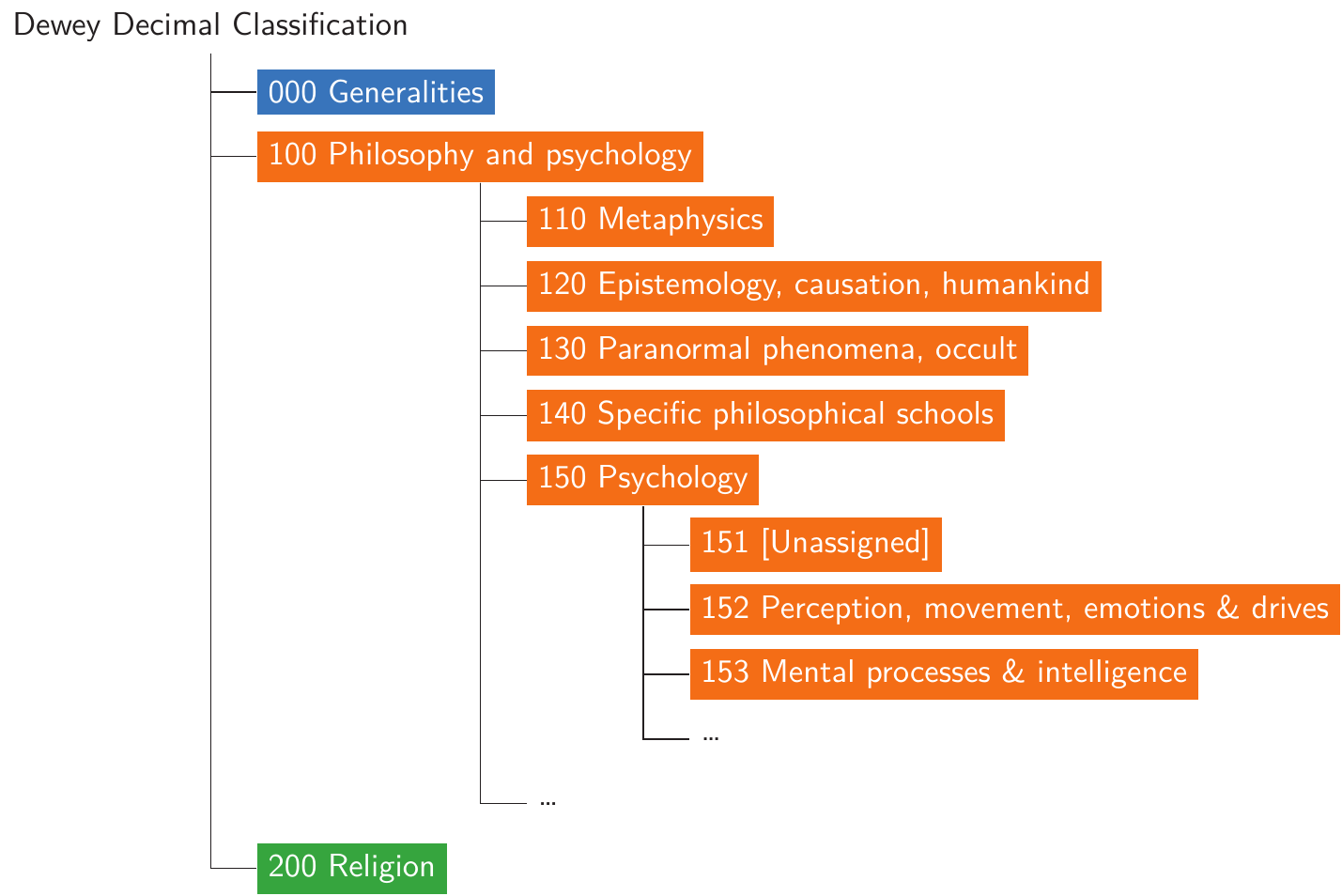}
	\caption{A subtree of the DDC displaying a snapshot of the second class (100) on the first three levels.}
	\label{fig:DDC Subtree}
\end{figure}

According to Definition \ref{def:Definitional Setting}, the derivation of TNs from LMNs requires the specification of a \textit{Reference Classification System} (RCS) $\classification = (V_{\classification}, A_{\classification})$.
For this purpose, we utilize the \textit{Dewey Decimal Classification} (DDC), a system that is well-established in the area of (digital) libraries.
As a result, the generalized tree $\classification$ from Definition \ref{def:Definitional Setting} degenerates into an ordinary tree since the DDC has no arcs superimposing its kernel hierarchy (see Figure \ref{fig:DDC Subtree} for a subtree of the DDC).
As a classifier $\classify$, which addresses the DDC, we use $\classify \coloneq \textit{text2ddc}$ \cite{Uslu:Mehler:Niekler:Baumartz:2018}, a topic classifier based on neural networks, which has been trained for a variety of languages \cite{Baumartz:Uslu:Mehler:2018}.\footnote{See \url{https://textimager.hucompute.org/DDC/}}
Starting from the output of Module 1 (NLP), we use text2ddc to map each input text $x$ to the distribution of the 5 top-ranked DDC classes that best match the content of $x$ as predicted by text2ddc.
Since text2ddc reflects the three-level topic hierarchy of the DDC, this classifier can output a subset of 98 classes of the \nth{2} (two classes of this level are unspecified) and a subset of 641 classes of the 3rd DDC level for each input text.\footnote{We did not have training for all 3rd-level classes (which are partly unspecified). 
See \cite{Uslu:Mehler:Niekler:Baumartz:2018} and the appendix for details.}
Thus, each topic network of each input corpus is represented on two levels of increasing thematic resolution.
Note that text2ddc classifies input texts of any size (from single words to entire texts in order to meet challenge \ref{enumerate:R3}, page \pageref{enumerate:R3}) and works as a multi-label classifier for processing thematically ambiguous input texts.
By using an RCS, text2ddc meets challenge \ref{enumerate:R2} simply by referring to the labels of the topic classes of the DDC.
Further, since text2ddc is trained with the help of a reference corpus, it can detect topics, even if they occur only once in a text (this is needed to meet challenge \ref{enumerate:R5}) and guarantees comparability for different input corpora (challenge \ref{enumerate:R1}).
text2ddc is based on fastText whose time complexity is $O(h \log_2(k))$, where \enquote{$k$ is the number of classes and $h$ the dimension of the text representation} \cite[2][]{Joulin:Grave:Bojanowski:Mikolov:2016} (making this classifier competitive compared to TMs). 

Figures \ref{fig:Dresden city wiki}, \ref{fig:TTN and ATNs of Muenchen City Wiki}, \ref{fig:First and Second Orbit of Integralrechnung} and \ref{fig:First and Second Orbit of Kernkraftwerk} show examples of TTNs and ATNs generated by means of text2ddc by addressing the second level of the DDC. 
Each of these topic networks was generated for a subset of articles of the German Wikipedia that are at most 2 clicks away from the respective start article $x$ (for the statistics of the corpora underlying these topic networks see Section \ref{sec:resources}). 
Formally speaking, let $G = (V,A)$ be a directed graph and $v\in V$; the $n$th orbit induced by $v$ is the subgraph
\begin{equation}\label{math:Orbit n-graph}
G_v^n = (V_v^n,A_v^n),\; 
V_v^n = \{w\in V\mid \delta(v,w) \le n\},\;
A_v^n = \{ (r,s)\in A\mid r,s\in V_v^n\}
\end{equation}
that is induced by the subset of vertices whose geodetic distance $\delta(v,w)$ from $v$ is at most $n$ (cf.\ \cite{Dehmer:2008:a}).
We compute the first and the second orbit of a set of Wikipedia articles (so that $G$ denotes Wikipedia's web graph).
This is done to obtain a basis for comparison for the evaluation of topic networks derived from special wikis.
Since Wikipedia is probably more strongly regulated than these special wikis, we expect higher disparities between networks of different groups (Wikipedia vs.\ special wiki) and smaller differences for networks of the same group.

\subsubsection{Module 3: Network Induction}\label{sec:Module 3: Network Induction}

Network induction is done according to the formal model of the Section \ref{sec:A Formal Model of Topic Networks}. 
It starts with inducing an LMN $\mathcal{L}(X,2)$ for each input corpus $X$.
That is, for each corpus $X$ we generate a text network $L_1$ and an agent network $L_2$ according to Definition \ref{def:Linguistic Multilayer Networks}:
\begin{enumerate}
\item In this paper, $X$ always denotes the set of texts (web documents) of a corresponding wiki $W$ 
so that the text layer $L_1 = (\Prime{V}{1}, \Prime{A}{1}, \Prime{\mu}{1}, \Prime{\nu}{1}, \Prime{\lambda}{1}, \Prime{\kappa}{1})$ of the LMN $\mathcal{L}(X,2)$, in which $L_2$ is an agent network defined below, can be used to represent the web graph \cite{Baldi:Frasconi:Smyth:2003} of this wiki.
Thus, for any two texts $x,y$ that are linked in $W$, we generate an arc $a = (v,w)\in \Prime{A}{1}$ where $\Prime{\nu}{1}(a) = 1$ and $\Prime{\kappa}{1}(a) = \mathtt{hyperlink}$. 
Further, for $\forall x\in \Prime{V}{1}\!: \Prime{\mu}{1}(x) = 1 \wedge \Prime{\lambda}{1}(x) = x$.

\item The author layer $L_2 = (\Prime{V}{2}, \Prime{A}{2}, \Prime{\mu}{2}, \Prime{\nu}{2}, \Prime{\lambda}{2}, \Prime{\kappa}{2})$ of the LMN $\mathcal{L}(X,2)$ corresponding to $L_1$ (see Definition \ref{def:Linguistic Multilayer Networks}) is generated as follows:
$\Prime{V}{2}$ is the set of all registered authors or TCP/IP addresses of anonymous users working on texts in $X$ so that 
$\forall v\in \Prime{V}{2}\!: \Prime{\lambda}{2}(v)$ maps to this name or IP address, respectively.
Let $\activity(r,x)$ be the sum of all additions made by author $r \in \Prime{V}{2}$ to any revision of the edit history of text $x$; we use $\activity(r,x)$ to approximate the more difficult to measure concept of authorship as introduced by Brandes et al.\ \cite{Brandes:Kenis:Lerner:vanRaaij:2009}.
Then we define: $\forall r\in \Prime{V}{2}\!: \Prime{\mu}{2}(r) = \sum_{x\in \Prime{V}{1}} \activity(r,x)$.
Further, $\Prime{A}{2}$ is the set of all arcs $(r,s)$ between users $r,s\in \Prime{V}{2}$, for which there is at least one text $x$ to which both contribute so that $\activity(r,x),\activity(s,x) > 0$.
Then, we define \cite[cf.][]{Mehler:Gleim:Hemati:Uslu:2017}: 
\begin{equation}\label{math:co-authorship}
\Prime{\nu}{2}(r,s) = \sum_{x\in \Prime{V}{1}} 2 \frac{\min(\activity(r,x),\activity(s,x))}{\sum_{u\in \Prime{V}{2}} \activity(u,x)}\in (0,1]
\end{equation}
Finally, $\Prime{\kappa}{2}(a) = \mathtt{coauthorship}$.
Obviously, $L_2$ is symmetric.
\end{enumerate}

Now, given the definitional setting $(\classification, \classify, \mathcal{L}(X,2))$, where $\classification, \classify$ are instantiated in terms of Section \ref{sec:Module 2: Topic Classification}, we induce a TTN $\topicnetwork(L_1) = (\Prime{V}{L_1}, \Prime{A}{L_1}, \Prime{\mu}{L_1}, \Prime{\nu}{L_1}, \Prime{\lambda}{L_1}, \Prime{\kappa}{L_1})$ according to Definition \ref{def:Text Topic Network} by means of appropriately defined monotonically increasing functions $\alpha_1, \beta_1, \gamma_1, \delta_1$.
To this end, we utilize the set 
\begin{equation}
\classify_x^{V_{\classification}} = \{ \classify_x(\dot{v})> \classify_{\min} \mid \dot{v}\in V_\classification \}
\end{equation}
of the membership values of text $x\in \Prime{V}{1}$ to the topics in $V_\classification$, where the parameter $\classify_{\min}$ denotes a lower bound of an acceptable degree of aboutness. 
We set $\classify_{\min} \coloneq 0$.
Further, by 
\begin{equation}
\bar{\classify} = \frac{1}{|\mathbb{Y}|}\sum_{y\in \mathbb{Y}} y
\end{equation}
we denote the mean value of the set $\mathbb{Y} = \cup_{x\in \Prime{V}{1}}\classify_x^{V_{\classification}}$ of selected topic membership values and by 
$\max(\mathbb{X},m)$ we denote the $m\in\{1,\ldots, |\mathbb{X}|\}$ largest value of the arbitrary set $\mathbb{X}$. 
Finally, we select a number $0 < m_{\bot} < |V_{\classification}|$ and define $\forall v\in V, \forall x\in \Prime{V}{1}$ thereby instantiating the parameters $\alpha,\beta,\gamma,\delta$ of the Formulas \ref{math:Cotextual Topic Network:1}--\ref{math:Cotextual Topic Network:4} of Definition \ref{def:Text Topic Network}: 
\begin{alignat}{2}\label{math:alpha beta}
\alpha &\coloneq \alpha_1 = \identity \\
\beta(\StackXY{x}{\dot{v}}{\classify}{\rclassify})
&\coloneq 
	\beta_1(\StackX{x}{\dot{v}}{\classify}) = 
	\beta_1(\classify(x, \Prime{\lambda}{}(v))) = 
	\beta_1(\classify_x(\dot{v}))
\nonumber \\
&= 
\begin{cases}
\classify_x(\dot{v}) & \classify_x(\dot{v}) \in 
\{ 
  r \in \classify_x^{V_{\classification}} |\, 
  \exists m \le m_{\bot}\!: 
   r = \max (\classify_x^{V_{\classification}},m) 
	 \ge \bar{\classify}
\} \label{math:Beta-Function of TTNs} \\
0 & \text{else}
\end{cases}
\\
\gamma &\coloneq \gamma_1 = \identity \\
\delta[
\StackXY{x}{\dot{v}}{\classify}{\rclassify}, 
\StackXY{y}{\dot{w}}{\classify}{\rclassify}, 
\StackX{x}{y}{\Prime{\nu}{1}} 
] &\coloneq \delta_1(
\StackX{x}{\dot{v}}{\classify}, 
\StackX{y}{\dot{w}}{\classify}, 
\StackX{x}{y}{\Prime{\nu}{1}} 
) \nonumber \\
&= 
\begin{cases}
\beta_1(\classify_x(\dot{v})) \beta_1(\classify_y(\dot{w})) & (x,y) \in \Prime{A}{1} \\
0 & \text{else}
\end{cases} 
\end{alignat}
According to Formula \ref{math:Beta-Function of TTNs}, $\beta_1(\XStackY{x}{\dot{v}}{\classify}{\rclassify}) = \classify_x(\dot{v})$ iff $\classify_x(\dot{v})$ is one of the $m_{\bot}$ highest membership values of $x$ to the topics in $V_{\classification}$, supposed that $\classify_x(\dot{v}) > \bar{\classify}$.
Otherwise, $\beta_1(\XStackY{x}{\dot{v}}{\classify}{\rclassify}) = 0$.
In this paper, we experiment with $m_{\bot} = 5$. 
The higher the value of $m_{\bot}$, the more sensitive the generation of $\topicnetwork(L_1)$ to the thematic ambiguity of the underlying texts. 
However, since $\classify$ creates a membership value for each pair of texts and topics, we use $\bar{\classify}$ as a lower bound of aboutness (in the sense of addressing a topic known by $\classify$) so that irrelevant classifications $\classify_x(\dot{v})$ do not affect $\mu_{L_1}(v)$.

Regarding the ATN $\topicnetwork(L_1, \{L_2\}) = (\Prime{V}{L_2}, \Prime{A}{L_2}, \Prime{\mu}{L_2}, \Prime{\nu}{L_2}, \Prime{\lambda}{L_2}, \Prime{\kappa}{L_2})$ corresponding to the TTN $\topicnetwork(L_1)$, we have to define monotonically increasing functions ${\alpha_2}, {\beta_2}, {\gamma_2}, {\delta_2}$.
To this end, we use several auxiliary functions:
\begin{itemize}
	\item By $\overline{\activity(\cdot,\cdot)}$ we denote the mean activity per author per Wikipedia article.
	\item By $\averageactivity$ we denote the average number of active authors per Wikipedia article. 
\end{itemize}
\begin{table}
	\resizebox{0.55\textwidth}{!}{
		\small
\end{comment}
 
\selectlanguage{english}
 
\nprounddigits{0}
\begin{tikzpicture}[
Kasten/.style={rectangle, fill=SeminarSehrHellGrau, draw=white, thin, anchor=west, minimum height=0.7cm}
]
\matrix [
	matrix of nodes,
	nodes=Kasten,
 	column 1/.append style={nodes={text width=4.5cm, fill=SeminarSehrHellGrau, text=black, align=left}},
 	column 2/.append style={nodes={text width=1.4cm, fill=SeminarSehrHellGrau, text=black, align=right}},
 	column 3/.append style={nodes={text width=1.4cm, fill=SeminarSehrHellGrau, text=black, align=right}},
 	row 1/.append style={nodes={fill=SeminarHellBlau, text=black, align=left}},
 	row 1 column 2/.append style={nodes={fill=SeminarHellBlau, text=black, align=left}},
] {
Corpus of Articles & $\averageactivity$ & $\overline{\activity(\cdot,\cdot)}$ \\ 
without redirects (\numprint{2195812}) & \nprounddigits{2}\numprint{27.343497075341606} & \nprounddigits{2}\numprint{234.51789920714248} \\
with redirects (\numprint{3657483}) & 
\nprounddigits{2}\numprint{17.06883203558294} & \nprounddigits{2}\numprint{226.61389387486702} \\
};
\end{tikzpicture}

	}
	\caption{Estimates of the average number of active authors per Wikipedia article ($\averageactivity$) and the average activity of authors per article ($\overline{\activity(\cdot,\cdot)}$) differentiated for the complete set of articles in the German Wikipedia (download at 2018-07-01) with and without redirect articles (numbers of articles in parentheses).}
	\label{tab:Wikipedia-based Estimations of Activity}
\end{table}

The corresponding estimators are found in Table \ref{tab:Wikipedia-based Estimations of Activity}.
Now, consider the set $\Prime{V}{2}(x)$ of all active authors of text $x$ and the set $\classify_v(\Prime{V}{1})$ of all texts that potentially contribute to $\Prime{\mu}{L_2}(v)$ and thus to the weight of the vertex $v\in \Prime{V}{L_2}$:
\begin{alignat}{2}
\Prime{V}{2}(x) &= \{ r\in\Prime{V}{2}\mid \activity(r,x) > 0\} \\
\classify_v(\Prime{V}{1}) &= \{x\in \Prime{V}{1}\mid \beta_1(\classify_x(\dot{v})) > 0\}
\end{alignat}
Then we define the following functions and ratios:
\begin{alignat}{3}
\scale &=
\begin{cases}
(0,1]^2 \to (0,2] \\
\scale(a,b) \mapsto 2\frac{a}{a+b}\\
\end{cases} \\
\omega_{x} &= 
\scale(|\Prime{V}{2}(x)|, \averageactivity) && \in (0,2] 
\label{math:rescaling 1}\\
\omega_{v} &= \frac{1}{|\classify_v(\Prime{V}{1})|}
\sum_{x\in \classify_v(\Prime{V}{1})} \omega_{x} &&\in (0,2]
\label{math:rescaling 2}
\end{alignat}
$\scale$ is a function which is used to rescale below or above average values (see Formula \ref{math:rescaling 1}).
Formula \ref{math:rescaling 2} defines the mean of the rescaled numbers of active users per article in $\classify_v(\Prime{V}{1})$.
Based on these preliminaries and regarding the vertex weighting function $\Prime{\mu}{L_2}$, we define $\forall v\in V, \forall r\in \Prime{V}{2}$ thereby instantiating the functions $\alpha$ and $\beta$ of Formula \ref{math:two-layer topic network:vertices} of Definition \ref{def:Two-layer Topic Network}: 
\begin{alignat}{2}\label{math:alpha beta gamma delta}
\alpha\coloneq {\alpha}_2 \wedge \forall z\in\mathbb{R}\!: {\alpha}_2(z) &= \omega_{v} \cdot z \\
{\beta}[ 
	\StackXY{x}{\dot{v}}{\classify}{\rclassify}, 
	\StackXY{r}{\dot{v}}{\oclassify}{\roclassify}, 
	\StackXY{r}{x}{\Prime{\nu}{i\myto 1}}{\Prime{\nu}{1\myto i}}
]
&\coloneq {\beta_2}(
	\StackX{x}{\dot{v}}{\classify}, 
	\StackX{r}{x}{\Prime{\nu}{2\myto 1}}
)
\nonumber \\
&= \beta_1(\classify_x(\dot{v})) \cdot
\begin{cases}
\frac{1}{p}\frac{\activity(r,x)}{\cdot \sum_{s\in\Prime{V}{2}} \activity(s,x)} & \activity(r,x) < \overline{\activity(\cdot,\cdot)} \\[.7em]
\frac{\activity(r,x)}{\sum_{s\in\Prime{V}{2}} \activity(s,x)} & \activity(r,x) = \overline{\activity(\cdot,\cdot)} \\[.7em]
p \frac{\activity(r,x)}{\sum_{s\in\Prime{V}{2}} \activity(s,x)} & \activity(r,x) > \overline{\activity(\cdot,\cdot)} 
\end{cases}\label{math:beta of ATN}
\end{alignat}
In the present paper, we experiment with $p = 2$.
To understand this definition, we have to run through the cases of Formula \ref{math:beta of ATN}:
\begin{enumerate}
\item The case $\activity(r,x) = \overline{\activity(\cdot,\cdot)}$: 
Suppose that for each $x\in \classify_v(\Prime{V}{1})$ the following condition holds: 
$\forall r,s\in \Prime{V}{2}(x)\!: \activity(r,x) = \activity(s,x) = \overline{\activity(\cdot,\cdot)}$. 
In this case, we obtain for each $x\in \classify_v(\Prime{V}{1})$ the following result:
\begin{equation}
\sum_{r\in\Prime{V}{2}} \beta_1(\classify_x(\dot{v})) \frac{\activity(r,x)}{\sum_{s\in\Prime{V}{2}} \activity(s,x)} = 
\beta_1(\classify_x(\dot{v})) \sum_{r\in\Prime{V}{2}} \frac{\activity(r,x)}{\sum_{s\in\Prime{V}{2}} \activity(s,x)} = \beta_1(\classify_x(\dot{v}))
\end{equation}
In other words: 
If all authors of all texts contributing to the weight of a topic contribute to these texts according to the average activity, the weight of this topic in the ATN corresponds to that of the corresponding TTN. 
In this case, the average activity does not bias the weight of a topic in the ATN compared to the same topic in the corresponding TTN.
Obviously, this scenario gives us a \textit{neutral point} or, more specifically, a \textit{calibration point} for the comparison of ATNs and TTNs. 
Such a calibration point allows us to interpret any down- or upward deviation of the topic weights in both networks, since no deviation means average activity and average number of active users.
However, this consideration presupposes that $\omega_{v} = 1$ so that ${\alpha_2} = \alpha_1 = \identity$.
If $\omega_{v} > 1$, then the number of authors of texts contributing to the weight of $v$ is on average higher than expected on the basis of Wikipedia, so that the weight of the topic $v$ in the ATN is \enquote{biased upwards} compared to the weight of the same topic in the corresponding TTN.
Conversely, if $\omega_{v} < 1$, then the number of authors of texts contributing to the weight of $v$ is on average smaller than expected, so that $v$'s weight in the ATN is \enquote{biased downwards} compared to the weight of the same topic in the corresponding TTN.
This scenario teaches us the different roles of ${\alpha_2}$ and ${\beta_2}$ with respect to the weighting of the $\beta_1$ values:
while ${\beta_2}$ operates as a function of the activities of authors, ${\alpha_2}$ considers their number.
\item The case $\activity(r,x) < \overline{\activity(\cdot,\cdot)}$: 
suppose for each $s\not= r$ that $\activity(s,x) = \overline{\activity(\cdot,\cdot)}$ while $\activity(r,x) < \overline{\activity(\cdot,\cdot)}$.
Then, we conclude:
\begin{alignat}{2}
\beta_1(\classify_x(\dot{v}))
\Bigl(
	\sum_{t\in\Prime{V}{2}\setminus \{r\}} \frac{\activity(t,x)}{\sum_{s\in\Prime{V}{2}} \activity(s,x)} +
	\frac{1}{p}\frac{\activity(r,x)}{\sum_{s\in\Prime{V}{2}} \activity(s,x)} 
\Bigr) &< \beta_1(\classify_x(\dot{v})) \Leftrightarrow \nonumber \\
\sum_{t\in\Prime{V}{2}\setminus \{r\}} \frac{\activity(t,x)}{\sum_{s\in\Prime{V}{2}} \activity(s,x)} +
\frac{1}{p}\frac{\activity(r,x)}{\sum_{s\in\Prime{V}{2}} \activity(s,x)} 
 &< 1 \Leftrightarrow \nonumber \\
\frac{1}{p}\frac{\activity(r,x)}{\sum_{s\in\Prime{V}{2}} \activity(s,x)} 
&< 
\frac{\activity(r,x)}{\sum_{s\in\Prime{V}{2}} \activity(s,x)} 
\Leftrightarrow 1 < p 
\end{alignat}
Thus, for $p> 1$ we penalize the contribution of a below-average active author of a text to the weight of the topic to which this text contributes.
The different effects of $\omega_{v} \lesseqgtr 1$ have already been discussed.

\item The case $\activity(r,x) > \overline{\activity(\cdot,\cdot)}$: 
if we suppose now that $\forall s\not= r\!: \activity(s,x) = \overline{\activity(\cdot,\cdot)}$ while $\activity(r,x) > \overline{\activity(\cdot,\cdot)}$, we conclude that for $p> 1$, we reward the contribution of an above-average active author of a text to the weight of the topic to which this text contributes.
\end{enumerate}

\noindent In a nutshell: ${\alpha_2}$ and ${\beta_2}$ implement the following proportionality assumptions:
\begin{itemize}
\item By ${\alpha_2}$ we penalize or reward under- or above-average co-authorships:
the higher the above-average number of authors contributing to the texts of a topic, the higher the reward effect, the higher the weight of the topic.
And vice versa: the lower the below-average number of authors contributing to the texts of a topic, the higher the penalty effect, the lower the weight of the topic.
\item By ${\beta_2}$ we penalize or reward under- or above-average activities of single authors:
the higher the above-average activity of a single author contributing to a text of a topic, the higher the reward effect, the higher the contribution of this author-text pair to the weight of the topic. 
And vice versa: the lower the below-average activity of a single author contributing to a text of a topic, the higher the penalty effect, the lower the contribution of this author-text pair to the weight of the topic.
\end{itemize}

Finally, we define the functions ${\gamma_2}$ and ${\delta_2}$ to get instantiations of the functions ${\gamma}$ and ${\delta}$ of Formula \ref{math:two-layer topic network:arcs} of Definition \ref{def:Two-layer Topic Network} (or, in the generalized case, of Formula \ref{math:n-level topic networks - nu} of Definition \ref{math:Jointly induced Topic Network}).
This is done by means of the following auxiliary function:
\begin{alignat}{2}
\Prime{\tilde{\nu}}{2}(r,s) &=
\scale(\Prime{\nu}{2}(r,s), \Prime{\overline{\nu}}{2}) \in \mathbb{R}^+
\end{alignat}
where $\Prime{\overline{\nu}}{2}$ estimates the average degree of co-authorship in Wikipedia according to Formula \ref{math:co-authorship}.\footnote{We estimate $\Prime{\overline{\nu}}{2}$ by means of \numprint{10000} randomly selected Wikipedia articles so that $\Prime{\overline{\nu}}{2}\coloneq \nprounddigits{6}\numprint{0.0027564072092594585}$.}
$\Prime{\tilde{\nu}}{2}(r,s)$ is a readjustment of $\Prime{\nu}{2}(r,s)$ in relation to the mean value $\Prime{\overline{\nu}}{2}$:
the higher the above-average co-authorship, the higher the value of $\Prime{\tilde{\nu}}{2}$ and the lower the below-average co-authorship, the lower the value of $\Prime{\tilde{\nu}}{2}$.
Then, we define: 
\begin{alignat}{2}
{\gamma} &\coloneq {\gamma_2} = \identity \\
{\delta}[
\StackXY{x}{\dot{v}}{\classify}{\rclassify},
\StackXY{y}{\dot{w}}{\classify}{\rclassify},
\StackXY{r}{\dot{v}}{\vartheta}{\roclassify},
\StackXY{s}{\dot{w}}{\vartheta}{\roclassify},
\StackXY{r}{x}{\Prime{\nu}{i\myto 1}}{\Prime{\nu}{1\myto i}}, 
\StackXY{s}{y}{\Prime{\nu}{i\myto 1}}{\Prime{\nu}{1\myto i}}, 
\StackX{r}{s}{\Prime{\nu}{i}},
\StackX{x}{y}{\Prime{\nu}{1}}
] 
& \coloneq \nonumber \\
{\delta_2}(
\StackX{x}{\dot{v}}{\classify},
\StackX{y}{\dot{w}}{\classify},
\StackX{r}{x}{\Prime{\nu}{2\myto 1}}, 
\StackX{s}{y}{\Prime{\nu}{2\myto 1}}, 
\StackX{r}{s}{\Prime{\nu}{2}},
\StackX{x}{y}{\Prime{\nu}{1}}
)
&
= \nonumber \\
\begin{cases}
\Prime{\tilde{\nu}}{2}(r,s) \cdot
{\beta_2}(\classify_x(\dot{v})) \cdot
{\beta_2}(\classify_y(\dot{w}))
& (x,y) \in \Prime{A}{1} \wedge (r,s) \in \Prime{A}{2} \\
0 & \text{else}
\end{cases}
& 
\end{alignat}
In this definition, ${\beta_2}(\classify_x(\dot{v}))$ quantifies the link $\StackX{x}{\dot{v}}{\classify}$ and the link $\StackX{r}{x}{\Prime{\nu}{2\myto 1}}$ (cf.\ Formula \ref{math:two-layer topic network:arcs}), the product ${\beta_2}(\classify_x(\dot{v})){\beta_2}(\classify_y(\dot{w}))$ quantifies the link $\Stack{x}{y}{\Prime{\nu}{1}}$ and $\Prime{\tilde{\nu}}{2}(r,s)$ quantifies the link $\Stack{r}{s}{\Prime{\nu}{2}}$.
The calibration point of arc weighting is now reached under the conditions of the following scenario (for the first two conditions see above):
\begin{alignat}{2}
{\beta_2}(\classify_x(\dot{v})) &= \beta_1(\classify_x(\dot{v})) \\
{\beta_2}(\classify_y(\dot{w})) &= \beta_1(\classify_y(\dot{w})) \\ \Prime{\tilde{\nu}}{2}(r,s) &= 1 
\end{alignat}
Under these conditions, the authors $r$ and $s$ contribute to text $x$ and $y$ at an average level while interacting at an average level of co-authorship.
In this case, the (co-)authorship of both authors does not influence the strength of the corresponding arc in the ATN: 
neither in terms of reducing nor of increasing $\Prime{\nu}{2}(v,w)$.
Note that the size of an ATN (i.e., the number of its arcs) is always less than or equal to that of the corresponding TTN, since the arcs present in a TTN are merely re-weighted in the corresponding ATN: 
no new arcs are added.
The same holds for the order of the ATN since there is no node in a TTN for which there is no author authoring it.

Our instantiation of multiplex text and author topic networks has shown two points: 
firstly, we demonstrated a single parameter setting as an element of a huge parameter space spanned by parameters such as 
$p$, 
$\Prime{\overline{\nu}}{2}$, 
$\overline{\activity(\cdot,\cdot)}$, 
$\averageactivity$, 
$\classify$,
$\alpha_1$,
$\alpha_2$, 
$\beta_1$, 
$\beta_2$, 
$\gamma_1$, 
$\gamma_2$, 
$\delta_1$, 
$\delta_2$ 
etc.\footnote{In the latter eight cases various information links are included as candidate parameters. Formula \ref{math:beta of ATN} shows, for example that out of the six possible information links, only two are evaluated to instantiate $\beta_2$. Obviously, numerous alternatives exist to instantiate this function.}
Secondly, anyone who complains about the apparently inherent parameter explosion in our approach should consider the hyperparameter spaces of neuronal networks as an object of parameter optimizations.
Regardless of the heuristic character of our approach, compared to the black box character of neural networks, its settings are extensible on the basis of the schematic framework provided by Definition \ref{def:Multiplex Topic Network} of MTNs and the definitions it is based upon. 
At the same time, this approach guarantees interpretability as long as the different ingredients entering our model via formulas of the sort as Formula \ref{math:n-level topic networks - mu} and Formula \ref{math:n-level topic networks - nu} fulfill this condition -- in order to meet challenge \ref{enumerate:R7}.

\subsubsection{Module 4: Network Randomization}\label{sec:Module 4: Network Randomization}

Randomization is conducted to assess the significance of our findings.
This is necessary because there is currently no related classification in the area examined here that can serve this role.
To fill this gap, we compute the following randomizations:
\begin{enumerate}
	\item \textbf{Baseline B1:} A lower bound of a baseline is obtained by randomly assigning the object networks onto the gold standard (target) classes. 
	This can be done by informing the assignment about the true cardinality of these classes {(B11)} or not {(B12)}.
	We opt for B11 since this variant yields a higher F-score, making it more difficult to surpass.
	Of course, any serious network representation and classification model should go beyond this baseline.
	B1 will be averaged over \numprint{100000} iterations.
	
	\item \textbf{Baseline B2:} 
	An alternative is to randomize the input networks and to derive vector representations (according to Section \ref{sec:Module 3: Network Induction}), which ultimately undergo the same classification process as the original networks.
	That is, the input networks are randomly rewired to generate Erd\H{o}s-R{\'e}nyi (ER) graphs, for which we ask whether they are separable by the same classification model.\footnote{An alternative, not considered here, would be to randomize the topic classification of the underlying texts.}
	If this is successful (in terms of high $F$-scores\footnote{The $F$-score is a measure of the accuracy of a classification, that is, the harmonic mean of its precision and recall.}), then we conclude that the network representation model or the operative classifier is not informative enough regarding the hypothetical class memberships of the input networks.
	Conversely, the lower the average $F$-scores obtained by classifying the randomized networks compared to the classification of the original ones, the more informative the representation model or the classification procedure regarding the underlying hypotheses.
	By keeping the model constant while varying the classifier we can ultimately attribute this (non-)informativity to the underlying representation model.
	Conversely, by keeping the classifier constant while varying the model we can attribute this informativity to the classification model.
	B2 will be repeated 100 times. 
	
	\item \textbf{Baseline B3:}
	A third baseline results from randomizing the matrices that form the input of the target classifiers. 
	This means that instead of calculating graph invariants or similarity values to feed the classifiers, we use matrices whose dimensions are chosen uniformly at random from the domain of the corresponding invariants or (dis-)similarity measures.\footnote{We require that the main diagonal of the random matrix is 1 and that it is symmetric.}
	If the classification based on the original networks does not exceed this baseline, we are again informed about a deficit of our representation model.
	Evidently, we are looking for models that significantly exceed this baseline; otherwise we would have to accept that the same classifiers perform better on random values than on our feature model.
	B3 will be repeated 100 times. 
	
	\item \textbf{Baseline B4:}
	Finally, we start from randomly reorganizing the set of observations into random classes while using the same representation model to separate the resulting random gold standard.\footnote{Obviously, we have to prevent that the gold standard is ever part of the set of these randomizations.}
	We choose the variant of using randomized cardinalities of the random classes rather than keeping the sizes of the gold standard. Tests have shown that this approach tends to generate higher $F$-scores than the latter.
	If our network representation and classification model does not outperform this baseline, we learn that the underlying invariants used to characterize the networks are not \textit{specific} enough: 
	rather, they can be related to random classifications of the same objects using the same feature space.
	Obviously we are looking for a model characterizing the gold standard (\textit{tendency to specificity}) and not a random counterpart of it (\textit{tendency to non-specificity}).
	B4 is averaged over 100 repetitions.
\end{enumerate}

B1 is a lower bound: 
models that fall under this bound are obsolete.
B2 concerns the evaluation of the network representation or classification model. 
B3 focuses on evaluating the classification model, and B4 aims to evaluate the specificity of the operative feature model.

\subsubsection{Module 5: Network Quantification}\label{sec:Module 5: Network Quantification}

Module 5 is a preparatory step for a subset of network similarity measures.
This relates to so-called topology-based approaches to graph similarity \cite{Mehler:2008:a,Abramov:Mehler:2011:a,Macindoe:Whitman:2010,Li:Semerci:Yener:Zaki:2011,Li:Dong:Shi:Dehmer:2017}.
The idea behind this approach is to map input networks onto vectors of graph indices or invariants to compare them with each other.
That is, graph similarity is traced back to similarity in vector space: 
the higher the number of indices for which two graphs resemble each other, the more similar the graphs.
The apparatus that we employ in this context is described next.

\subsubsection{Module 6: Graph Similarity Analysis}\label{sec:Module 6: Graph Similarity Analysis}

Our hypothesis about thematic networks on geographical places says that these networks are similar in terms of the skewness of their thematic focus and their network structure, regardless of whether the underlying texts are written by different communities and regardless of the framing theme.
To test this hypothesis, we apply the framework of graph similarity measurement which allows for mapping the second of these three reference points by exploring the structure of topic networks as well as features of their nodes.
Since graph similarity measurement is generally known to be computational complex, we take profit from the fact of dealing with \textit{labeled} graphs.
By using alignments of the labels of the nodes of the graphs to be compared, we reduce the time complexity of these approaches enormously.

The literature knows a number of approaches for graph similarity measurement.
Among other things, this includes the following approaches (see Emmert-Streib et al.\ \cite{Emmert-Streib:Dehmer:Shi:2016} for an overview \cite[cf.][]{Koutra:et:al:2011,Koutra:et:al:2016}; the paper does not aim at a comprehensive study of them, but focuses on a selected subset):

\begin{enumerate}
	\item \textit{Graph Edit Distance} (GED) based approaches \cite{Bunke:Dickinson:Kraetzl:Wallis:2006,Ibragimov:Malek:Guo:Baumbach:2013,Wallis:Shoubridge:Kraetz:Ray:2001} and their relatives (e.g.\ the \textit{Vertex and Edge Overlap} (VEO) \cite{Papadimitriou:Dasdan:Garcia-Molina:2008}),
	\item spherical \cite{Dehmer:2008:a} or neighborhood-related approaches \cite[cf.][]{Koutra:et:al:2016} and
	\item network topology-related approaches \cite{Mehler:2008:a,Abramov:Mehler:2011:a,Macindoe:Whitman:2010,Li:Semerci:Yener:Zaki:2011,Li:Dong:Shi:Dehmer:2017,Papadimitriou:Dasdan:Garcia-Molina:2008}.
\end{enumerate}
We will develop and test candidates of each of these classes.

GED-based methods are well studied in the area of web mining \cite{Schenker:Bunke:Last:Kandel:2005}. 
Since we are dealing with labeled graphs, we can compute the GED directly from the vertex and edge sets of the input graphs \cite{Bunke:Dickinson:Kraetzl:Wallis:2006,Koutra:et:al:2016}. 
Let $G_1 = (\Prime{V}{1}, \Prime{A}{1}, \Prime{\mu}{1}, \Prime{\nu}{1}, \Prime{\lambda}{1}, \Prime{\kappa}{1}), G_2 = (\Prime{V}{2}, \Prime{A}{2}, \Prime{\mu}{2}, \Prime{\nu}{2}, \Prime{\lambda}{2}, \Prime{\kappa}{2})$ be two TNs, then their GED is computed as follows:
\begin{alignat}{2}
\GED(G_1, G_2) &= 
|V_1| + |V_2| - 2|V_{\classification}(V_1)\cap V_{\classification}(V_2)| + |A_1| + |A_2| - 2|V_{\classification}(A_1)\cap V_{\classification}(A_2)| \in \mathbb{R}^+_0 
\label{math:ges:1}
\end{alignat}
where $V_{\classification}(A_i) = \{(\dot{v},\dot{w})\mid (v,w)\in A_i\}, i = 1..2$.
Since we are targeting graph similarities, we consider $\ges$ instead of $\GED$, where overlaps of vertex and arc sets are equally weighted:
\begin{alignat}{2}
\ges(G_1, G_2) &= 1 - \frac{1}{2}\left(\frac{|V_1| + |V_2| - 2|V_{\classification}(V_1)\cap V_{\classification}(V_2)|}{|V_1| + |V_2|} + \frac{|A_1| + |A_2| - 2|V_{\classification}(A_1)\cap V_{\classification}(A_2)|}{|A_1| + |A_2|}\right)
\in [0, 1] \label{math:ges:2}
\end{alignat}
The same is done in the case of Wallis' approach to graph distance \cite{Wallis:Shoubridge:Kraetz:Ray:2001} which is adapted as follows to get a similarity measure:
\begin{alignat}{2}
\wal(G_1, G_2) &= \frac{|V_{\classification}(V_1)\cap V_{\classification}(V_2)| + |V_{\classification}(A_1)\cap V_{\classification}(A_2)|}{|V_1| + |V_2| + |A_1| + |A_2| - |V_{\classification}(V_1)\cap V_{\classification}(V_2)| - |V_{\classification}(A_1)\cap V_{\classification}(A_2)|}
&&\in [0, 1] \label{math:wal}
\end{alignat}

A relative of $\ges$ is the \textit{Vertex/Edge Overlap} (VEO) graph similarity measure \cite{Papadimitriou:Dasdan:Garcia-Molina:2008}:
\begin{alignat}{2}
\veo(G_1, G_2) &= 2\frac{|V_{\classification}(V_1)\cap V_{\classification}(V_2)| + |V_{\classification}(A_1)\cap V_{\classification}(A_2)|}{|V_1| + |V_2| + |A_1| + |A_2|} \\ &
= 1 - \frac{\GED(G_1, G_2)}{|V_1| + |V_2| + |A_1| + |A_2|} &&\in [0, 1] \label{math:veo} 
\end{alignat}

Since node and arc weights are not taken into account by these measures, we compute the following variant of $\ges$ to close this gap: 
\begin{alignat}{2}
\forall x,y\in \mathbb{R}^+_0\!: \Distance(x,y) &= \frac{|x - y|}{\max(x, y)} &&\in [0, 1] \\
\forall v\in V_1 \forall w\in V_2\!: \wges(v, w) &= 
\begin{cases}
\Distance(\mu_{1}(v), \mu_{2}(w)) & \dot{v} = \dot{w} \\
0 & \text{else}
\end{cases}
&&\in [0, 1] \\
&\forall a = (v, w)\in A_1 \forall b = (x, y)\in A_2\!: \nonumber\\
\wges(a, b) &= \begin{cases}
\Distance(\nu_{1}(a), \nu_{2}(b)) & \dot{v} = \dot{x} \wedge \dot{w} = \dot{y} \\
0 & \text{else}
\end{cases}
&&\in [0, 1] \\
\wges(V_1, V_2) &= \frac{|V_1| + |V_2| - 2 \sum_{v\in V_1, w\in V_2} \Distance(\mu_{1}(v), \mu_{2}(w))}{|V_1| + |V_2|} &&\in [0, 1] \\
\wges(A_1, A_2) &= \frac{|A_1| + |A_2| - 2 \sum_{a\in A_1, b\in A_2} \Distance(\nu_{1}(a), \nu_{2}(b))}{|A_1| + |A_2|} &&\in [0, 1] \\
\wges(G_1, G_2) &= \frac{\wges(V_1, V_2) + \wges(A_1, A_2)}{2} &&\in [0, 1] \label{math:wges}
\end{alignat}
$\wges$ is sensitive to arc \cite{Koutra:et:al:2016} \textit{and} to vertex weights of TNs, the latter measuring the membership degree of the underlying texts to the topic represented by the corresponding vertex.
We say that such measures are \textit{dual weight-dependent}. 
These measures are of high interest since they cover more information of the underlying networks than single weight- or even weight-independent measures (cf.\ the axiom of edge weight sensitivity of Koutra et al.\ \cite{Koutra:et:al:2016}).

GED and its relatives share a view of similarity, according to which graphs are considered to be more similar the more (equally weighted) vertices and arcs they share.
This notion of similarity is contrasted by spherical approaches (see above) as exemplified by DeltaCon \cite{Koutra:et:al:2016}.
Roughly speaking, according to DeltaCon, the more similar two graphs resemble each other from the perspective of their vertices, the more similar they are.
Since DeltaCon is not dual weight-dependent, we consider a dual weight-dependent relative of it.
To this end, we compute the cosine of the vectors of geodetic distances for each pair of equally labeled vertices.
Since topic networks can differ in their order, we first have to align their node sets to make them comparable -- this is also needed because we aim for a dual weight-dependent measurement. 
The required alignment is addressed by means of the following auxiliary graphs $G_{12}$ and $G_{21}$:
\begin{alignat}{2}
\forall i, j\in \{1, 2\}, i \not= j\!: G_{ij} &= (V_{ij}, A_i, \mu_{ij}, \nu_{i}, l_{ij}) \\
V_{ij} &= V_i \cup \{w\in V_j\mid \nexists v\in V_i\!: \dot{v} = \dot{w}\} \\
\forall v\in V_{ij}\!: \mu_{ij}(v) &= 
\begin{cases}
\mu_{i}(v) & v \in V_i \\
0 & \text{else}
\end{cases} \\
\forall v\in V_{ij}\!: l_{ij}(v) &= 
\begin{cases}
l_{i}(v) & v \in V_i \\
l_{j}(v) & \text{else}
\end{cases}
\end{alignat}

$G_{12}$ and $G_{21}$ are needed to make $G_1$ and $G_2$ comparable whose symmetric difference $V_1 \bigtriangleup V_2$ can be non-empty while their vertex labeling functions share the same codomain (since $G_1$ and $G_2$ belong to the same multiplex topic network according to Definition \ref{def:Multiplex Topic Network}).
Obviously, $|G_{12}| = |G_{21}|$ so that for each $v\in V_i, w\in V_{ij}\setminus V_i; i,j \in \{1,2\}, i\not= j,$ there is no path from $v$ to $w$ in $G_{ij}$. 
Cases in which no such path exists are denoted by $v\not\sim w$; otherwise, if such a path exists, we denote by $\ged_{ij}(v,w)$ the length of the shortest path, that is, the geodetic distance between $v$ and $w$ in $G_{ij}$.
As we deal with graph similarities, we first transform the distance values into similarity values:
\begin{alignat}{2}\label{math:Geodesic Proximity}
\forall v,w\in V_{ij}\!: \gep_{ij}^{[\omega,\iota]}(v, w) &= 
\begin{cases}
1 - \frac{\ged_{ij}^{[\omega,\iota]}(v, w)}{|V_{ij}|} & v,w \in V_i \\
0 & \text{else}
\end{cases}
&&\in [0, 1]
\end{alignat}
$\gep$ is short for \textit{geodetic proximity}. 
With the denominator $|V_{ij}|$ we penalize situations in which there is no path between $v$ and $w$, that is, $v\not\sim w$. 
The parameter $\omega\in \{\text{w}, \neg{\text{w}}\}$ specifies, whether the geodetic distance $\ged_{ij}^{[\omega,\iota]}$ and the geodetic proximity $\gep_{ij}^{[\omega,\iota]}$ are computed for the weighted ($\text{w}$) or unweighted ($\neg{\text{w}}$) variant of $G_{ij}$.
If $\omega = \text{w}$, we assume that each arc weighting value is normalized by means of the non-zero maximum value assumed by the arc weighting function for this network.\footnote{This means that a graph $G_2$, which is obtained from a graph $G_1$ by multiplying the weights of all arcs of $G_1$ by a factor $c>0$, will be equal to $G_1$ in terms of the graph similarity measure to be introduced now (insensitivity to certain scalings).}
$\iota\in \mathbb{R}^+_0$ specifies the maximum geodetic distance to be considered: 
beyond this value, nodes $w$ are considered to be of maximum geodetic distance $|V_{ij}|$ to $v$ -- irrespective of their real distance.
For $\iota \ge |V_{ij}|$, we have to compute all geodetic distances. 
For values of $\iota \ll |V_{ij}|$ (e.g.\ $\iota = 2$), we arrive at variants of $\gep_{ij}$ that are less time complex.
We consider the variant $\iota = \infty$ s that we take all path-related information into account.
Now, we calculate the dual weight-dependent cosine of $G_1$ and $G_2$ as follows:
\small
\begin{alignat}{2}
\forall v\in V_{12} \forall w\in V_{21}\!: \cos[\omega,\iota](v, w) &= 
\frac{\sum\limits_{x\in V_{12}, y\in V_{21}, \dot{x}= \dot{y}} \gep_{ij}^{[\omega,\iota]}(v, x)\gep_{ij}[\omega,\iota](w, y)}{\sqrt{\sum\limits_{u\in V_{12}} \gep_{ij}^{[\omega,\iota]}(v, u)^2}\sqrt{\sum\limits_{u\in V_{21}} \gep_{ij}^{[\omega,\iota]}(w, u)^2}} &&\in [0, 1] \label{math:Vertex-related cosine}\\
\cos_{\mathcal{A}}[\omega,\iota,\phi,\mathbb{L}](G_1, G_2) &= \frac{\sum\limits_{v\in V_{12}, w\in V_{21}, \dot{v}= \dot{w}\in \mathbb{L}} \phi(v,w)\cos[\omega,\iota](v, w)}{\sum\limits_{v\in V_{12}, w\in V_{21}, \dot{v}= \dot{w} \in \mathbb{L}} \phi(v,w)} &&\in [0, 1] \label{math:Edge-related cosine} \\
\cos_{\mathcal{V}}(G_1, G_2) &= 
\frac{\sum\limits_{v\in V_{12}, w\in V_{21}, \dot{v}= \dot{w}} \mu_{12}(v)\mu_{21}(w)}{\sqrt{\sum\limits_{v\in V_{12}} \mu_{12}(v)^2}\sqrt{\sum\limits_{w\in V_{21}} \mu_{21}(w)^2}} &&\in [0, 1] \label{math:Cosine of two Vertex Sets} \\
\cos_{\mathcal{A}\mathcal{V}}[\omega,\iota,\phi,\mathbb{L}](G_1, G_2) &= \frac{\cos_{\mathcal{V}}(G_1, G_2) + \cos_{\mathcal{A}}[\omega,\iota,\phi](G_1, G_2)}{2} &&\in [0, 1] \label{math:cosine Graph Similarity}
\end{alignat}
\normalsize
$\cos[\omega,\iota,\phi,\mathbb{L}](G_1, G_2)$ is the weighted cosine of the vectors of geodetic proximities of the same-named vertices in $G_{12}$ and $G_{21}$. 
In this article, we consider two instantiations of parameter $\phi$:
\begin{alignat}{2}
\forall v\in V_{12}, w\in V_{21}, \dot{v}= \dot{w}\!: \phi_1(v,w) &= 1 \label{math:phi as Average} 
\\
\forall v\in V_{12}, w\in V_{21}, \dot{v} = \dot{w}\!: \phi_2(v,w) &= \max(d(v), d(w)) \label{math:phi as Weeighting by Degree}
\end{alignat}
$\phi_1$ implements an arithmetic mean.
$\phi_2$ is a function of the degree centrality \cite{Freeman:1978} of its arguments: 
the more linked a topic in a network, the higher its impact onto the similarity of the input networks.
The similarity view behind this approach is that while $\cos_X[\omega,\iota,\phi_1,\mathbb{L}], X\in \{ \mathcal{A}, \mathcal{A}\mathcal{V} \}$, treats all -- peripheral or central -- nodes equally, $\cos_X[\omega,\iota,\phi_2,\mathbb{L}]$ gives central nodes more influence.
Take the example of two city networks \cite{Blanchard:Volchenkov:2009}: 
it is plausible to say that if city networks look similar from the point of view of their central places, this should have more impact on the general similarity assessment than similarities from the point of view of peripheral locations.
An extension would be to use more informative node weighting measures (e.g.\ closeness centrality).
Finally, parameter $\mathbb{L}$ limits the number of vertices for which cosine values are computed.
In the unlimited case, $\mathbb{L} \coloneq \mathbb{L}_{12} = \{l_{12}(v)\mid v\in V_{12}\}$.
\begin{table}[t]
	\centering
	\begin{tikzpicture}
	[Kasten/.style={rectangle, minimum height=0.60cm, anchor=west}, Path/.style={font={\small\vphantom{Ag}}, midway, above},]
	\matrix [matrix of nodes, nodes=Kasten, row sep=0.05cm, column sep=0.05cm, font={\small\vphantom{Ag}}, 
	row 1/.append style={nodes={fill=SeminarSehrHellBlau}},
	column 1/.append style={nodes={text width=0.30cm, align=right, fill=SeminarSehrHellBlau}},
	column 2/.append style={nodes={text width=3.25cm, align=left, fill=SeminarSehrSehrHellGrau}},
	column 3/.append style={nodes={text width=5.0cm, align=left, fill=SeminarSehrSehrHellGrau}},
	column 4/.append style={nodes={text width=1.5cm, align=center, fill=SeminarSehrSehrHellGrau}},
	column 5/.append style={nodes={text width=1.5cm, align=left, fill=SeminarSehrSehrHellGrau}},
	column 6/.append style={nodes={text width=1.00cm, align=right, fill=SeminarSehrSehrHellGrau}},
	] (Link) {
		& Measure & Approach & Formula & Reference \\
		1. & $\ges$ & graph edit similarity & (\ref{math:ges:2}) & \cite{Bunke:Dickinson:Kraetzl:Wallis:2006} \\
		2. & $\wal$ & graph edit similarity & (\ref{math:wal}) & \cite{Wallis:Shoubridge:Kraetz:Ray:2001} \\
		3. & $\veo$ & vertex and edge overlap & (\ref{math:veo}) & \cite{Papadimitriou:Dasdan:Garcia-Molina:2008} \\
		4. & $\wges$ & weighted graph edit similarity & (\ref{math:wges}) & { } \\
		5. & $\cos_{\mathcal{A}}[\text{w},\infty,\phi_1,\mathbb{L}_{12}]$ & cosine graph similarity & (\ref{math:Edge-related cosine}) & { } \\
		6. & $\cos_{\mathcal{A}\mathcal{V}}[\text{w},\infty,\phi_1,\mathbb{L}_{12}]$ & cosine graph similarity & (\ref{math:cosine Graph Similarity}) & { } \\
		7. & $\cos_{\mathcal{A}\mathcal{V}}[\text{w},\infty,\phi_2,\mathbb{L}_{12}]$ & cosine graph similarity & (\ref{math:cosine Graph Similarity}) & { } \\
		8. & $\cos_{\mathcal{A}}[\neg{\text{w}},\infty,\phi_1,\mathbb{L}_{12}]$ & cosine graph similarity & (\ref{math:Edge-related cosine}) & { } \\
		9. & $\cos_{\mathcal{A}\mathcal{V}}[\neg{\text{w}},\infty,\phi_1,\mathbb{L}_{12}]$ & cosine graph similarity & (\ref{math:cosine Graph Similarity}) & { } \\
		10. & NetSimile & topological similarity & (\ref{math:NetSimile}) & \cite{Berlingerio:et:al:2013} \\
		11. & ToSi & topological similarity & (\ref{math:ToSi}) & { } \\
	};
	\end{tikzpicture}
	\caption{The list of measures of graph similarity used for computing the similarities of topic networks.}\label{tab:Graph Similarity Measures}
\end{table}
It is easy to see that Formulas \ref{math:Edge-related cosine}, \ref{math:Cosine of two Vertex Sets} and \ref{math:cosine Graph Similarity} are similarity measures. For $X\in \{ \mathcal{A}, \mathcal{V}, \mathcal{A}\mathcal{V} \}$, this can be shown as follows: 
\begin{enumerate}
\item \textit{Symmetry:} 
$\cos_{X}[\omega,\iota,\phi,\mathbb{L}](G_1, G_2) = \cos_{X}[\omega,\iota,\phi,\mathbb{L}](G_2, G_1)$ since the Formulas 
\ref{math:Vertex-related cosine}--\ref{math:cosine Graph Similarity} are all symmetric.
\item \textit{Positivity:} 
Since we are considering only positive arc weights, it always holds that 
\begin{equation}
\cos_X[\omega,\iota,\phi,\mathbb{L}](G_1, G_1) \ge 0
\end{equation}
for any $\omega,\iota,\phi$ and $\mathbb{L}\not=\emptyset$.
\item \textit{Upper bound:} 
$\cos[\omega,\iota,\phi,\mathbb{L}](G_1, G_1) = 1$ for any $\omega,\iota,\phi$ and $\mathbb{L}\not=\emptyset$ and thus:
\[\forall G_2\not= G_1\!: \cos[\omega,\iota,\phi,\mathbb{L}](G_1, G_1) \ge \cos[\omega,\iota,\phi,\mathbb{L}](G_1, G_2)\]
\end{enumerate}

It is worth noticing that the range of values of Formula \ref{math:Vertex-related cosine} and of Formula \ref{math:Cosine of two Vertex Sets} is limited to $[0,1]$, since the values of $\gep$ are always positive and we only consider positive membership values of texts to topic nodes.

So far we looked at measures that mostly processed the arc set $A$ of TNs. 
This is contrasted by measures operating on topological indices of graphs.
An example is NetSimile \cite{Berlingerio:et:al:2013} which is based on the idea of characterizing networks by vectors of graph indices, which mostly draw on theories of social networks or egonets.
Starting from seven local, node-related structural features (e.g.\ node degree, node clustering, or size of a node's egonet\footnote{See Berlingerio et al.\ \cite{Berlingerio:et:al:2013} for the details of this approach.}), it computes the mean and the first four moments of the corresponding distributions to generate 35 dimensional feature vectors per network where the Canberra Distance is used to compute their distances:
let $\vec{x}, \vec{y}\in \mathbb{R}^k$ be two vectors, then their Canberra Distance is defined as 
\begin{equation}\label{math:NetSimile}
d_{\mathit{Can}}(\vec{x}, \vec{y}) = 
\sum_{i=1}^k \frac{|\vec{x}_i - \vec{y}_i|}{|\vec{x}_i + \vec{y}_i|}
\end{equation}
Soundarajan et al.\ \cite{Soundarajan:et:al:2014} show that NetSimile is consistently close to the consensus among all measures studied by them, showing that it approximates the results of more complex competitors. 
This finding makes NetSimile a first choice in any comparative study of graph similarities.

Following on from this success, we introduce a topology-related approach to graph similarity, which draws on the hierarchical classification of the texts underlying the topic networks by reference to the \textit{Dewey Decimal Classification} (DDC) (see Section \ref{sec:Module 2: Topic Classification}).
Starting from a pretest which essentially showed that graph invariants of complex network theory \cite{Newman:2003:a} do not sufficiently distinguish networks from their random counterparts, we decided to calculate a series of graph indices that evaluate the assignment of topics to the second level of the DDC.
More specifically, we compute three node type-sensitive variants of the four cluster coefficient $\Cws$ \cite{Watts:Strogatz:1998}, $\Cbr$ \cite{Bollobas:Riordan:2003}, $\Cbbpv$ \cite{Barrat:Barthelemy:Pastor-Satorras:Vespignani:2004:a} and 
$\Czh$ \cite{Zhang:Horvath:2005} \cite[cf.][]{Kalna:Higham:2006}.
This variation can be exemplified by means of $\Cws$:
to derive the desired variants from $\Cws$, we use the following scheme, where $\mathit{mode}\in \{\mathit{intra}, \mathit{inter}, \mathit{heter}\}$ serves as a parameter to distinguish these alternatives ($d_i$ is the degree of $v_i\in V$):
\begin{eqnarray}\label{math:ToSi}
\CwsX = \frac{1}{n}\sum_{i = 1}^{n} 2 \frac{\adjXi(v_i)}{d_i^2 - d_i} \in [0,1] 
\end{eqnarray}
$\adjintra(v_i)$ is the number of adjacent neighbors of $v_i\in V$ sharing their \nth{2}-level topic classification with $v_i$, $\adjinter(v_i)$ is the number of adjacent neighbors of $v_i$ whose identical classification differs from that of $v_i$ and $\adjheter(v_i)$ is the number of adjacent neighbors of $v_i$ whose classification differs among each other and from that of $v_i$.\footnote{A 4th case is that $v_i$ shares with a single neighbors its \nth{2}-level topic while differing from the topics of all other neighbors.}
In this way, we compute for each of the cluster values $\Cws$ (unweighted), $\Cbr$ (unweighted), $\Cbbpv$ (weighted), $\Czh$ (weighted) three variants considering intra- and interrelational as well as heterogeneous type-sensitive clustering so that topic networks are finally represented by 12-dimensional feature vectors which are compared using the cosine measure.
We call this approach \ToSi\ (as short for \textit{topological similarity}).

As a result of this candidate show of graph similarity measures we consider the set of measures displayed in Table \ref{tab:Graph Similarity Measures} for measuring the similarities of topic networks in order to shed light on Hypothesis \ref{hypo:Hypothesis 1}, part (2).

\subsubsection{Module 7 and 8: Machine Learning and Classification Analysis}\label{sec:Module 8 -- Machine Learning -- and Module 9 -- Classification Analysis}
We conduct experiments in supervised learning with the aim of training classifiers to detect the layer (TTN or ATN) to which a topic network of a MTN belongs and the genre of the corpus from which the underlying LMN is derived.
That is, our machine learning starts from a set of $n$ genres $\mathcal{G}_i$, $i = 1..n$, each of which is represented by a set $\mathbb{C}_{i} = \{ C_{ij}\mid j = 1..n_{i}\}$ of text corpora $C_{ij}$ (see Figure \ref{fig:Classification Scenarios}).
The set $\{ \mathbb{C}_{i} \mid i = 1..n \}$ defines a gold standard for which we assume that 
$\forall i, j = 1..n, i \not= j\!: \mathbb{C}_{i}\cap \mathbb{C}_{j} = \emptyset$.
Next, for each corpus $C_{ij}$ of each genre $\mathcal{G}_i$, we span an LMN $\mathcal{L}(C_{ij}, 2)$ that in turn is used to derive a two-layer MTN $\mathcal{M}(C_{ij}, 2) = (\mathbb{M}_{ij}, \mathbb{D}_{ij}) \mapsfrom C_{ij}$ such that $\mathbb{M}_{ij} = \{ M_{ij}, N_{ij} \}$ consists of exactly two topic networks: 
a TTN $M_{ij}$ and an ATN $N_{ij}$ both derived from $\mathcal{L}(C_{ij}, 2)$. 
In this way, we obtain the set $\mathbb{M}_{\mathit{ttn}}$ and the set $\mathbb{M}_{\mathit{atn}}$ of all TTNs and ATNs, respectively, both derived from $\mathcal{L}(C_{ij}, 2)$ according to Section \ref{sec:Module 3: Network Induction}.
Next, each of the sets $\mathbb{M}_{\mathit{ttn}}$ and $\mathbb{M}_{\mathit{atn}}$ is randomized according to the procedure described in Section \ref{sec:Module 4: Network Randomization} (Baseline B2).
In this way, we obtain the sets $\mathbb{M}'_{\mathit{ttn}}$ and $\mathbb{M}'_{\mathit{atn}}$ as the randomized counterparts of $\mathbb{M}_{\mathit{ttn}}$ and $\mathbb{M}_{\mathit{atn}}$.
As a result, we distinguish a range of classification experiments ({\small\setlength{\fboxsep}{0.5mm}\fbox{1}--\fbox{14}}) only a subset of which will be conducted in Section \ref{sec:Experimentation} to tackle Hypothesis \ref{hypo:Hypothesis 1}.
We start with distinguishing TTNs from ATNs.
The underlying classification hypothesis is:
\begin{hypothesis}\label{hypo:TTNs versus ATNs}
Topic networks of the same layer (also called mode) (i.e.\ TTN or ATN) are more similar than networks of different modes.\footnote{This concerns Scenario {\small\setlength{\fboxsep}{0.5mm}\fbox{1}} (observed data) and Scenario {\small\setlength{\fboxsep}{0.5mm}\fbox{6}} (randomized data) in Figure \ref{fig:Classification Scenarios}.}
\end{hypothesis}
The similarity of TNs will be quantified by means of the apparatus of Section \ref{sec:Module 6: Graph Similarity Analysis}.
Regardless of which genre (\textit{urban} vs.\ \textit{regional} vs.\ \textit{encyclopedic communication}) the underlying corpus belongs to, Hypothesis \ref{hypo:TTNs versus ATNs} assumes that one can always distinguish TTNs from ATNs by their structure, while TTNs and ATNs are less distinguishable among themselves.
This scenario is depicted in Figure \ref{fig:Classification Scenarios} by Arrow {\small\setlength{\fboxsep}{0.5mm}\fbox{1}}.
If we falsify the alternative to this hypothesis, we can assume that (poor, rich or moderate) thematic intertextuality, as manifested by TTNs, is different form co-authorship-based networking of topics in ATNs. 
Collaboration- and intertextuality-based networking would then differ in a way that characterizes their layer.
In order to test genre sensitivity as disregarded by Hypothesis \ref{hypo:TTNs versus ATNs}, we carry out two experiments: 
one in which we classify TTNs (ATNs) by genre and one in which we combine both classifications by simultaneously classifying by genre and layer.
When classifying by genre, we distinguish TNs derived from city wikis (\textit{urban communication}), regional wikis (\textit{regional communication}) and from subnetworks of Wikipedia (\textit{knowledge communication}) (see Section \ref{sec:Module 2: Topic Classification}). 
Finally, we generate two control classes of wikis and Wikipedia-based networks outside of these three genres. 
The corresponding wikis are sampled in a way that their members are rather dissimilar. 
Our similarity measurement should therefore not work with them.\label{page:Control CLasses}
In a nutshell, the underlying classification hypothesis is:

\begin{figure}[t]
	\includegraphics[width=0.95\textwidth]{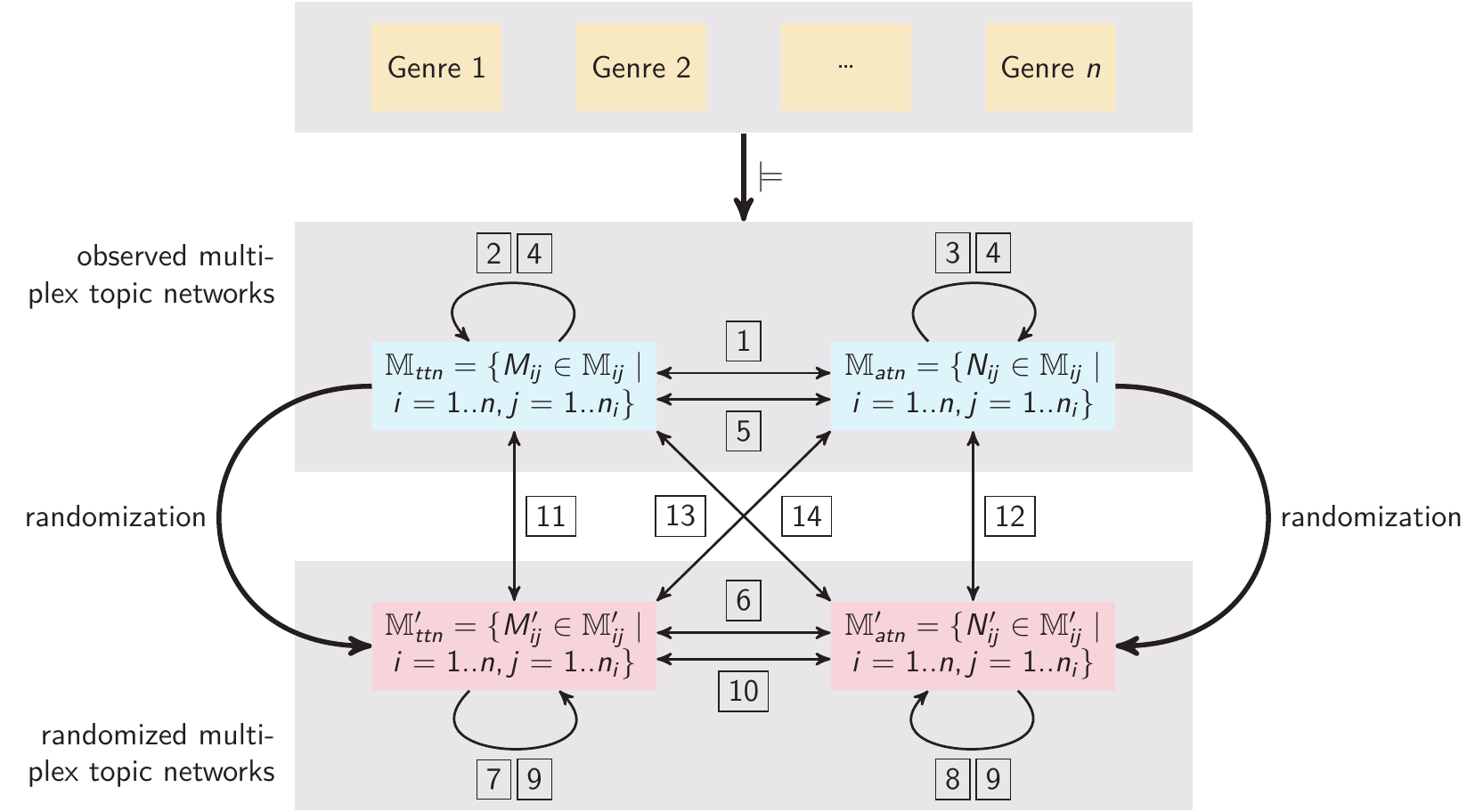}
	\caption{From sets of corpora of different genres to multiplex topic networks and their randomizations:
	corpora of different genres are the starting point for spanning LMNs which are then used derive two-layer multiplex topic networks ($\models$). 
	In a second step, randomized counterparts according to section \ref{sec:Module 4: Network Randomization} are derived from these MTNs to obtain a further basis for evaluating their significance.
	In this way we arrive at fourteen candidate scenarios for classifying topic networks.
	}
	\label{fig:Classification Scenarios}
\end{figure}

\begin{hypothesis}\label{hypo:Genre-oriented Classification}
Topic networks of the same genre are more similar than those of different genres.\footnote{This concerns the scenarios 
{\small\setlength{\fboxsep}{0.5mm}\fbox{2}}, 
{\small\setlength{\fboxsep}{0.5mm}\fbox{3}} and 
{\small\setlength{\fboxsep}{0.5mm}\fbox{4}} (observed) and the scenarios {\small\setlength{\fboxsep}{0.5mm}\fbox{7}}, 
{\small\setlength{\fboxsep}{0.5mm}\fbox{8}} and 
{\small\setlength{\fboxsep}{0.5mm}\fbox{9}} (random data) in Figure \ref{fig:Classification Scenarios}.}
\end{hypothesis}
As we consider the genre-sensitive classification in the context of the layer-sensitive one, we get different classification scenarios:
\begin{enumerate}
\item Scenario {\setlength{\fboxsep}{0.5mm}\fbox{2}} in Figure \ref{fig:Classification Scenarios} denotes the task of training a classifier that detects TTNs of the same genre while distinguishing TTNs of different ones.
If this is successful, we can assume that the TTNs analyzed here are genre-sensitive or that the communication functions that we hypothetically associate with these genres influence the structure of these TTNs.
\item Scenario {\setlength{\fboxsep}{0.5mm}\fbox{3}} from Figure \ref{fig:Classification Scenarios} regards the analog experiment for the genre-sensitive classification of ATNs. 

\item Scenario {\setlength{\fboxsep}{0.5mm}\fbox{4}} concerns the alternative in which the modal difference of TTNs and ATNs is ignored in order to classify topic networks independently of their modal difference according to their underlying genre.

\item This scenario is contrasted with Scenario {\setlength{\fboxsep}{0.5mm}\fbox{5}}, which considers classifiers for simultaneously detecting the genre and layer of TNs.
The underlying classification hypothesis is:
\end{enumerate}

\begin{hypothesis}\label{hypo:Genre-and Mode-oriented Classification}
Topic networks of the same layer and genre are more similar than networks of different layers or genres.\footnote{This concerns Scenario {\small\setlength{\fboxsep}{0.5mm}\fbox{5}} (observed data) and Scenario {\small\setlength{\fboxsep}{0.5mm}\fbox{10}} (random data) in Figure \ref{fig:Classification Scenarios}.}
\end{hypothesis}

Falsifying the alternative to part (2) of Hypothesis \ref{hypo:Hypothesis 1} implies that TNs derived from corpora written by different communities by addressing different thematic frames (e.g.\ cities) appear nevertheless similar in their gestalt.
Such a finding is very unlikely in cases in which the underlying corpora serve very different communication functions: 
Hypothesis \ref{hypo:Hypothesis 1} is not saying that everything is similar irrespective of the heterogeneity of the underlying function or the thematic orientation.
Thus, a genre-oriented classification that shows that TNs of the same genre (serving a certain communication function and having a certain thematic orientation), are more similar than those belonging to different genres, would rather correspond to such a finding.
From this point of view, Hypothesis \ref{hypo:Genre-oriented Classification} and \ref{hypo:Genre-and Mode-oriented Classification} are of interest:
to deal with them experimentally could pave the way for testing the second part (2) of Hypothesis \ref{hypo:Hypothesis 1}.

As explained in Section \ref{sec:Module 4: Network Randomization}, we randomize input networks so that we obtain five additional classification scenarios labeled {\setlength{\fboxsep}{0.5mm}\fbox{6}}--{\setlength{\fboxsep}{0.5mm}\fbox{10}} in Figure \ref{fig:Classification Scenarios}.
The experiments corresponding to these scenarios will be conducted here, as far as they concern the baseline scenario B2 of Section \ref{sec:Module 4: Network Randomization}.
Furthermore, scenarios are to be enumerated which attempt to distinguish observed networks directly from their randomized counterparts.
In this context, Scenario {\setlength{\fboxsep}{0.5mm}\fbox{11}} aims at distinguishing TTNs from their randomized counterparts by means of the classifiers trained to detect TTNs.
Analogously, Scenario {\setlength{\fboxsep}{0.5mm}\fbox{12}} considers ATNs in relation to their randomized counterparts, while 
Scenario {\setlength{\fboxsep}{0.5mm}\fbox{13}} aims to separate observed topic networks (whether ATNs or TTNs) from randomized ones. 
Finally, Scenario {\setlength{\fboxsep}{0.5mm}\fbox{14}} extends the latter scenario by trying to additionally account for the modal difference of ATNs and TTNs.
These scenarios are only listed for theoretical reasons.

\section{Experimentation}\label{sec:Experimentation}

To test Hypothesis \ref{hypo:Hypothesis 1} and its relatives (i.e.\ Hypothesis \ref{hypo:TTNs versus ATNs}, \ref{hypo:Genre-oriented Classification} and \ref{hypo:Genre-and Mode-oriented Classification}), we conduct several experiments using two resources: 
a corpus of special wikis, called the \textit{Frankfurt Regional Wiki Corpus}, and a corpus of subnetworks of Wikipedia that mostly contain information about cities and regions.

\subsection{Tools and Resources}\label{sec:resources}
The \textit{Frankfurt Regional Wiki Corpus} (FRWC) contains 43 wikis collected from online wiki lists.\footnote{E.g.\ \url{https://de.wikipedia.org/wiki/Regiowiki}}
Table~\ref{table:general_wiki_stats} shows the statistics of this corpus, which is divided into three genres:
\Cities relates to wikis describing certain cities, \Regions includes wikis focusing on a specific region, while the residual class \Others collects wikis that are not off-topic w.r.t.\ regional communication, but are unusual in their structure or the described rhemes.
We consider only articles that are not redirects.
Wiki authors use redirect pages to lead readers of articles with outdated, incorrect or alternative spelling titles to the desired target page.
We remove all such redirects and rewire all affected links accordingly.
As a result, the number of processed articles is smaller than their overall number (see Table~\ref{table:general_wiki_stats}).
In addition to the FRWC we extracted a corpus of Wikipedia subgraphs (see Section \ref{sec:Module 2: Topic Classification} for the formal definition of these graphs and Table~\ref{table:general_wikisubgraph_stats} for the corpus statistics).
Subsequently, we denote the two variants in this Wikipedia corpus \WPRegioOne and \WPRegioTwo.
We choose 25 articles about cities or regions matching the titles of the wikis in the FRWC and additionally include the subgraphs of six off-topic articles to build two additional corpora, called \WPOthersOne and \WPOthersTwo, for purposes of comparison.
\nprounddigits{0}
\begin{table}[!htb]  
	\hspace*{-1.75cm}
	\resizebox{0.75\textwidth}{!}{
		\begin{minipage}[b]{6.0cm}
		\begin{tikzpicture}
	[Kasten/.style={rectangle, minimum height=0.40cm, anchor=west}, Path/.style={font={\small\vphantom{Ag}}, midway, above},]
	\matrix [matrix of nodes, nodes=Kasten, row sep=0.03cm, column sep=0.03cm, font={\scriptsize\vphantom{Ag}}, 
	row 1/.append style={nodes={fill=SeminarSehrHellBlau}},	
	column 1/.append style={nodes={text width=1.5cm, align=left, fill=SeminarSehrSehrHellGrau}}, 
	column 2/.append style={nodes={text width=0.8cm, align=left, fill=SeminarSehrSehrHellGrau}},
	column 3/.append style={nodes={text width=0.8cm, align=right, fill=SeminarSehrSehrHellGrau}},
	column 4/.append style={nodes={text width=0.8cm, align=right, fill=SeminarSehrSehrHellGrau}},
	column 5/.append style={nodes={text width=0.8cm, align=right, fill=SeminarSehrSehrHellGrau}},
	column 6/.append style={nodes={text width=0.8cm, align=right, fill=SeminarSehrSehrHellGrau}},
	%
	row 28/.append style={nodes={fill=SeminarSehrHellGrau}},
	row 29/.append style={nodes={fill=SeminarSehrHellGrau}},
	row 30/.append style={nodes={fill=SeminarSehrHellGrau}},
	row 31/.append style={nodes={fill=SeminarSehrHellGrau}},
	row 32/.append style={nodes={fill=SeminarSehrHellGrau}},
	row 33/.append style={nodes={fill=SeminarSehrHellGrau}},
	row 34/.append style={nodes={fill=SeminarSehrHellGrau}},
	row 35/.append style={nodes={fill=SeminarSehrHellGrau}},
	row 36/.append style={nodes={fill=SeminarSehrHellGrau}},
	row 37/.append style={nodes={fill=SeminarSehrHellGrau}},
	row 38/.append style={nodes={fill=SeminarSehrHellGrau}},
	row 39/.append style={nodes={fill=SeminarSehrHellGrau}}
	] (Link) {
Wiki & \#art. 1 & \#art. 2 & \#rev. & \#authors \\
Baden-Baden & \numprint{999} & \numprint{844} & \numprint{3576} & \numprint{138} \\
Boppard & \numprint{24} & \numprint{23} & \numprint{107} & \numprint{17} \\
Cuxhaven & \numprint{2884} & \numprint{2722} & \numprint{28284} & \numprint{619} \\
Dresden & \numprint{11479} & \numprint{9796} & \numprint{76776} & \numprint{2702} \\
Erfurt & \numprint{2275} & \numprint{2267} & \numprint{30314} & \numprint{129} \\
Esslingen & \numprint{252} & \numprint{219} & \numprint{2646} & \numprint{353} \\
Fürth & \numprint{9686} & \numprint{8055} & \numprint{109467} & \numprint{2546} \\
Görlitz & \numprint{1897} & \numprint{1735} & \numprint{11412} & \numprint{555} \\
Hamm & \numprint{16602} & \numprint{14439} & \numprint{99307} & \numprint{1353} \\
Karlsruhe & \numprint{38870} & \numprint{25575} & \numprint{306143} & \numprint{11002} \\
Köln & \numprint{3925} & \numprint{3184} & \numprint{13394} & \numprint{400} \\
Linz & \numprint{6776} & \numprint{4250} & \numprint{28923} & \numprint{343} \\
Lüneburg & \numprint{105} & \numprint{96} & \numprint{422} & \numprint{108} \\
Lustenau & \numprint{812} & \numprint{553} & \numprint{3185} & \numprint{241} \\
München & \numprint{20344} & \numprint{15829} & \numprint{111681} & \numprint{8016} \\
Münster & \numprint{4096} & \numprint{3703} & \numprint{24226} & \numprint{984} \\
Olsberg & \numprint{376} & \numprint{360} & \numprint{2403} & \numprint{140} \\
Reutlingen & \numprint{583} & \numprint{545} & \numprint{3122} & \numprint{368} \\
Schiltach & \numprint{505} & \numprint{489} & \numprint{560} & \numprint{14} \\
Schorndorf & \numprint{1035} & \numprint{1005} & \numprint{4778} & \numprint{73} \\
Strausberg & \numprint{3906} & \numprint{3668} & \numprint{12860} & \numprint{111} \\
Stuttgart & \numprint{1260} & \numprint{1076} & \numprint{6784} & \numprint{228} \\
Tübingen & \numprint{4749} & \numprint{4211} & \numprint{38540} & \numprint{1513} \\
Weißenburg & \numprint{436} & \numprint{393} & \numprint{5436} & \numprint{63} \\
Wulfen & \numprint{746} & \numprint{722} & \numprint{23218} & \numprint{767} \\
Würzburg & \numprint{22432} & \numprint{17661} & \numprint{283773} & \numprint{2726}  \\
};
	\end{tikzpicture}
		\end{minipage}
		\hspace*{0.3cm}
		\begin{minipage}[b]{5cm}
		\begin{tikzpicture}
[Kasten/.style={rectangle, minimum height=0.40cm, anchor=west}, Path/.style={font={\small\vphantom{Ag}}, midway, above},]
\matrix [matrix of nodes, nodes=Kasten, row sep=0.03cm, column sep=0.03cm, font={\scriptsize\vphantom{Ag}}, 
row 1/.append style={nodes={fill=SeminarSehrHellBlau}},	
column 1/.append style={nodes={text width=1.8cm, align=left, fill=SeminarSehrSehrHellGrau}}, 
column 2/.append style={nodes={text width=0.8cm, align=left, fill=SeminarSehrSehrHellGrau}},
column 3/.append style={nodes={text width=0.8cm, align=right, fill=SeminarSehrSehrHellGrau}},
column 4/.append style={nodes={text width=0.8cm, align=right, fill=SeminarSehrSehrHellGrau}},
column 5/.append style={nodes={text width=0.8cm, align=right, fill=SeminarSehrSehrHellGrau}},
column 6/.append style={nodes={text width=0.8cm, align=right, fill=SeminarSehrSehrHellGrau}},
%
row 28/.append style={nodes={fill=SeminarSehrHellGrau}},
row 29/.append style={nodes={fill=SeminarSehrHellGrau}},
row 30/.append style={nodes={fill=SeminarSehrHellGrau}},
row 31/.append style={nodes={fill=SeminarSehrHellGrau}},
row 32/.append style={nodes={fill=SeminarSehrHellGrau}},
row 33/.append style={nodes={fill=SeminarSehrHellGrau}},
row 34/.append style={nodes={fill=SeminarSehrHellGrau}},
row 35/.append style={nodes={fill=SeminarSehrHellGrau}},
row 36/.append style={nodes={fill=SeminarSehrHellGrau}},
row 37/.append style={nodes={fill=SeminarSehrHellGrau}},
row 38/.append style={nodes={fill=SeminarSehrHellGrau}},
row 39/.append style={nodes={fill=SeminarSehrHellGrau}}
] (Link) {
Wiki & \#art. 1 & \#art. 2 & \#rev. & \#authors \\
Ahrweiler & \numprint{24194} & \numprint{22814} & \numprint{149345} & \numprint{690} \\
Attersee/Attergau & \numprint{922} & \numprint{813} & \numprint{17944} & \numprint{53} \\
Dithmarschen & \numprint{2155} & \numprint{1712} & \numprint{29981} & \numprint{185} \\
Ennstal & \numprint{12774} & \numprint{11936} & \numprint{76721} & \numprint{135} \\
Franken & \numprint{5511} & \numprint{4510} & \numprint{78371} & \numprint{887} \\
Göttingen & \numprint{8695} & \numprint{7755} & \numprint{36393} & \numprint{488} \\
Niederbayern & \numprint{33751} & \numprint{20504} & \numprint{196525} & \numprint{1392} \\
Pforzheim-Enz & \numprint{14763} & \numprint{12821} & \numprint{67604} & \numprint{3213} \\
Rhein-Main & \numprint{5276} & \numprint{2801} & \numprint{17290} & \numprint{40} \\
Rhein-Neckar & \numprint{12241} & \numprint{10413} & \numprint{62830} & \numprint{2807} \\
Sachenanhalt & \numprint{4644} & \numprint{4173} & \numprint{36264} & \numprint{1153} \\
Waldviertel & \numprint{266} & \numprint{264} & \numprint{1906} & \numprint{124} \\
};
\end{tikzpicture}

\begin{tikzpicture}
[Kasten/.style={rectangle, minimum height=0.40cm, anchor=west}, Path/.style={font={\small\vphantom{Ag}}, midway, above},]
\matrix [matrix of nodes, nodes=Kasten, row sep=0.03cm, column sep=0.03cm, font={\scriptsize\vphantom{Ag}}, 
row 1/.append style={nodes={fill=SeminarSehrHellBlau}},	
column 1/.append style={nodes={text width=1.8cm, align=left, fill=SeminarSehrSehrHellGrau}}, 
column 2/.append style={nodes={text width=0.8cm, align=left, fill=SeminarSehrSehrHellGrau}},
column 3/.append style={nodes={text width=0.8cm, align=right, fill=SeminarSehrSehrHellGrau}},
column 4/.append style={nodes={text width=0.8cm, align=right, fill=SeminarSehrSehrHellGrau}},
column 5/.append style={nodes={text width=0.8cm, align=right, fill=SeminarSehrSehrHellGrau}},
column 6/.append style={nodes={text width=0.8cm, align=right, fill=SeminarSehrSehrHellGrau}},
%
row 28/.append style={nodes={fill=SeminarSehrHellGrau}},
row 29/.append style={nodes={fill=SeminarSehrHellGrau}},
row 30/.append style={nodes={fill=SeminarSehrHellGrau}},
row 31/.append style={nodes={fill=SeminarSehrHellGrau}},
row 32/.append style={nodes={fill=SeminarSehrHellGrau}},
row 33/.append style={nodes={fill=SeminarSehrHellGrau}},
row 34/.append style={nodes={fill=SeminarSehrHellGrau}},
row 35/.append style={nodes={fill=SeminarSehrHellGrau}},
row 36/.append style={nodes={fill=SeminarSehrHellGrau}},
row 37/.append style={nodes={fill=SeminarSehrHellGrau}},
row 38/.append style={nodes={fill=SeminarSehrHellGrau}},
row 39/.append style={nodes={fill=SeminarSehrHellGrau}}
] (Link) {
	Wiki & \#art. 1 & \#art. 2 & \#rev. & \#authors \\
	Graz & \numprint{10226} & \numprint{9436} & \numprint{35490} & \numprint{32} \\
	RegioWikiAT & \numprint{12085} & \numprint{8551} & \numprint{113436} & \numprint{3221} \\
	Wallis & \numprint{3174} & \numprint{3149} & \numprint{18054} & \numprint{86} \\
	Wetzikon & \numprint{1737} & \numprint{1302} & \numprint{23999} & \numprint{446} \\
	Wien-Geschichte & \numprint{45473} & \numprint{43919} & \numprint{296467} & \numprint{402} \\
};
\end{tikzpicture}
		\end{minipage}
	}
	\caption{Statistics of the FRWC showing the number of articles with (\#art.\ 1) and without (\#art.\ 2) redirects, the number of revisions (\#rev.) and the number of distinct authors (\#authors). The last three columns disregard redirecting articles. Left table: genre \Cities; upper right: genre \Regions; lower right: genre \Others.
	The German, Austrian and Swiss wikis were downloaded in early 2018.
	}
	\label{table:general_wiki_stats}
\end{table}

\nprounddigits{0}
\begin{table}[!htb]  
	\centering
	\resizebox{0.85\textwidth}{!}{
		\begin{tikzpicture}
	[Kasten/.style={rectangle, minimum height=0.30cm, anchor=west}, Path/.style={font={\small\vphantom{Ag}}, midway, above},]
	\matrix [matrix of nodes, nodes=Kasten, row sep=0.03cm, column sep=0.03cm, font={\scriptsize\vphantom{Ag}}, 
	row 1/.append style={nodes={fill=SeminarSehrHellBlau}},	
	column 1/.append style={nodes={text width=0.4cm, align=right, fill=SeminarSehrSehrHellGrau}}, 
	column 2/.append style={nodes={text width=3.2cm, align=left, fill=SeminarSehrSehrHellGrau}}, 
	column 3/.append style={nodes={text width=1.3cm, align=right, fill=SeminarSehrSehrHellGrau}},
	column 4/.append style={nodes={text width=1.3cm, align=right, fill=SeminarSehrSehrHellGrau}},
	column 5/.append style={nodes={text width=1.3cm, align=right, fill=SeminarSehrSehrHellGrau}},
	column 6/.append style={nodes={text width=1.3cm, align=right, fill=SeminarSehrSehrHellGrau}},
	column 7/.append style={nodes={text width=1.3cm, align=right, fill=SeminarSehrSehrHellGrau}},
	column 8/.append style={nodes={text width=1.3cm, align=right, fill=SeminarSehrSehrHellGrau}},
	row 25/.append style={nodes={minimum height=0.50cm}}
	row 26/.append style={nodes={minimum height=0.50cm}},
	row 27/.append style={nodes={minimum height=0.50cm}},
	row 28/.append style={nodes={minimum height=0.30cm}},
	row 20/.append style={nodes={minimum height=0.30cm}},
	row 30/.append style={nodes={minimum height=0.30cm}}
	row 31/.append style={nodes={minimum height=0.30cm}}
	] (Link) {
& Seed Article & \#articles 1 & \#revisons 1 & \#authors 1 & \#articles 2 & \#revisons 2 & \#authors 2 \\
1. & Ahrweiler & \numprint{90} & \numprint{66217} & \numprint{16772} & \numprint{11413} & \numprint{5602327} & \numprint{930621} \\
2. & 
Dithmarschen & \numprint{210} & \numprint{156862} & \numprint{38180} & \numprint{30386} & \numprint{10006785} & \numprint{1506634} \\
3. & 
Dresden & \numprint{1615} & \numprint{1180743} & \numprint{239747} & \numprint{127675} & \numprint{27746644} & \numprint{3566957} \\
4. & 
Erfurt & \numprint{943} & \numprint{850786} & \numprint{179282} & \numprint{100052} & \numprint{23644822} & \numprint{3158299} \\
5. & 
Fürth & \numprint{504} & \numprint{598687} & \numprint{130445} & \numprint{77663} & \numprint{19481686} & \numprint{2657440} \\
6. & 
Görlitz & \numprint{790} & \numprint{468641} & \numprint{99606} & \numprint{62896} & \numprint{17305177} & \numprint{2431331} \\
7. & 
Göttingen & \numprint{922} & \numprint{786663} & \numprint{170082} & \numprint{93726} & \numprint{22448816} & \numprint{2995497} \\
8. & 
Hamm & \numprint{764} & \numprint{697437} & \numprint{150502} & \numprint{82099} & \numprint{20436567} & \numprint{2799384} \\
9. & 
Karlsruhe & \numprint{1021} & \numprint{842723} & \numprint{180652} & \numprint{97484} & \numprint{23178185} & \numprint{3103192} \\
10. & 
Köln & \numprint{1485} & \numprint{1090676} & \numprint{223801} & \numprint{122446} & \numprint{26851098} & \numprint{3483785} \\
11. & 
Linz & \numprint{816} & \numprint{602346} & \numprint{130520} & \numprint{79376} & \numprint{20188052} & \numprint{2792374} \\
12. & 
Metropolregion Rhein-Neckar & \numprint{296} & \numprint{157356} & \numprint{37960} & \numprint{23250} & \numprint{8608771} & \numprint{1388939} \\
13. & 
München & \numprint{1421} & \numprint{1077626} & \numprint{216774} & \numprint{120725} & \numprint{26727317} & \numprint{3472725} \\
14. & 
Münster & \numprint{1139} & \numprint{894916} & \numprint{193090} & \numprint{103436} & \numprint{24330809} & \numprint{3251427} \\
15. & 
Niederbayern & \numprint{239} & \numprint{142392} & \numprint{33551} & \numprint{22466} & \numprint{7796961} & \numprint{1222744} \\
16. & 
Rhein-Main-Gebiet & \numprint{390} & \numprint{297276} & \numprint{65804} & \numprint{42238} & \numprint{12750028} & \numprint{1870354} \\
17. & 
Sachsen-Anhalt & \numprint{603} & \numprint{459933} & \numprint{96116} & \numprint{59565} & \numprint{16392237} & \numprint{2291304} \\
18. & 
Schorndorf & \numprint{362} & \numprint{226153} & \numprint{51264} & \numprint{32562} & \numprint{11738799} & \numprint{1746169} \\
19. & 
Steirisches Ennstal & \numprint{43} & \numprint{19702} & \numprint{6322} & \numprint{4400} & \numprint{2101467} & \numprint{386487} \\
20. & 
Strausberg & \numprint{265} & \numprint{215854} & \numprint{49617} & \numprint{30284} & \numprint{10602198} & \numprint{1579390} \\
21. & 
Stuttgart & \numprint{1317} & \numprint{1089313} & \numprint{215788} & \numprint{123906} & \numprint{26648581} & \numprint{3403376} \\
22. & 
Tübingen & \numprint{623} & \numprint{385288} & \numprint{85266} & \numprint{54525} & \numprint{15884637} & \numprint{2265358} \\
23. & 
Wetzikon & \numprint{204} & \numprint{145207} & \numprint{33914} & \numprint{20607} & \numprint{8044399} & \numprint{1306780} \\
24. & 
Wien & \numprint{1380} & \numprint{874419} & \numprint{170952} & \numprint{102792} & \numprint{23357095} & \numprint{3087254} \\
25. & 
Würzburg & \numprint{959} & \numprint{885109} & \numprint{185495} & \numprint{106381} & \numprint{24484274} & \numprint{3216674} \\
26. & 
Hydraulik &\numprint{121} & \numprint{59874} & \numprint{19400} & \numprint{8287} & \numprint{3600636} & \numprint{700341} \\
27. & 
Integralrechnung & \numprint{194} & \numprint{75082} & \numprint{21787} & \numprint{6708} & \numprint{2663563} & \numprint{508606} \\
28. & 
Kernkraftwerk & \numprint{287} & \numprint{196202} & \numprint{49279} & \numprint{20773} & \numprint{8195232} & \numprint{1387491} \\
29. & 
Neuronales Netz & \numprint{85} & \numprint{27878} & \numprint{9750} & \numprint{3739} & \numprint{1488680} & \numprint{332714} \\
30. & 
Schlacht bei Waterloo & \numprint{200} & \numprint{97290} & \numprint{25614} & \numprint{18674} & \numprint{6990403} & \numprint{1097749} \\
31. & 
Zecken & \numprint{112} & \numprint{58582} & \numprint{16350} & \numprint{7500} & \numprint{3896913} & \numprint{734269} \\
};
\end{tikzpicture}
	}
	\caption{Wikipedia-based corpora: 
	number of content articles (\#articles $n$), revisions (\#revisions $n$) and authors (\#authors $n$) of non-redirecting articles in \WPRegioOne ($n=1$) and \WPRegioTwo ($n=2$) of the German Wikipedia dump from 2018-07-01 (subgraphs 1-25); the variable $n$ codes the $n$th orbit (see Formula \ref{math:Orbit n-graph}).
	Subgraphs 26-31 are used to generate the corpora \WPOthersOne and \WPOthersTwo.
	}
	\label{table:general_wikisubgraph_stats}
\end{table}

We process the content, link structure and meta data (e.g.\ authorship-related information) of all articles in our corpora.
This includes their history, that is, the chains of revisions which led to their current state.
We do not consider past states of link structure and content itself but incorporate the authorship and the amount of content being added or removed per revision (see Section \ref{sec:Module 3: Network Induction}).
The wikis considered here are based on MediaWiki.
The structure of their articles varies from wiki to wiki, so that HTML-based extractions are error-prone.
To circumvent this problem, we use WikiDragon \cite{Gleim:Mehler:Song:2018}, a Java-based framework for importing and processing wikis offline.

For our experiments we used, adapted and newly developed several tools including the so-called \textit{GeneticClassifierWorkbench} (GCW), a Python library for performing feature selections and sensitivity analyses in classification experiments.
Since our experiments are based on feature vectors with a size of sometimes more than 100 features, a complete sensitivity analysis of all feature combinations was not possible.
Therefore, we conducted a genetic search for the best performing subset of features due to maximizing the $F$-score.
That is, a population of $p$ features is evaluated and mutated over a number of $t$ rounds.
Instances which score best are saved unchanged for the next round and partly added in a slightly mutated form.
The worst performing instances are removed and replaced by random feature combinations.
The Workbench is based on the Python library \textit{scikit-learn} \cite{Pedregosa:et:al:2011} allowing us to abstract from the underlying machine learning paradigm so that the same genetic search can be applied to optimize different classifiers.
We experimented with neural networks which produced similar results on our test data, but took too much time to be used for genetic searches and random baseline computations.
Therefore, we decided for \textit{Support Vector Machine}s (SVM) as the embedded method of supervised learning using the \textit{Radial Basis Function} (RBF) as a kernel.
Our source code is open source on GitHub (\url{https://github.com/texttechnologylab/GeneticClassifierWorkbench}).

\subsection{Classification Experiments}
We investigate the similarities of our seven corpora of regional wikis (\Cities, \Regions and \Others) and of Wikipedia-based subgraphs (\WPRegioOne, \WPRegioTwo, \WPOthersOne and \WPOthersTwo) (each defining a corpus of texts) in order to test Hypothesis \ref{hypo:Hypothesis 1} and its derivatives, that is, Hypothesis \ref{hypo:TTNs versus ATNs}, \ref{hypo:Genre-oriented Classification} and \ref{hypo:Genre-and Mode-oriented Classification}.
Thus, we distinguish up to seven target classes in our experiments.
For reasons of simplicity, we call each element of these corpora  \textit{wiki} and each of the seven classes \textit{genre}.
Unless otherwise stated, the experiments are performed on all of them.
In the case of \WPRegioTwo and \WPOthersTwo, we did not induce the corresponding ATNs, as some of these would have included several million edit events.
Thus, in this case we have at most five target classes.
Each experiment includes three consecutive steps:
\begin{enumerate}
\item \textit{The \emph{all} variant:} 
The first step, denoted by \textit{all}, is a hyperplane parameter optimization and evaluation using the entire feature set. 
The optimized parameters of the respective classifier are then used in subsequent steps.
Ideally, the parameters are optimized independently for each step, but this would have slowed down the genetic search.
\item \textit{The \emph{opt} variant:} 
In the \nth{2} step, denoted by \textit{opt}, genetic searches for optimal feature subsets are performed using a population of 20 feature vector instances and 50 rounds, trying to maximize the $F$-score of the classification. 
Note that these searches may only reach a local maximum.
\item \textit{The \emph{ext} variant:} 
For experiments which are not conducted on random baseline data, we perform an extended genetic search for optimal feature subsets based on 20 instances and 500 rounds. 
In an additional step, a bit-wise genetic optimization attempts to further minimize the number of used features while keeping or even improving the $F$-score, using 20 instances and 500 rounds.
\end{enumerate}

\subsubsection{Graph-Similarity based classification}
Using the apparatus of Section \ref{sec:Module 6: Graph Similarity Analysis}, each TN (ATN or TTN) of each MTN is represented by a vector of values indicating its similarities to the wikis of the underlying experiment.
Any such vector is separately computed for each of the 11 similarity measures of Table \ref{tab:Graph Similarity Measures}.
Thus, if $\mathbb{T}$ is the set of all TNs of whatever mode (ATN or TTN) and genre (\Cities, \Regions etc.) and if $\mathbb{T}'\subseteq \mathbb{T}$ is a subset of these TNs used in a classification experiment concerning the genres (target classes) $\text{Genre } i_1,\ldots \text{Genre } i_{j}$ (c.f.\ Figure \ref{fig:Classification Scenarios}), then each topic network $T\in \mathbb{T}'$ is represented for each similarity measure by a $|\mathbb{T}'|$-dimensional feature vector which is processed by the three-step algorithm described above.
If for a given similarity measure the topic networks derived from wikis of the same genre are mapped to neighboring similarity vectors, then they belong to overlapping neighborhoods in vector space: 
\textit{related networks are similar in their similarity and dissimilarity relations.}
In this way, TNs of the same genre should become as recognizable as TNs of different genres. 
Now we see why a genetic search for optimal subsets of features is necessary:
the reason is that otherwise we would assume that all dimensions of our feature vectors are equally informative -- an assumption that is probably wrong.

\begin{table}[t]
	\centering
	\nprounddigits{3}
	\begin{tikzpicture}
	[Kasten/.style={rectangle, minimum height=0.60cm, anchor=west}, Path/.style={font={\footnotesize\vphantom{Ag}}, midway, above},]	
	\matrix [matrix of nodes, nodes=Kasten, row sep=0.05cm, column sep=0.05cm, font={\footnotesize\vphantom{Ag}}, 
	row 1/.append style={nodes={fill=SeminarSehrHellBlau}},
	%
	column 1/.append style={nodes={text width=0.30cm, align=right, fill=SeminarSehrHellBlau}},
	column 2/.append style={nodes={text width=2.8cm, align=left, fill=SeminarSehrSehrHellGrau}},
	column 3/.append style={nodes={text width=0.6cm, align=right, fill=SeminarSehrHellGrau}},
	column 4/.append style={nodes={text width=0.7cm, align=right, fill=SeminarSehrHellGrau}},
	column 5/.append style={nodes={text width=0.65cm, align=right, fill=SeminarSehrHellGrau}},
	column 6/.append style={nodes={text width=0.6cm, align=right, fill=SeminarSehrSehrHellGrau}},
	column 7/.append style={nodes={text width=0.7cm, align=right, fill=SeminarSehrSehrHellGrau}},
	column 8/.append style={nodes={text width=0.8cm, align=right, fill=SeminarSehrSehrHellGrau}},
	column 9/.append style={nodes={text width=0.7cm, align=right, fill=SeminarSehrSehrHellGrau}},
	column 10/.append style={nodes={text width=0.8cm, align=right, fill=SeminarSehrSehrHellGrau}},
	] (Link) { 
 & Measure & all & opt & ext & B1 & B3 all & B3 opt & B4 all & B4 opt \\ 
1. & $\ges$ & \numprint{0.6534404106470174} & \numprint{0.7525813682116204} & \numprint{0.7979984030560783} & \numprint{0.14279573663003836} & \numprint{0.13038814952295522} & \numprint{0.28648049747499443} & \numprint{0.12051556127052013} & \numprint{0.21273233579084075} \\
2. & $\wal$ & \numprint{0.6486374549819928} & \numprint{0.7508649675215026} & \numprint{0.7880437643273304} & \numprint{0.14279573663003836} & \numprint{0.13038814952295522} & \numprint{0.28648049747499443} & \numprint{0.10891658212278733} & \numprint{0.21629737902123555} \\
3. & $\veo$ & \numprint{0.6770113509884829} & \numprint{0.7726911150734933} & \numprint{0.8164697006835121} & \numprint{0.14279573663003836} & \numprint{0.13038814952295522} & \numprint{0.28648049747499443} & \numprint{0.1204902372234965} & \numprint{0.22080064106411723} \\
4. & $\wges$ &	\numprint{0.5594083467552856} & \numprint{0.6202807339092361} & \numprint{0.6501141176882417} & \numprint{0.14279573663003836} & \numprint{0.13038814952295522} & \numprint{0.28648049747499443} & \numprint{0.12012229508186045} & \numprint{0.198734804408144} \\
5. & $\cos_{\mathcal{A}}[\text{w},\infty,\phi_1,\mathbb{L}_{12}]$ & \numprint{0.6383933887034438} & \numprint{0.7223274764451235} & \numprint{0.7635285163016255} & \numprint{0.14279573663003836} & \numprint{0.13038814952295522} & \numprint{0.28648049747499443} & \numprint{0.11886268111858804} & \numprint{0.21065184993651465} \\
6. & $\cos_{\mathcal{A}\mathcal{V}}[\text{w},\infty,\phi_1,\mathbb{L}_{12}]$ & \numprint{0.7289939457564402} & \numprint{0.768134623996693} & \textbf{\numprint{0.853066585485869}} & \numprint{0.14279573663003836} & \numprint{0.13038814952295522} & \numprint{0.28648049747499443} & \numprint{0.12496166851979694} & \numprint{0.22256852282815698} \\
7. & $\cos_{\mathcal{A}\mathcal{V}}[\text{w},\infty,\phi_2,\mathbb{L}_{12}]$ & \numprint{0.6940259740259741} & \numprint{0.7662034262034262} & \numprint{0.832293363314451} & \numprint{0.14279573663003836} & \numprint{0.13038814952295522} & \numprint{0.28648049747499443} & \numprint{0.12671380482759498} & \numprint{0.22946734094640026} \\
8. & $\cos_{\mathcal{A}}[\neg{\text{w}},\infty,\phi_1,\mathbb{L}_{12}]$ & \numprint{0.6421010668418311} & \numprint{0.6814861793774114} & \numprint{0.7168186235533174} & \numprint{0.14279573663003836} & \numprint{0.13038814952295522} & \numprint{0.28648049747499443} & \numprint{0.12239154589122826} & \numprint{0.21202169976203178} \\
9. & $\cos_{\mathcal{A}\mathcal{V}}[\neg{\text{w}},\infty,\phi_1,\mathbb{L}_{12}]$ & \textbf{\numprint{0.7415196479769677}} & \textbf{\numprint{0.773102386335469}} & \numprint{0.7900375814661528} & \numprint{0.14279573663003836} & \numprint{0.13038814952295522} & \numprint{0.28648049747499443} & \numprint{0.1024745058984006} & \numprint{0.15643639740273782} \\
10. & NetSimile & \numprint{0.4785673420561859} & \numprint{0.6294710496068731} & \numprint{0.7220723413688953} & \numprint{0.14279573663003836} & \numprint{0.13038814952295522} & \numprint{0.28648049747499443} & \numprint{0.1268092963300572} & \numprint{0.229382657244954} \\
11. & ToSi & \numprint{0.3903569974998546} & \numprint{0.4329537875756363} & \numprint{0.46522186147186145} & \numprint{0.14279573663003836} & \numprint{0.13038814952295522} & \numprint{0.28648049747499443} & \numprint{0.10822046220541263} & \numprint{0.1485720504461578} \\
};
\end{tikzpicture}	
	\caption{$F$-scores of classifying TTNs into seven target classes (\Cities, \Regions, \Others, \WPRegioOne, \WPRegioTwo, \WPOthersOne and \WPOthersTwo) by means of SVMs using RBF kernels.
	\textit{Column \emph{all}:} $F$-scores, if all features are used by the similarity measure (row).
	\textit{Column \emph{opt}:} $F$-scores, if a subset of features selected by the genetic search is used.
	\textit{Column \emph{ext}:} $F$-scores, if a subset of features selected by the extended genetic search is used. 
	The last five columns display the $F$-scores of the random baselines B1, B3 and B4, in the case of B3 and B4 differentiated for the variants \textit{all} and \textit{opt}.}
	\label{table:ttn_graphsim_classification_summary}
\end{table}
\begin{table}[t]
	\centering
	\nprounddigits{3}
	\begin{tikzpicture}
	[Kasten/.style={rectangle, minimum height=0.60cm, anchor=west}, Path/.style={font={\footnotesize\vphantom{Ag}}, midway, above},]	
	\matrix [matrix of nodes, nodes=Kasten, row sep=0.05cm, column sep=0.05cm, font={\footnotesize\vphantom{Ag}}, 
	row 1/.append style={nodes={fill=SeminarSehrHellBlau}},
	%
	column 1/.append style={nodes={text width=0.3cm, align=right, fill=SeminarSehrHellBlau}},
	column 2/.append style={nodes={text width=2.8cm, align=left, fill=SeminarSehrSehrHellGrau}},
	column 3/.append style={nodes={text width=0.6cm, align=right, fill=SeminarSehrHellGrau}},
	column 4/.append style={nodes={text width=0.7cm, align=right, fill=SeminarSehrHellGrau}},
	column 5/.append style={nodes={text width=0.65cm, align=right, fill=SeminarSehrHellGrau}},
	column 6/.append style={nodes={text width=0.6cm, align=right, fill=SeminarSehrSehrHellGrau}},
	column 7/.append style={nodes={text width=0.7cm, align=right, fill=SeminarSehrSehrHellGrau}},
	column 8/.append style={nodes={text width=0.8cm, align=right, fill=SeminarSehrSehrHellGrau}},
	column 9/.append style={nodes={text width=0.7cm, align=right, fill=SeminarSehrSehrHellGrau}},
	column 10/.append style={nodes={text width=0.8cm, align=right, fill=SeminarSehrSehrHellGrau}},
	column 11/.append style={nodes={text width=0.7cm, align=right, fill=SeminarSehrSehrHellGrau}},
	column 12/.append style={nodes={text width=0.8cm, align=right, fill=SeminarSehrSehrHellGrau}},
	] (Link) { 
 & Measure & all & opt & ext & B1 & B2 all & B2 opt & B3 all & B3 opt & B4 all & B4 opt \\ 
1. & $\ges$ & \numprint{0.5983340168603327} & \numprint{0.6493770491803279} & \numprint{0.7524118738404453} & \numprint{0.19991273846154414} & \numprint{0.22553224196757854} & \numprint{0.325173333654575} & \numprint{0.1818679684633869} & \numprint{0.39729957392948684} & \numprint{0.17554107981829542} & \numprint{0.2944440720498837} \\
2. & $\wal$ & \numprint{0.6098663887652489} & \numprint{0.6353455433455433} & \numprint{0.70720041862899} & \numprint{0.19991273846154414} & \numprint{0.16775580592411643} & \numprint{0.22223682419788238} & \numprint{0.1818679684633869} & \numprint{0.39729957392948684} & \numprint{0.15832569212679753} & \numprint{0.2888297023565099} \\
3. & $\veo$ & \numprint{0.6360963695493462} & \numprint{0.7064364207221351} & \numprint{0.7832505753688019} & \numprint{0.19991273846154414} & \numprint{0.21266837828535498} & \numprint{0.306217424427005} & \numprint{0.1818679684633869} & \numprint{0.39729957392948684} & \numprint{0.16958034103737765} & \numprint{0.3080817801522055} \\
4. & $\wges$ & \numprint{0.4575011655011655} & \numprint{0.5758976317799848} & \numprint{0.6180395794681509} & \numprint{0.19991273846154414} & \numprint{0.31091498493684605} & \numprint{0.34789948979168406} & \numprint{0.1818679684633869} & \numprint{0.39729957392948684} & \numprint{0.1732613292702617} & \numprint{0.2808113643971854} \\
5. & $\cos_{\mathcal{A}}[\text{w},\infty,\phi_1,\mathbb{L}_{12}]$ & \numprint{0.5668642162435266} & \numprint{0.6729172305526985} & \numprint{0.7367254698489567} & \numprint{0.19991273846154414} & - & - & \numprint{0.1818679684633869} & \numprint{0.39729957392948684} & \numprint{0.17287907702804123} & \numprint{0.29998283907649187} \\
6. & $\cos_{\mathcal{A}\mathcal{V}}[\text{w},\infty,\phi_1,\mathbb{L}_{12}]$ & \textbf{\numprint{0.7398384399131703}} & \numprint{0.7771923314780458} & \numprint{0.853665000268774} & \numprint{0.19991273846154414} & \numprint{0.24154052907070767} & \numprint{0.4402360659194667} & \numprint{0.1818679684633869} & \numprint{0.39729957392948684} & \numprint{0.1812180920100091} & \numprint{0.31951474474662545} \\
7. & $\cos_{\mathcal{A}\mathcal{V}}[\text{w},\infty,\phi_2,\mathbb{L}_{12}]$ & \numprint{0.6115979070873633} & \textbf{\numprint{0.815834837298332}} & \textbf{\numprint{0.8752197423004876}} & \numprint{0.19991273846154414} & - & - & \numprint{0.1818679684633869} & \numprint{0.39729957392948684} & \numprint{0.18678210259962674} & \numprint{0.3401235397118218} \\
8. & $\cos_{\mathcal{A}}[\neg{\text{w}},\infty,\phi_1,\mathbb{L}_{12}]$ & \numprint{0.558931216931217} & \numprint{0.5996719576719578} & \numprint{0.6521916592724045} & \numprint{0.19991273846154414} & - & - & \numprint{0.1818679684633869} & \numprint{0.39729957392948684} & \numprint{0.18175087382587726} & \numprint{0.3071654284830821} \\
9. & $\cos_{\mathcal{A}\mathcal{V}}[\neg{\text{w}},\infty,\phi_1,\mathbb{L}_{12}]$ & \numprint{0.7212940187915361} & \numprint{0.8108239522223982} & \numprint{0.8649130763416478} & \numprint{0.19991273846154414} & \numprint{0.23984379471303058} & \numprint{0.46407453546761934} & \numprint{0.1818679684633869} & \numprint{0.39729957392948684} & \numprint{0.18217413385486306} & \numprint{0.3172001654117379} \\
10. & NetSimile & \numprint{0.46720575022461813} & \numprint{0.5069832338342597} & \numprint{0.6095894909688013} & \numprint{0.19991273846154414} & \numprint{0.4936847368715241} & \numprint{0.6023374088915949} & \numprint{0.1818679684633869} & \numprint{0.39729957392948684} & \numprint{0.17255756841880251} & \numprint{0.27196348884722554} \\
11. & ToSi & \numprint{0.43126601341704324} & \numprint{0.5666370471633629} & \numprint{0.5853383458646617} & \numprint{0.19991273846154414} & - & - & \numprint{0.1818679684633869} & \numprint{0.39729957392948684} & \numprint{0.17903593913793603} & \numprint{0.2535734199546465} \\
};
\end{tikzpicture}	
	\caption{$F$-scores of classifying ATNs into five classes (\Cities, \Regions, \Others, \WPRegioOne and \WPOthersOne) by means of SVMs using RBF kernels.
	\textit{Column \emph{all}:} $F$-scores using all features in terms of the respective similarity measure.
	\textit{Column \emph{opt}:} using a subset of features detected according to a genetic search. 
	\textit{Column \emph{ext}:} subset selection according to extended genetic optimization. 
	Additionally, $F$-scores of random baselines B1, B2, B3 and B4 are displayed, in the latter three cases differentiated for the variants \textit{all} and \textit{opt}.}
	\label{table:atn_graphsim_classification_summary}
\end{table}
Relating to Hypothesis \ref{hypo:Genre-oriented Classification}, Table \ref{table:ttn_graphsim_classification_summary} and Table \ref{table:atn_graphsim_classification_summary} summarize our findings regarding the genre-sensitive classification of TTNs and ATNs, respectively.
Cosine-based measures always perform best. 
Especially in the case of ATNs we see that accounting for arcs \textit{and} for nodes secures better performance:
dual weight-dependent measures (see Section \ref{sec:Module 6: Graph Similarity Analysis}) outperform single weight-dependent or weight-insensitive measures.
However, in the case of TTNs, we also see that as long as we do not perform an extended optimization (ext), the measure $\cos_{\mathcal{A}\mathcal{V}}[\neg{\text{w}},\infty,\phi_1,\mathbb{L}_{12}]$, which disregards arc weights, is a best performer.
Of special interest is $\cos_{\mathcal{A}\mathcal{V}}[\text{w},\infty,\phi_2,\mathbb{L}_{12}]$, the best performer regarding the classification of ATNs (Table \ref{table:atn_graphsim_classification_summary}), which is not only arc and node sensitive, but also weights nodes as a function of their degree centrality and therefore covers the highest amount of structural information among all candidates considered here. 
This measure is also a robust candidate working at a high level in both experiments (it is the \nth{2} best performer in the case of TTNs if being optimized by an extended genetic search). 
Thus, we conclude that spherical measures clearly outperform GED-related approaches and especially network-topology-based approaches (ToSi and NetSimile) which perform worst: 
the kind of information we seek is apparently ignored or \enquote{abstracted away} by the latter measures.
However, NetSimile has at least a high optimization potential (see the column ext in Table \ref{table:ttn_graphsim_classification_summary}) -- a potential which is missing in the case of ToSi.
In any event, non of the measures considered here is outperformed by our baselines.
But in Table \ref{table:ttn_graphsim_classification_summary} we also see that B3 (opt) approaches ToSi (all); in Table \ref{table:atn_graphsim_classification_summary} we make analog observations also by example of other measures.
A serious problem concerns NetSimile in relation to Baseline B2 regarding the classification of ATNs (Table \ref{table:atn_graphsim_classification_summary}): 
the baseline surpasses the topology-related measure whether being optimized (opt) or not (all).
The graph indices collected by NetSimile have obviously difficulties in making observed networks distinguishable from their random counterparts -- at least in some of the cases considered here.
B3 is also of interest with regard to the classification of ATNs, which achieves $F$-scores of up to 40\% and thus makes representation models based on measures such as NetSimile, ToSi and $\wges$ problematic candidates. 
The values of B4 opt are also remarkably high and can therefore be regarded as a challenge for the measures.

\begin{figure}[t]
	\resizebox{1.00\textwidth}{!}{
		\includegraphics[]{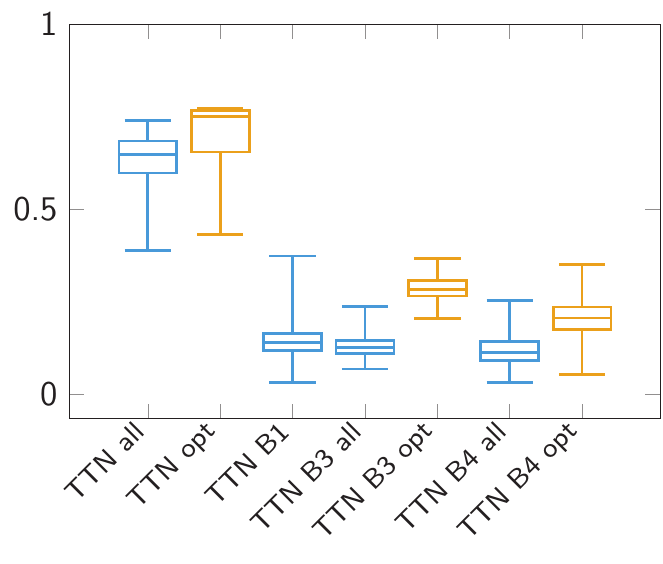}
		\includegraphics[]{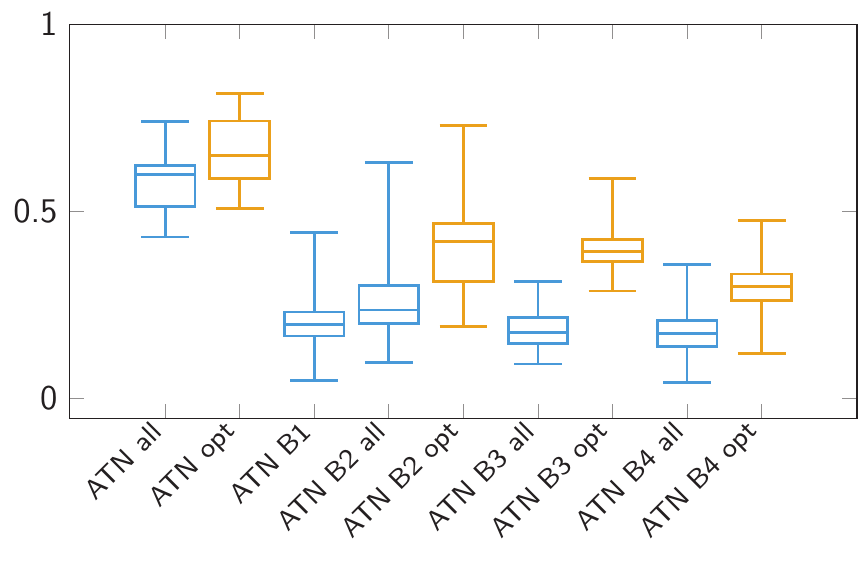}
	}
	\caption{Left: 
	boxplots of $F$-scores obtained for classifying TTNs contrasted by the baselines B1, B3 and B4.
	Right: 
	boxplots of $F$-scores obtained for classifying ATNs contrasted by the baselines B1, B2, B3 and B4.
	}
	\label{fig:graphsim_ttn_random_baselines}
\end{figure}

Figure \ref{fig:graphsim_ttn_random_baselines} shows that the baselines B1, B3 and B4 are outperformed by the results obtained for TTNs.
However, it also shows that feature optimization affects the random baselines.
This is particularly evident in the case of B3, which is based on random matrices.
This gain in $F$-score can be explained by random numbers that allow the target classes to be separated -- at least to some extent.
These features are then selected by the genetic feature selection.
The baseline results for ATNs show a similar picture (see Figure \ref{fig:graphsim_ttn_random_baselines}, right).
Regarding B2, we make the following observations in Figure \ref{fig:graphsim_ttn_random_baselines} (right) (for reasons of complexity we did not consider all measures to compute B2):
although the best B2 candidates are better than the average $F$-scores calculated on the basis of real data, B2 is clearly surpassed on average. 
Thus, we come to the conclusion that we found effective measures for comparing networks -- this concerns in particular the spherical approach based on the cosine measure.
From these experiments we conclude:
\begin{enumerate}
\item Hypothesis \ref{hypo:Genre-oriented Classification} is not falsified: 
we know the genre of a topic network by its structure. 
Note that this only concerns Scenario {\setlength{\fboxsep}{0.5mm}\fbox{2}} and {\setlength{\fboxsep}{0.5mm}\fbox{3}} of Figure \ref{fig:Classification Scenarios} -- Scenario {\setlength{\fboxsep}{0.5mm}\fbox{4}} is not computed here.
Similarly, by calculating our baselines, this also involves the scenarios {\setlength{\fboxsep}{0.5mm}\fbox{7}} and {\setlength{\fboxsep}{0.5mm}\fbox{8}} while ignoring Scenario {\setlength{\fboxsep}{0.5mm}\fbox{9}}.
The classification benefits especially from information that is explored by dual weight-dependent measures.
This holds regardless of the mode (ATN or TTN).
\item Spherical measures should be preferred to GED-based measures, and these in turn to topology-based measures:
\begin{equation}
\text{spherical} \succ \text{GED} \succ \text{topological}
\end{equation}
\end{enumerate}

\begin{figure}[t]
	\includegraphics[width=0.60\textwidth]{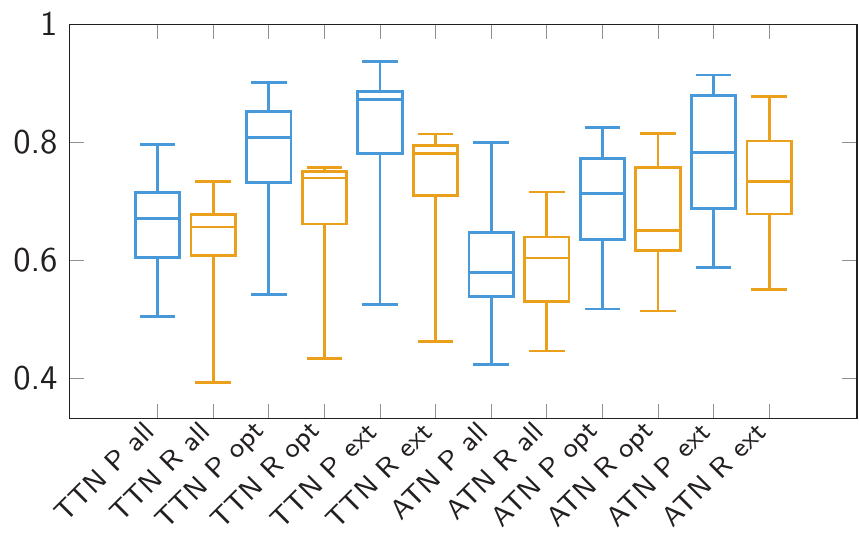}
	\caption{Boxplots of precision (P) and recall (R) values ($y$-axis) induced by the measures of Table \ref{tab:Graph Similarity Measures} and underlying the $F$-scores of Table \ref{table:ttn_graphsim_classification_summary} (first six columns) and Table \ref{table:atn_graphsim_classification_summary} (last six columns). 
	Distributions are distinguished by considering all features (all) or subsets of them generated by the genetic optimizations \textit{opt} or \textit{ext}.}
	\label{fig:graphsim_ttn_atn_overall_measures_boxplots}
\end{figure}

The boxplots in Figure \ref{fig:graphsim_ttn_atn_overall_measures_boxplots} give another perspective on the classification results by summarizing the distributions of precision and recall values generated by the graph similarity measures.
Except for the results on ATN using all features, the average precision is higher than the average recall.
The figure also demonstrates the strong effect of feature selection.

\begin{figure}[t]
	\includegraphics[width=0.75\textwidth]{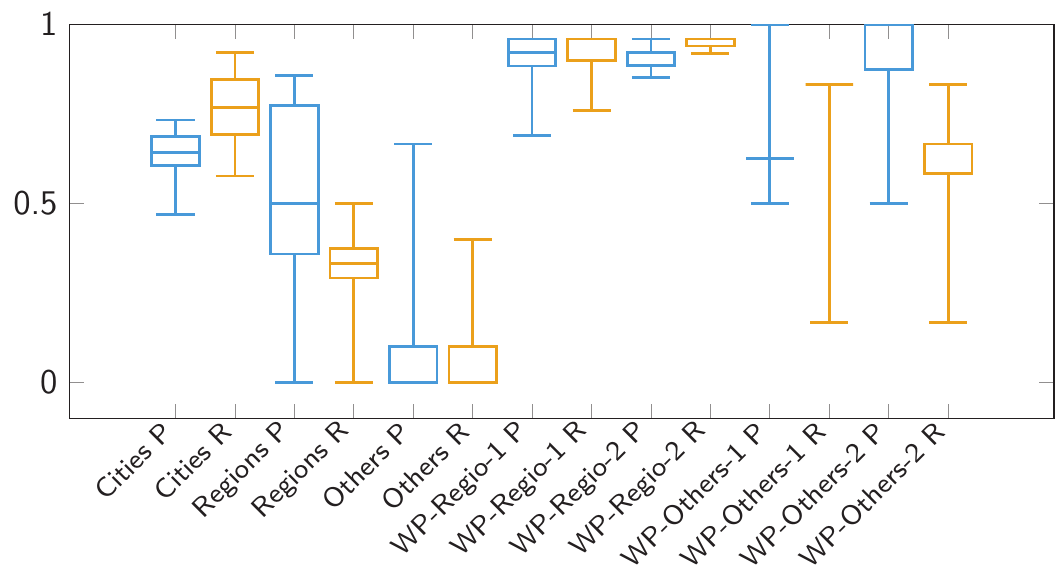}
	\caption{Boxplots of precision (P) and recall (R) values ($y$-axis) induced by the measures of Table \ref{tab:Graph Similarity Measures} underlying the $F$-scores of Table \ref{table:ttn_graphsim_classification_summary}. Distributions are distinguished by the respective target class of the classification.}
	\label{fig:graphsim_ttn_per_category_measures_boxplots}
\end{figure}

\begin{figure}[t]
	\includegraphics[width=0.65\textwidth]{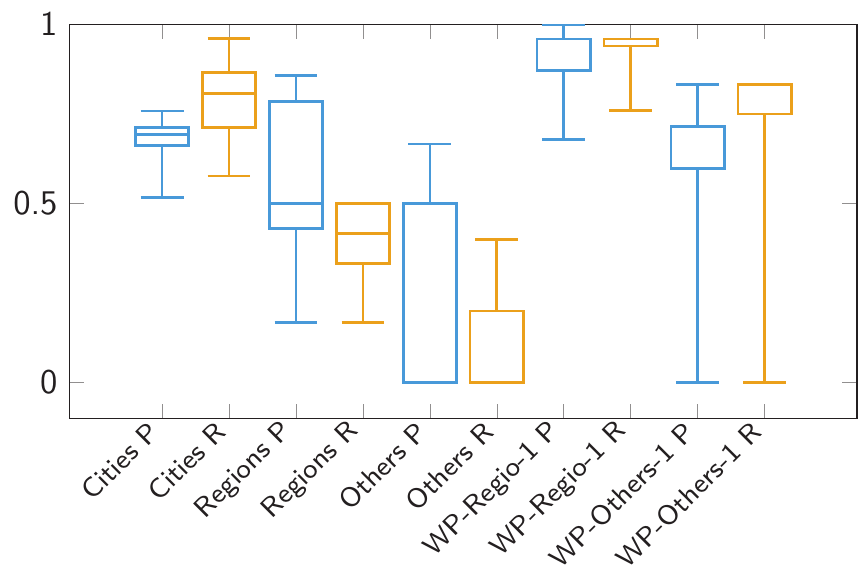}
	\caption{Boxplots of precision (P) and recall (R) values ($y$-axis) induced by the measures of Table \ref{tab:Graph Similarity Measures} underlying the $F$-scores of Table \ref{table:atn_graphsim_classification_summary}. Distributions are distinguished by the respective target class of the classification.}
	\label{fig:graphsim_atn_per_category_measures_boxplots}
\end{figure}

So far, we considered classifications as a whole and thus abstracted from the scores obtained for individual genres.
The boxplots in Figure \ref{fig:graphsim_ttn_per_category_measures_boxplots} give insights into these genre-related scores regarding the classification of TTNs by means of the extended feature optimization (ext).
The members of the genre \Cities are well identified: in terms of recall and precision.
The genre \Regions is far less separable and causes many classification errors (low recall).
Apparently, this class contains more heterogeneous TTNs. 
In any event, the Wikipedia-based genres \WPRegioOne and \WPRegioTwo are very well separated.
By contrast, instances of the category \Others are extremely difficult to detect (as predicted in Section \ref{sec:Module 8 -- Machine Learning -- and Module 9 -- Classification Analysis}, page \pageref{page:Control CLasses}).
Similarly, elements of the classes \WPOthersOne and \WPOthersTwo are difficult to identify -- albeit to a minor degree.
Thus we conclude: the upper bound of separability concerns Wikipedia-based regional wikis. 
The corresponding subgraphs are very similar.
This upper bound is approached by city wikis. 
Region wikis are less homogeneous, making the corresponding class \Regions rather blurred and therefore question its status as a genre.
Figure~\ref{fig:graphsim_atn_per_category_measures_boxplots} shows the corresponding results of classifying ATNs.
The general picture is quite similar to that of the TTNs.

\begin{figure}[t]
	\includegraphics[width=1.00\textwidth]{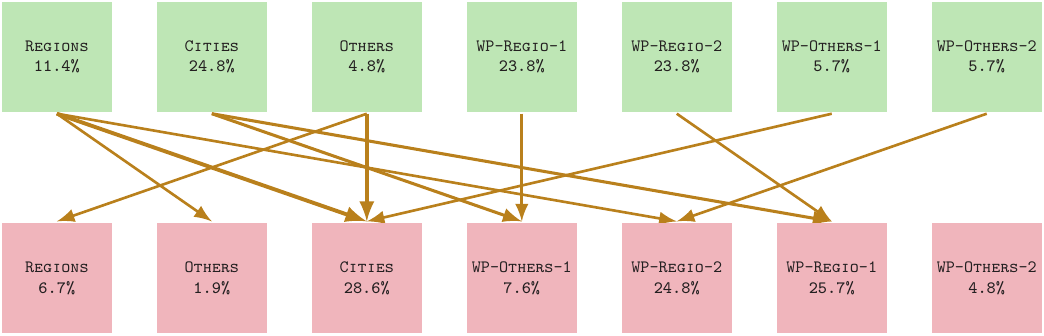}
	\caption{Error analysis regarding the classification of TTNs by means of $\cos_{\mathcal{A}\mathcal{V}}[\neg{\text{w}},\infty,\phi_1,\mathbb{L}_{12}]$.}
	\label{fig:ttn_mismatch_example.tex}
\end{figure}

We take another perspective on the results to examine classification errors.
The best results on TTNs using all features is achieved by $\cos_{\mathcal{A}\mathcal{V}}[\neg{\text{w}},\infty,\phi_1,\mathbb{L}_{12}]$.
Figure~\ref{fig:ttn_mismatch_example.tex} shows to what degree wikis of a target class are wrongly classified using this measure.
The labels show the proportion of the categories according to the gold standard (top) and the classification result (bottom).
The picture is diverse, but some details become clear:
wikis of the classes \Regions and \Others are often falsely categorized as \Cities.
City wikis on the other hand are wrongly classified as \WPOthersOne or \WPRegioOne.

\begin{figure}[t]
	\includegraphics[width=0.7\textwidth]{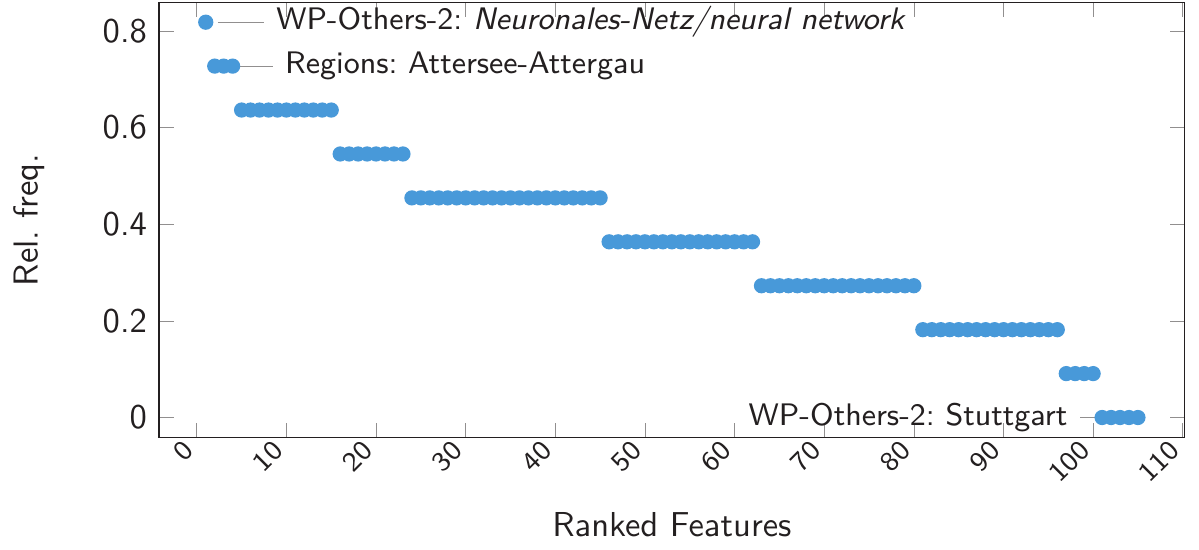}
	\caption{Ranking of the relative frequencies of features as a result of being selected by the extended genetic feature optimization in the classification of TTNs.}
	\label{fig:graphsim_ttn_feature_use_distribution}
\end{figure}

Genetic feature selection has proven to increase $F$-score significantly.
In the extended optimization (ext) the last step is to minimize the number of features used.
Since our features stand for similarities to networks, we have to ask whether some of the wikis underlying these networks are more relevant for the differentiation of the target classes than others -- possibly because of their prototypical status.
If all wikis were equally important, an equal distribution of the frequencies with which these features are selected by the genetic optimization would be expected.
Figure \ref{fig:graphsim_ttn_feature_use_distribution} shows the corresponding rank frequency distribution: 
it shows that we are far from evenly distributed features.
From this we conclude that the selection of features is indispensable and that the underlying wikis are very different in their roles in our classification experiments.

\begin{figure}[t]
	\includegraphics[width=0.15\textwidth]{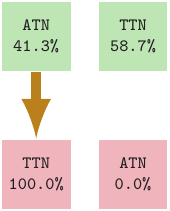}
	\caption{Error analysis regarding the classification of TTNs vs.\ ATNs by means of $\cos_{\mathcal{A}\mathcal{V}}[\neg{\text{w}},\infty,\phi_1,\mathbb{L}_{12}]$.}
	\label{fig:ttn_mismatch_example_2cat}
\end{figure}

Next, we try to distinguish TTNs from ATNs thereby addressing Hypothesis \ref{hypo:TTNs versus ATNs} (or more specifically Scenario {\small\setlength{\fboxsep}{0.5mm}\fbox{1}} of Figure \ref{fig:Classification Scenarios}).
The error analysis in Figure \ref{fig:ttn_mismatch_example_2cat} shows that networks of these two modes are not separable using our approach. 
Table \ref{table:ttn_atn_12cat_graphsim_classification_summary.tex} differentiates this outcome by reporting the results obtained for different measures. 
It shows that this classification scenario is far exceeded by Baseline B1 and is therefore irrelevant.
From this result we conclude that ATNs are so similar to their corresponding TTNs that they cannot be distinguished by our measures, or alternatively: our similarity measures are not suitable to distinguish them. 
This is not surprising, as the order and size of an ATN always corresponds to the order and size of the TTN from which it was derived, so that they can only differ by the weighting of their nodes and arcs.
By concerning Hypothesis \ref{hypo:Genre-and Mode-oriented Classification} and thus by distinguishing twelve target classes (in the case of \WPOthersTwo and \WPRegioTwo we do not induce ATNs), Table \ref{table:ttn_atn_12cat_graphsim_classification_summary.tex} shows a somehow different scenario: 
though the $F$-scores are still rather low, Baseline B1 is clearly outperformed when using a cosine measure for graph similarity measurement.
From this observation, we conclude that while Hypothesis \ref{hypo:TTNs versus ATNs} is falsified, there is at least a potential regarding the simultaneous distinction of genre and mode: ATNs do not uniformly resemble their corresponding TTNs.

\begin{table}[t]
	\centering
	\nprounddigits{3}
	\resizebox{1.0\textwidth}{!}{
		\begin{tikzpicture}
	[Kasten/.style={rectangle, minimum height=0.60cm, anchor=west}, Path/.style={font={\footnotesize\vphantom{Ag}}, midway, above},]	
	\matrix [matrix of nodes, nodes=Kasten, row sep=0.05cm, column sep=0.05cm, font={\footnotesize\vphantom{Ag}}, 
	row 1/.append style={nodes={fill=SeminarSehrHellBlau}},
	%
	column 1/.append style={nodes={text width=0.2cm, align=right, fill=SeminarSehrHellBlau}},
	column 2/.append style={nodes={text width=2.35cm, align=left, fill=SeminarSehrSehrHellGrau}},
	column 3/.append style={nodes={text width=0.55cm, align=right, fill=SeminarSehrHellGrau}},
	column 4/.append style={nodes={text width=0.55cm, align=right, fill=SeminarSehrHellGrau}},
	column 5/.append style={nodes={text width=0.55cm, align=right, fill=SeminarSehrHellGrau}},
	column 6/.append style={nodes={text width=0.55cm, align=right, fill=SeminarSehrSehrHellGrau}},
	] (Link) { 
 & Measure & all & opt & ext & B1 \\ 
1. & $\ges$ & \numprint{0.36971830985915494} & \numprint{0.36971830985915494} & \numprint{0.36971830985915494} & \numprint{0.5000075012870889} \\
2. & $\veo$ & \numprint{0.36971830985915494} & \numprint{0.36971830985915494} & \numprint{0.36971830985915494} & \numprint{0.5000075012870889} \\
3. & $\cos[\text{w},\infty,\phi_1,\mathbb{L}_{12}]$ & \numprint{0.36971830985915494} & \numprint{0.36971830985915494} & \numprint{0.36971830985915494} & \numprint{0.5000075012870889} \\
4. & $\cos[\neg{\text{w}},\infty,\phi_1,\mathbb{L}_{12}]$ & \numprint{0.36971830985915494} & \numprint{0.36971830985915494} & \numprint{0.36971830985915494} & \numprint{0.5000075012870889} \\
};
\end{tikzpicture}
		\hfill	
		\begin{tikzpicture}
	[Kasten/.style={rectangle, minimum height=0.60cm, anchor=west}, Path/.style={font={\footnotesize\vphantom{Ag}}, midway, above},]	
	\matrix [matrix of nodes, nodes=Kasten, row sep=0.05cm, column sep=0.05cm, font={\footnotesize\vphantom{Ag}}, 
	row 1/.append style={nodes={fill=SeminarSehrHellBlau}},
	%
	column 1/.append style={nodes={text width=0.2cm, align=right, fill=SeminarSehrHellBlau}},
	column 2/.append style={nodes={text width=2.35cm, align=left, fill=SeminarSehrSehrHellGrau}},
	column 3/.append style={nodes={text width=0.55cm, align=right, fill=SeminarSehrHellGrau}},
	column 4/.append style={nodes={text width=0.55cm, align=right, fill=SeminarSehrHellGrau}},
	column 5/.append style={nodes={text width=0.55cm, align=right, fill=SeminarSehrHellGrau}},
	column 6/.append style={nodes={text width=0.55cm, align=right, fill=SeminarSehrSehrHellGrau}},
	] (Link) { 
 & Measure & all & opt & ext & B1 \\ 
1. & $\ges$ & \numprint{0.15150829562594267} & \numprint{0.17814011367840446} & \numprint{0.19421911421911423} & \numprint{0.08245726495726498} \\
2. & $\veo$ & \numprint{0.1812003220260436} & \numprint{0.22826249380871233} & \numprint{0.2585691802339971} & \numprint{0.08245726495726498} \\
3. & $\cos[\text{w},\infty,\phi_1,\mathbb{L}_{12}]$ & \numprint{0.3148876868431056} & \numprint{0.3630259271907441} & \numprint{0.4074489504206485} & \numprint{0.08245726495726498} \\
4. & $\cos[\neg{\text{w}},\infty,\phi_1,\mathbb{L}_{12}]$ & \numprint{0.28403881963206146} & \numprint{0.3390420436327762} & \numprint{0.408509840964903} & \numprint{0.08245726495726498} \\
};
\end{tikzpicture}		
	}
	\caption{Left: 
	$F$-scores obtained for different measures and optimizations by classifying ATNs vs.\ TTNs according to Scenario {\small\setlength{\fboxsep}{0.5mm}\fbox{1}} of Table \ref{fig:Classification Scenarios} -- two target classes are considered. B1 considers Scenario {\small\setlength{\fboxsep}{0.5mm}\fbox{6}} of Figure \ref{fig:Classification Scenarios}.
	Right: $F$-scores obtained for different measures and optimizations by classifying simultaneously for mode and genre according to Scenario {\small\setlength{\fboxsep}{0.5mm}\fbox{5}} -- twelve target classes are considered. B1 considers Scenario {\small\setlength{\fboxsep}{0.5mm}\fbox{10}}.
	}
	\label{table:ttn_atn_12cat_graphsim_classification_summary.tex}
\end{table}

\begin{figure}[t]
	\resizebox{1.0\textwidth}{!}{
		\includegraphics[]{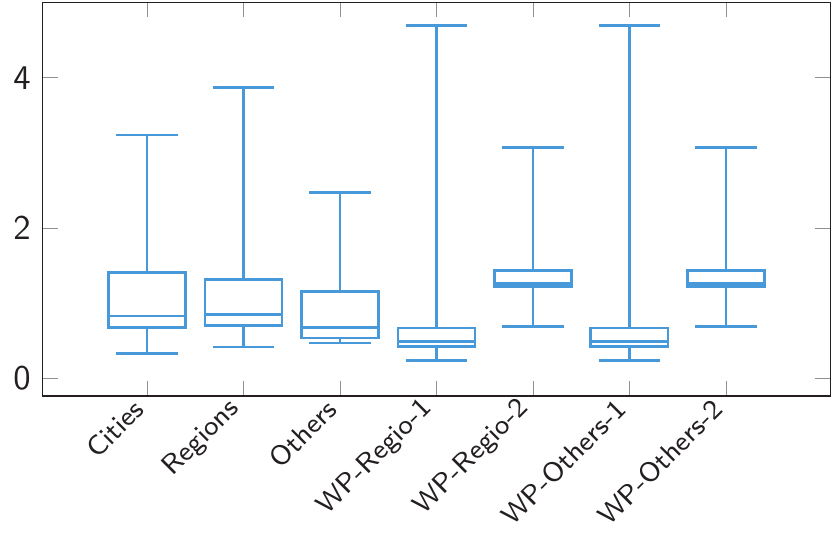}
		\hfill
		\includegraphics[]{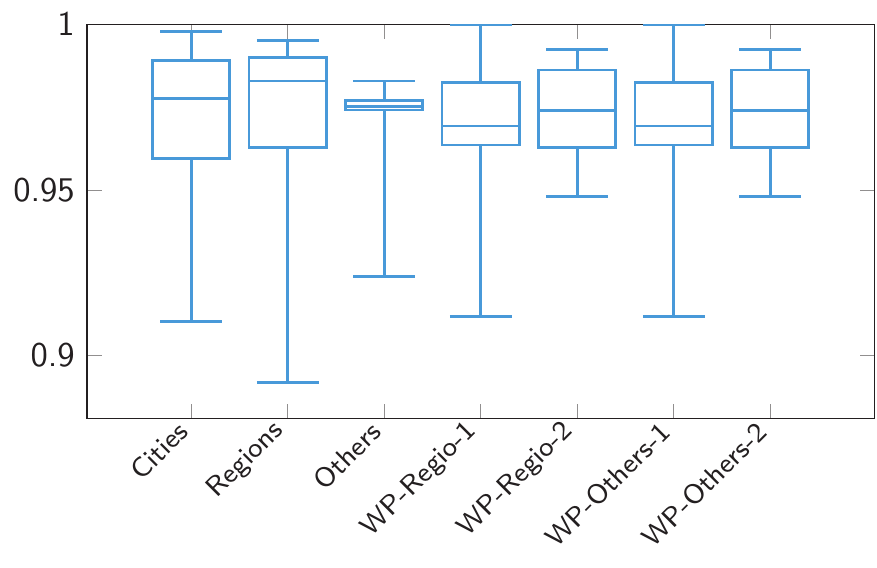}
	}
	\caption{Left: 
	boxplots of the distribution of the exponents of the power laws fitted to the weight distributions of the nodes in the TTNs differentiated by the target classes.
	Right: 
	the boxplots of the corresponding fitting accuracies computed by means of the adjusted R-squared.
	}
	\label{fig:graphsim_ttn_powerfit_per_category_boxplots}
\end{figure}

So far we considered part (2) of Hypothesis \ref{hypo:Hypothesis 1} by showing that TTNs (and also ATNs) with similar functions resemble each other, while differing from networks of other genres.
It remains to be shown that these networks are also thematically focused -- in a highly skewed manner.
To test this, we fit power laws to the distributions of node weights in TTNs.
Remember that these weights result from detecting textual instances of the topic represented by the respective node so that the more such instances are detected, the more salient the topic in the network.
Fitting a power law to such a distribution means that there is a minority of topics or just one topic that surpasses all other topics in its importance, while the majority of topics is of little or no importance.
The boxplots in Figure \ref{fig:graphsim_ttn_powerfit_per_category_boxplots} (left) show the distribution of the exponents of the power laws fitted to these distributions, differentiated by the genres considered here.
To assess the goodness of the fittings we compute the adjusted R-squares and display the value distributions in Figure \ref{fig:graphsim_ttn_powerfit_per_category_boxplots} (right).
Obviously, the fits are very good (the adjusted R-squares are on average above 95\%) while the averages of the exponents range between {$0.5$} and {$1.5$}:
From this analysis we conclude that the underlying wikis are all thematically focused and skewed by dealing with a minority of topics in depth.
The five most detected DDC labels per genre are shown in Table \ref{table:ttn_ddc3_distribution}. 
It shows that \textit{Transportation; ground transportation} is by far the most dominant topic in city wikis and in region wikis.
\textit{Obviously, these wikis are thematically focused in a highly skewed manner.}

\begin{table}[t]
	\centering
	\nprounddigits{3}
	\resizebox{0.75\textwidth}{!}{
		\selectlanguage{english}
\begin{tikzpicture}
[Kasten/.style={rectangle, minimum height=0.60cm, anchor=west}, Path/.style={font={\footnotesize\vphantom{Ag}}, midway, above},]
\matrix [matrix of nodes, nodes=Kasten, row sep=0.05cm, column sep=0.05cm, font={\footnotesize\vphantom{Ag}},
row 1/.append style={nodes={fill=SeminarSehrHellBlau}},
row 7/.append style={nodes={fill=SeminarSehrHellGrau}},
row 8/.append style={nodes={fill=SeminarSehrHellGrau}},
row 9/.append style={nodes={fill=SeminarSehrHellGrau}},
row 10/.append style={nodes={fill=SeminarSehrHellGrau}},
row 11/.append style={nodes={fill=SeminarSehrHellGrau}},
column 1/.append style={nodes={text width=0.6cm, align=center, fill=SeminarHellGrau}},
column 2/.append style={nodes={text width=1.0cm, align=left, fill=SeminarSehrSehrHellGrau}},
column 3/.append style={nodes={text width=2.5cm, align=right, fill=SeminarSehrSehrHellGrau}},
column 4/.append style={nodes={text width=1.4cm, align=right, fill=SeminarSehrSehrHellGrau}},
column 5/.append style={nodes={text width=0.6cm, align=right, fill=SeminarSehrSehrHellGrau}},
column 6/.append style={nodes={text width=4.5cm, align=left, fill=SeminarSehrSehrHellGrau}},
] (Link) {
	Rank & Genre & Node Weight Sum & Avg Weight & DDC & Description \\
	1. & City & \numprint{10325.830477445626} & \numprint{397.147326055601} & 388 & Transportation; ground transportation \\
	2. & City & \numprint{2404.6314207088935} & \numprint{92.48582387341898} & 943 & Central Europe; Germany \\
	3. & City & \numprint{1570.0095045424944} & \numprint{60.38498094394209} & 726 & Buildings for religious purposes \\
	4. & City & \numprint{1512.5357501609947} & \numprint{58.174451929269026} & 725 & Public structures \\
	5. & City & \numprint{964.261660482086} & \numprint{37.08698694161869} & 711 & Area planning \\
	1. & Region & \numprint{5127.546208847856} & \numprint{427.295517403988} & 388 & Transportation; ground transportation \\
	2. & Region & \numprint{1692.2665176595779} & \numprint{141.02220980496483} & 943 & Central Europe; Germany \\
	3. & Region & \numprint{1385.0130452736375} & \numprint{115.41775377280312} & 726 & Buildings for religious purposes \\
	4. & Region & \numprint{1289.7223422468796} & \numprint{107.47686185390664} & 551 & Geology, hydrology \& meteorology \\
	5. & Region & \numprint{1171.6560419821858} & \numprint{97.63800349851549} & 796 & Athletic \& outdoor sports \& games \\
	1. & Other & \numprint{5335.554611414674} & \numprint{1067.1109222829348} & 929 & Genealogy, names \& insignia \\
	2. & Other & \numprint{1640.0416563577219} & \numprint{328.00833127154436} & 726 & Buildings for religious purposes \\
	3. & Other & \numprint{715.0842530292961} & \numprint{143.01685060585922} & 723 & Architecture from ca. 300 to 1399 \\
	4. & Other & \numprint{701.2981019678218} & \numprint{140.25962039356438} & 725 & Public structures \\
	5. & Other & \numprint{680.3086912774457} & \numprint{136.06173825548916} & 720 & Architecture \\
};
\end{tikzpicture}
	}
	\caption{The five most detected DDC labels for the genres \Cities, \Regions and \Others.}
	\label{table:ttn_ddc3_distribution}
\end{table}

\begin{figure}[t]
	\resizebox{1.0\textwidth}{!}{
		\includegraphics[]{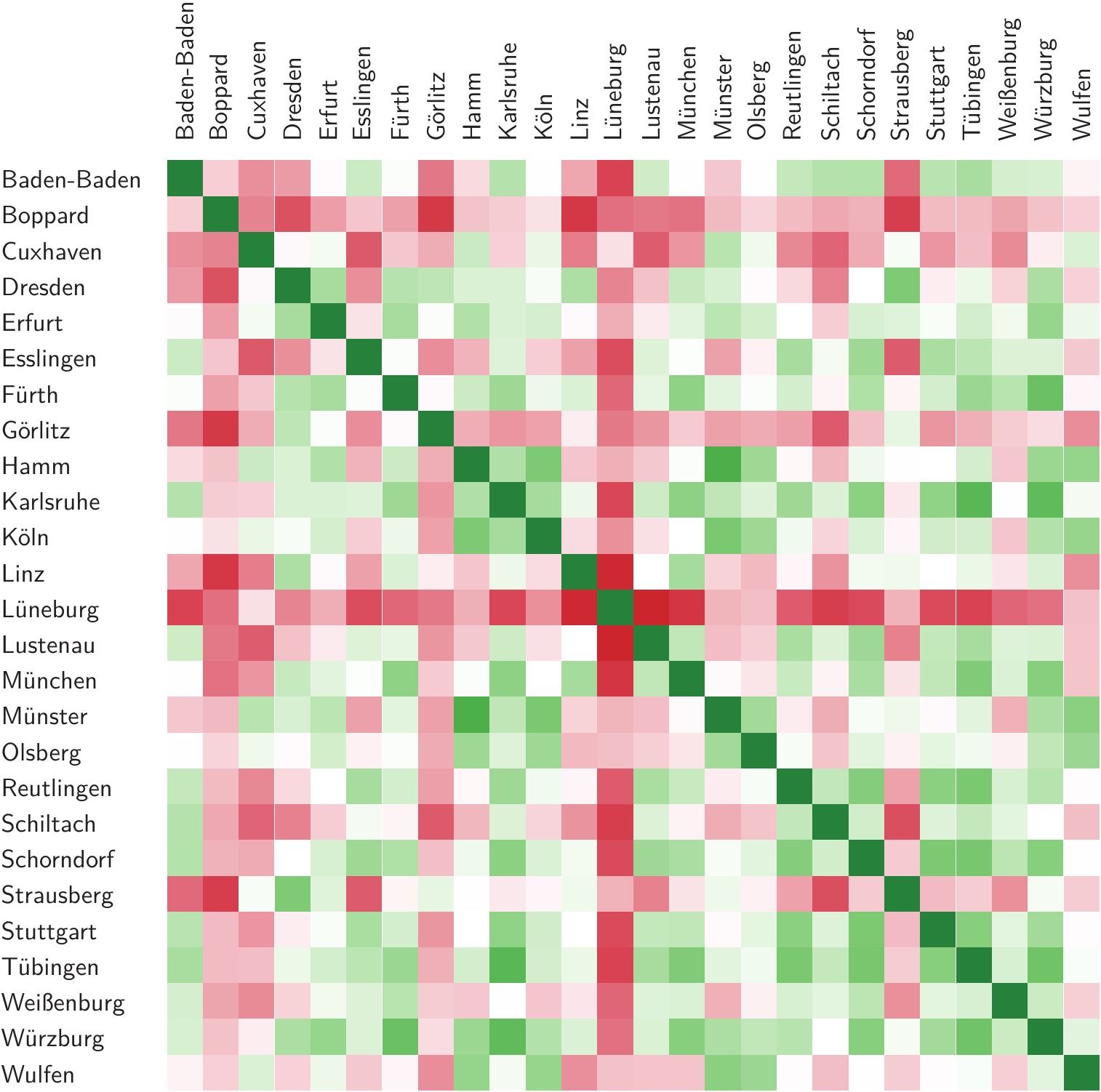}
		\hfill
		\includegraphics[]{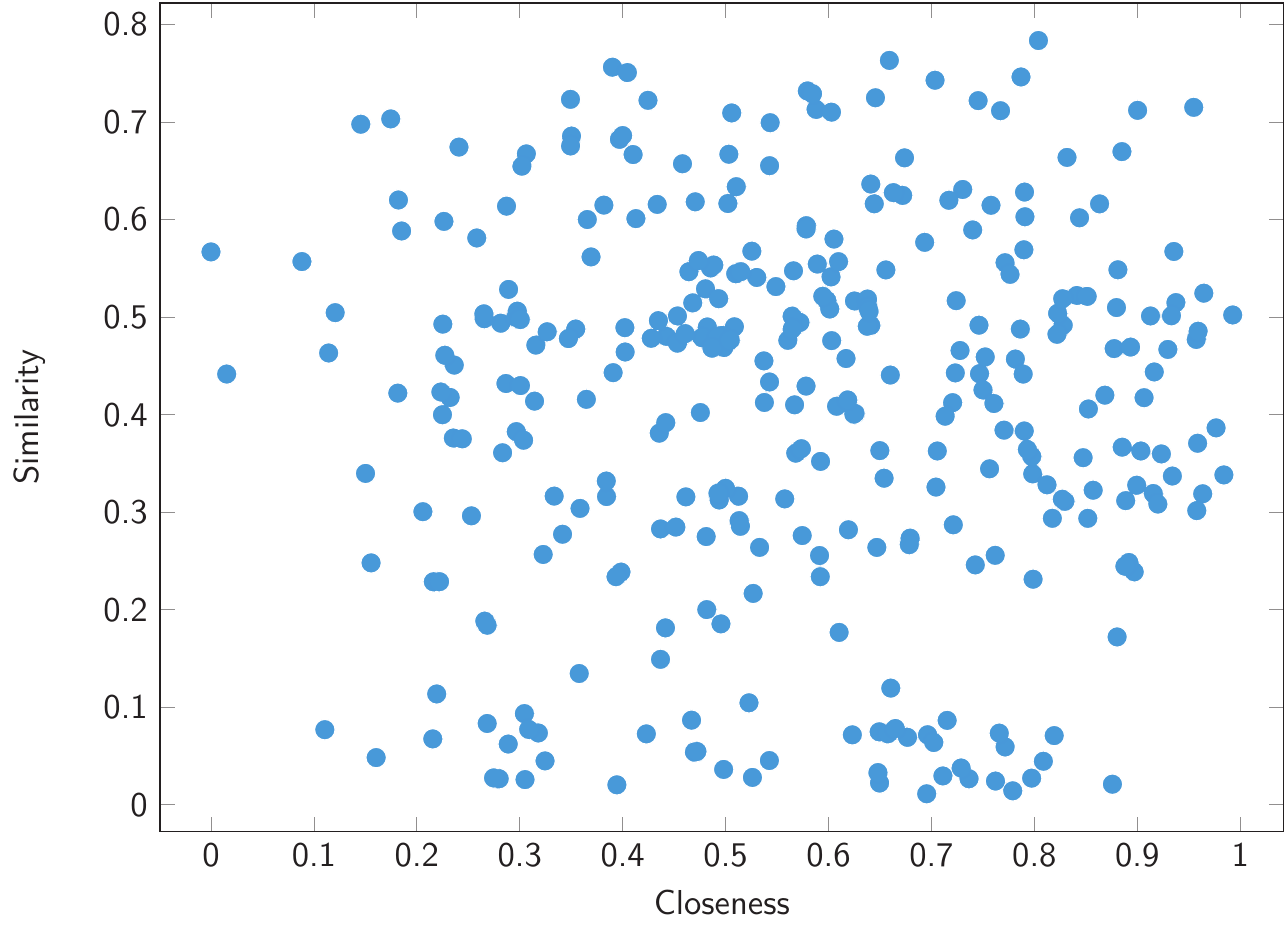}
	}
	\caption{Left: 
	the heatmap of thematic similarity and spatial closeness among city wikis.
	Red means that the wikis are thematically dissimilar and distant in space; 
	green means that they are thematically similar and close in space. 
	Right: the distribution of the similarities ($y$-axis) as a function of the closenesses ($x$-axis) of the different pairs of city wikis.
	}
	\label{fig:Heatmap of Place Distances and Similarities}
\end{figure}

\begin{figure}[t]
	\resizebox{1.0\textwidth}{!}{
		\includegraphics[]{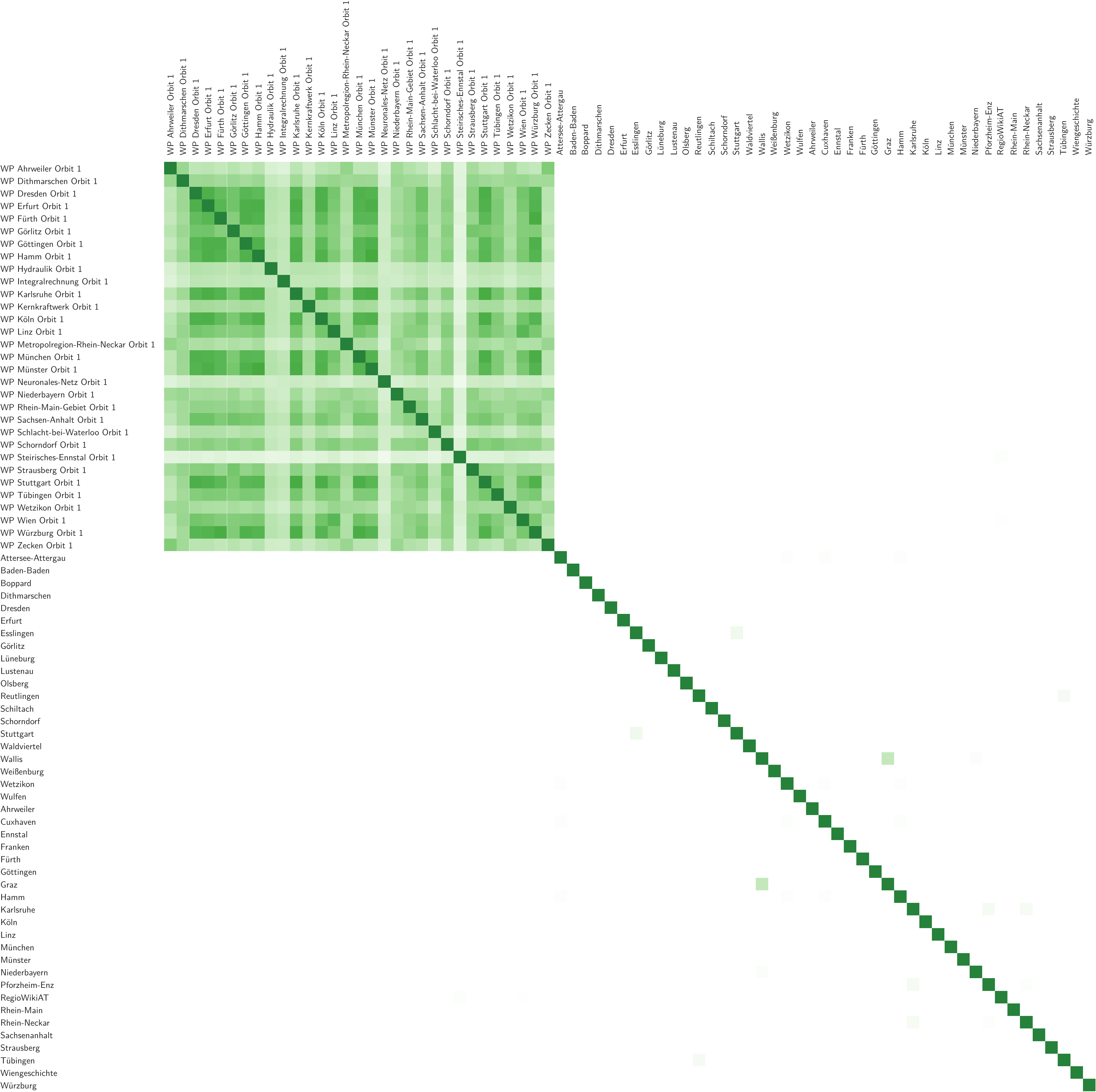}
	}
	\caption{The Fuzzy Jaccard overlap of the communities of registered authors of the wikis in the corpora \Cities, \Regions, \Others, \WPRegioOne and \WPOthersOne weighted by means of the writing activities of the authors: the greener the link, the higher the fuzzy overlap.}
	\label{fig:Heatmap of Overlapping Communities}
\end{figure}

It remains to be shown that our findings about urban wikis neither depend on the distances of the corresponding places nor on the communities writing these wikis.
Figure \ref{fig:Heatmap of Place Distances and Similarities} shows that the similarities detected by us do hardly correlate with the underlying distances of the places. 
In the heatmap in Figure \ref{fig:Heatmap of Place Distances and Similarities} (left), a connection between two city wikis is the greener, the closer and the more similar they are to each other, while a pair of wikis is the more red, the less similar and the more distant they are.
Similarity is measured by $\cos[\text{w},\infty,\phi_1,\mathbb{L}_{12}]$ while distance is converted into closeness and normalized to the unit interval (the values of the heatmap scale to $[-1,1]$ by calculating $-1 + \text{closeness} + \text{similarity}$).
Figure \ref{fig:Heatmap of Place Distances and Similarities} (right), shows that there is hardly a tendency to being more similar when being more close to each other. The lower similarity values are mostly induced by the rather unusually small wikis such as Boppard (see Table \ref{table:general_wiki_stats}).
Figure \ref{fig:Heatmap of Overlapping Communities} shows the Fuzzy Jaccard of the communities underlying the wikis, that is, the overlap of these communities weighted by the activities of their authors: 
the lower the number of shared authors of two wikis and the less active these authors, the lower the fuzzy overlap of these wikis.
The Fuzzy Jaccard is computed as follows \cite[cf.][]{Ramli:Mohamad:2009}:
let $\myauthors(\mathbb{W})$ be the set of all registered users contributing to any of the wikis in $\mathbb{W} = $ \Cities $\cup$ \Regions, \Others $\cup$ \WPRegioOne $\cup$ \WPOthersOne and let $\articles(W)$ be the set of all (non-redirect) articles of wiki $W\in\mathbb{W}$, then we compute
\begin{eqnarray}
\forall A,B\in \mathbb{W}\!:\; J_{\mu}(A,B) & = & \frac{\sum_{r\in \myauthors(\mathbb{W})} \mu_{A\cap B}(r)}{\sum_{r\in \myauthors(\mathbb{W})} \mu_{A\cup B}(r)} \in [0,1]
\end{eqnarray}
where 
\begin{alignat}{2}
\mu_{A\cap B}(r) &= \min\left(
\frac{\sum_{x\in \articles(A)} 
	\activity(r,x)}{\sum_{s\in \myauthors(A)}\sum_{x\in \articles(A)} \activity(s,x)}, 
\frac{\sum_{x\in \articles(B)} \activity(r,x)}{\sum_{s\in \myauthors(B)}\sum_{x\in \articles(B)} \activity(s,x)}
\right) \\
\mu_{A\cup B}(r) &= \max\left(
\frac{\sum_{x\in \articles(A)} 
	\activity(r,x)}{\sum_{s\in \myauthors(A)}\sum_{x\in \articles(A)} \activity(s,x)}, 
\frac{\sum_{x\in \articles(B)} \activity(r,x)}{\sum_{s\in \myauthors(B)}\sum_{x\in \articles(B)} \activity(s,x)}
\right) 
\end{alignat}
Figure \ref{fig:Heatmap of Place Distances and Similarities} shows that while among the Wikipedia-based extractions the overlap is remarkably high, it does nearly not exist between any of the city or region wikis:
these wikis are written by mostly completely different communities. 
The picture is not different if one considers all authors -- registered and unregistered.

\section{Discussion}\label{sec:Discussion}

Section \ref{sec:Experimentation} has shown that topic networks, whether TTNs or ATNs, are similar if they belong to the same genre, while they are characterized by a high degree of thematic focusing.
In order to operationalize this notion of network similarity, we tested, further or newly developed 11 different measures of network similarity by relying on four different paradigms of measuring the similarity of graphs (see Table \ref{tab:Graph Similarity Measures} and the discussion of graph/network similarity measures in Section \ref{sec:Module 6: Graph Similarity Analysis}) as instantiated by the complex networks studied here. All these measures and paradigms come along with a different notion of network similarity. We have shown that a subclass of them, especially cosine-based measures of network similarity, allow for detecting similarities of topic networks in line with Hypothesis 3 and 4. At the same time, the concept of network similarity underlying this class of dual weight-dependent measures seems to be the most promising from a research point of view, as it is based on node and arc weights and instantiates a very intuitive concept of network similarity: 
The more similar two networks are from the perspective of the more of their nodes, the more similar they are.
Thus, at the level of thematic abstraction examined here, there seems to be a hidden tendency to write about very prominent topics when it comes to thematizing places and linking the underlying texts in such a way that the resulting networks become almost indistinguishable.

Starting from this kind of thematic distortion of VGI as conveyed by online media, we now ask for a more general explanation of our findings.
The candidate we are considering for this purpose is given by \textit{Cognitive Maps} (CM) which were introduced as models of the cognitive representation and processing of spatial information to explain a number of different cognitive biases.
Because of bridging the gap between geographical information and its biased representation, CMs promise to be a candidate for our task.
At the same time, this notion allows for the connection of cognitive geography on the one hand and our generalized model of linguistic encoding of geographical information on the other (see Figure \ref{fig:TFL}).
The reason is that as mental representations, CMs are seen to integrate a wide range of representations of spatial objects, their relations and thematic units (see below).
We may argue now that we developed a method to represent and analyze a particular type of thematic information which can be subsumed under the latter list.
If this is true, then the thematic distortion observed by us could be seen as a result of the biased processing of geographic information by a community of agents dealing with the same place to generate a common cognitive map thereby manifesting a particular type of distributed cognition.
When creating such a common CM of the same place, agents tend to focus on a highly selected set of rhemes (see Figure \ref{fig:TFL}), even if there is no explicit agreement among these agents about this selection, and even if there is little or no direct communication between them and also irrespective of the focal place.
It seems that the agents participate in processes of distributed cognition in such a way that their own thematically distorted maps flow into the formation of a shared, stable but likewise distorted \enquote{thematic map}. 
These maps then appear as the result of a sort of swarm behavior regarding the formation of a particular distribution of the preference and salience of certain place-related rhemes.
From this perspective, topic networks serve as models of these thematic maps which in turn are parts of CMs.
To underpin this interpretation, we briefly summarize the research on CMs and, above all, ask about distortions that are distinguished by the research in this area.

Understood as mental representations of spatial knowledge, CMs have been subject of scientific work for decades.
Starting from different disciplinary perspectives, this research provides insights into how people perceive their environment, think about it and how this influences their spatial behavior.
The interdisciplinary research on CMs has led to a multitude of notions, research designs, and outcomes, the integration of which is still pending. 
Over the years, researchers worked, for example, with different terms for the mental representations in question such as \textit{cognitive maps} \cite{Tolman:1948}, \textit{environmental images} \cite{Lynch:1960}, \textit{mental maps} \cite{Gould:White:1986}, \textit{mental sketch maps} \cite{Gieseking:2013}, \textit{narrative space maps} \cite{Helferich:2014}, or \textit{internal representations} \cite{Portugali:1996}, where the constituent \textit{map} is most common.
However, there has been a discussion as to whether the term \textit{map} is generally misleading. 
In this context, Kitchin \cite[][3pp.]{Kitchin:1994} distinguishes approaches that understand CMs as
\begin{enumerate}
\item three-dimensional maps, 
\item an analogy to maps (because of their map-like characteristics), 
\item a metaphor for maps (because they function as if they were maps) or as
\item a hypothetical construct used to explain spatial behavior. 
\end{enumerate}

While we refer to cognitive maps as an auxiliary notion, we adhere to the fourth of these variants.
Regardless of this discussion, there is a greater consensus on some characteristics of CMs as mental representations: 
CMs are understood as complexes of mental images and concepts that humans have in mind when thinking about places, their \textit{location} (in terms of distance and direction), \textit{accessibility} (regarding questions like how to get there) and the \textit{meanings} associated with them.
They serve as a means of understanding spatial circumstances and as a frame of reference for the interpretation, preference and prediction of spatial structures, their relations and events in which they participate (see \cite[][100pp.,313]{Downs:Stea:1982}, \cite[][3]{Gould:White:1986} and \cite[][5p.]{Lynch:1960}).
Beyond that, they also serve as a basis for decision-making regarding spatial behavior (e.g.\ in route planning). 
In a nutshell, humans activate, generate and utilize CMs in spatial thinking and spatial behavior \cite[cf.][233]{Golledge:Stimson:1996}.
CMs are distinguished according to the entities they model. 
Kitchin and Blades \cite[5p.]{Kitchin:Blades:2002} distinguish CMs of \textit{object spaces} (e.g.\ rooms, cars), \textit{environmental spaces} (e.g.\ buildings, streets, neighborhoods, cities), \textit{geographical spaces} (e.g.\ regions, countries), \textit{panoramic spaces} and of \textit{map spaces} (including models) \cite[cf.][]{Freundschuh:Egenhofer:1997}. 
In this way, they cover existing as well as imagined places, where facts about the former can be mixed with imaginations of the latter \cite{Downs:Stea:2011}.
This list includes the kind of places that are central to our study, especially cities.

To build a bridge between the notion of CMs and our analysis, we need to look more closely at their content and the principles by which they are created.
Generally speaking, CMs are seen to cover at least two types of information (see \cite[][1p.]{Kitchin:1994} and \cite[][314p.]{Downs:Stea:2011}):
\begin{enumerate}
\item Regarding \textit{spatial cognition}, this concerns information about where entities are located in the environment of a person (location, distance and direction in relation to her location or to reference points like landmarks).

\item Regarding \textit{environmental cognition}, this concerns information about the kind of these entities, their attributes, meanings, valuations and attitudes that the person associates with them -- individually, socially or culturally mediated \cite[][224, 235]{Golledge:Stimson:1996}. 
\end{enumerate}

Our study focuses on the second part of this distinction: 
it is related to the rhemes that are associated with places as framing themes (see Section \ref{sec:Introduction}).
In any event, CMs are systematically characterized by distortions \cite[][315]{Downs:Stea:2011} concerning judgments about locations, distances and directions as well as the formation of preferences which effect spatial or environmental cognition.
One example is the \textit{localization effect} \cite{Gould:White:1986} according to which 
people can discriminate nearby places better and have stronger preferences for them \cite[see also][]{Golledge:Stimson:1996}.
This relates to errors in distance judgments depending on the perspective from which they are made: 
more differences are seen between closer areas than between more distant ones, so that shorter distances are exaggerated, while longer distances are underestimated \cite[][133]{Tversky:1992}.
Furthermore, spatial knowledge can be organized by reference to landmarks which \enquote{distort} places in their \enquote{neighborhood} so that buildings, for example, are judged to be closer to them than vice versa \cite[][134]{Tversky:1992}.
Tversky \cite[][135pp.]{Tversky:1992} describes additional modes of distortion: 
to remember the position and orientation of objects, humans isolate them from their background and organize them by referring to a general frame of reference (rotation) or to other figures (alignment).
While these examples primarily concern spatial cognition, the following bias focuses more on environmental cognition.
This concerns the hierarchical organization of conceptual systems according to which places of the same category are supposed to be closer in distance than places of different categories, while the direction of a category (with a direction slot) determines the one of its members \cite[][132p.]{Tversky:1992}.
Last but not least, Golledge and Stimson \cite{Golledge:Stimson:1996} describe distortions of the representation of urban spaces.
They observe that interactions influence the perception of a city in the sense that spatial information accumulates along the representations of the paths used to carry out these interactions. 
Likewise, structural properties of cities which are more salient than others are likely to become anchor points in CMs. 
In such maps, areas between used paths and anchor points may appear to be \enquote{folded} or \enquote{wrapped} so that preferred visited places are represented closer to each other. 
As a result, positional and relational errors can occur in perception (see \cite[][254]{Golledge:Stimson:1996} and \cite[][7]{Golledge:Gaerling:2001}).

To interpret our findings in the light of this research, we need to link the formation of CMs with linguistic processes.
The idea that this formation is substantially influenced by human language processing, so that geographical information is non-trivially encoded in linguistic structure, goes back to the work of Louwerse \cite[cf.][]{Louwerse:Benesh:2012} (see Section \ref{sec:Introduction}; see also Montello \& Freundschuh \cite[][171]{Montello:Freundschuh:1995} for an earlier hint on \enquote{\textit{obtain[ing] spatial knowledge through language}}).
In this context, Golledge \& Stimson \cite[][235]{Golledge:Stimson:1996} distinguish shared components of CMs from personalized ones by stating that
\enquote{\textit{The common elements facilitate communication with others about the characteristics of an environment; the idiosyncratic elements provide the basis of the personalized responses to such situations}}.
Our hypothesis is now that at the level of thematic abstraction as modeled here, the organization of platial rhemes shared by the members of a community is influenced by the general law of preferential order which is most prominently instantiated by Zipf's first law \cite{Zipf:1972}.
Such an organization makes the anticipation of a place rather expectable among the members of a community so that communication about this place is facilitated as predicted by Golledge \& Stimson \cite{Golledge:Stimson:1996}.

This Zipfian organization allows for relating our findings to the well known power-law-like degree distributions found in many natural, social, semiotic or technical networks (see \cite{Newman:2003:a,Newman:2010:a} and especially \cite{Newman:2005:a} for overviews of this and related research) and also by example of many linguistic systems -- especially on the text level \cite{Rapoport:1982,Tuldava:1995,Naranan:Balasubrahmanyan:1998}.
Because of this commonality, one might assume that we just detected a well-known \textit{text} or \textit{network} characteristic. 
Characteristic for our findings, however, is that we developed a measurement procedure that detects a \textit{text (corpus)-related semantic, thematic trend} -- with the help of network theory: 
Instead of counting directly observable arcs, for example, in ontological networks or co-occurrences in texts and instead of relying on monoplex networks \cite{Abramov:Mehler:2011:a,Amancio:et:al:2012,Cattuto:Barrat:Baldassarri:Schehr:Loreto:2009,Ferrer-i-Cancho:Sole:2001,Ferrer-i-Cancho:Mehler:Pustylnikov:Diaz-Guilera:Kurzform:2007:a,Mehler:2009:c:Langfassung,Baronchelli:et:al:2013}, we generated and analyzed a range of different networks in relation to each other in order to determine the corresponding thematic trend by means of multiplex networks. 
This is not to say that we first discovered a Zipfian process in the organization of linguistic networks, but rather that we observe such a process in a very specific area, in which it has not been observed before and which requires an appropriate explanation as elaborated so far.
Indeed, if thematic salience is skewed, and if skewed topic distributions derived from different corpora are similar not only topologically but also regarding the ranking of the majority of salient topics, such an observation requires explanation subject to the fact that the underlying text networks are constituted by different, distributed communities of authors.
It is the answer to this question that the paper was about.

At this point one might further object that we made a rather expectable observation in the sense that descriptions of cities, for example, are very likely related to rhemes like traffic, trade, culture, history etc.
However, this would mean underestimating our results:
(i) the thematic distortions observed by us are extremely skewed, (ii) they seem to emerge rather earlier in the development of a wiki\footnote{This is not shown here, but is the result of a pretest in which we looked at the life cycles of three different wikis. In future work we will analyze the underlying time series of multiplex topic networks in detail.} and (iii) they make both members of the same genre similar while allowing for distinguishing members of different genres.
To phrase it as a question: 
\textit{If the number of rhemes under which places are thematized is limited, why then should always a tiny subset of them dominate the discourse about a place and why then should the networking of these rhemes make discourses of the same genre identifiable?}
From this point of view, we argue that we discovered an additional form of the distortion of CMs, which means that the underlying place is always conceptualized from the point of view of a few but extremely preferred rhemes. 
When organizing their distributed processes of co-authorship, communities of authors seem to strive to a kind of thematic unification that makes different wikis serving alike functions looking structurally similar -- with respect to the preference order of themes and their networking.
It seems that people participate in processes of collaborative writing with a tendency to organize their thematic contributions and references in such a way that they remain shareable \cite{Freyd:1983} and communicable among members of the same community.
Ensuring shareability means securing the continued existence of the underlying wiki, which could otherwise collapse because of too many personalized or individualized fragmentations.
At this point we can speculate that people unconsciously prefer such thematic contributions that make their social roles and participations expectable and acceptable, whereby this selection behavior produces the described similarity of thematic maps as components of CMs.
In other words, the participants anticipate social roles and neglect their personal view of cities and regions, whose documentation would fragment the corresponding media thematically. 
Instead, they ignore the reproduction of their idiosyncratic, personalized views of places.
To say it in terms of the distinction made by Golledge \& Stimson \cite{Golledge:Stimson:1996} between shared and personalized components of CMs: 
participants overweight the former to the disadvantage of the latter to guarantee the shareability \cite{Freyd:1983,Freyd:2005} of CMs as a result of distributed cognition.

Note that in our study we did not simply map a frequency effect by our measurements: 
although we counted frequencies of topic assignments, they were determined by means of an inference process that went through a process of (machine) learning.
To support such an interpretation, however, a deeper analysis with a larger corpus of wikis and related media providing different functions is required.
This also requires experiments with other and above all much finer classification systems than the DDC to find out how much the use of the DDC has influenced our measurements.
And it requires a deeper analysis of the social roles of authors in online media, their interactions and the regulatory systems under which they interact.
But this already concerns future work.

\section{Conclusion}\label{sec:Conclusion}

We developed a novel model of topic networks in order to investigate the networking of rhemes addressing the same places in underlying corpora of natural language texts.
We developed our network model in a way that it enables thematic comparisons of previously unforeseen text corpora using an underlying reference corpus, offers a generic solution to the problem of topic labeling, is highly scalable and can therefore map even the smallest text snippets to topic distributions, simultaneously takes rare topics into account and is methodologically open and expandable.
Moreover, our model allows for comparatively investigating the networking of thematic units from different angles.
In this way, it is open and expandable as it allows for integrating different analytical perspectives into the study of the same semantic networks.
We exemplified our model by means of corpora of special wikis and extracts from Wikipedia in order to investigate how textual information encodes geographical information on the aboutness level of texts.
Our experiments show that the thematizations of different places on a certain level of abstraction are similar to each other in that they focus on a few themes in a highly distorted manner while networking them in similar ways. 
This happens regardless of whether the underlying media are generated by different communities and whether these communities address related or unrelated places in nearby or distant places.
We interpreted our findings in the context of the notion of cognitive maps.
To this end, we proposed to extend this notion in terms of thematic maps and argued that participants or interlocutors of online communication tend to organize their contributions in a way that makes them sharable.
This means that the contributions are abstracted and depersonalized at the aboutness level in such a way that the social roles of these participants become expectable and acceptable, while their personal views of places are reduced whose documentation would fragment the corresponding media thematically.
Ensuring shareability means securing the continued existence of the wiki, which could otherwise collapse in the face of too many personalized or individualized fragmentations.
Future work concerns several tasks:
We want to conduct deeper analyses based on larger corpora that manifest a greater variety of communication functions in order to shed more light on the genre sensitivity discovered in our study.
Beyond the DDC, we strive for the use of finer structured, higher resolution classification systems in order to model the contents of texts much more precisely.
Ideally this should be carried out with the help of systems like the category system of Wikipedia or even Wikidata, both of which develop as open topic universes \cite{Mehler:Waltinger:2009:b}.
Last but not least, a deeper analysis of the social roles of authors in online media and their co-authorship is required to gain a deeper understanding of the processes of linguistic encoding of geographical information.
This will be the task of future work.

\section*{Acknowledgment}

Financial support by the Federal Ministry of Education and Research (BMBF) via the \textit{Centre for the Digital Foundation of Research in the Humanities, Social, and Educational Sciences} CEDIFOR) is gratefully acknowledged.

\bibliographystyle{plain}
\bibliography{MultiplexTopicNetworks1.bib} 

\section*{Appendix}

\subsection*{text2ddc}

text2ddc is trained by means of corpora that are derived by integrating information from Wikidata, Wikipedia and the \textit{Integrated Authority File} (\textit{Gemeinsame Normdatei} -- GND) of the German National Library:
we explore the links of Wikipedia articles to entries in Wikidata containing the property attribute \url{https://www.wikidata.org/wiki/Property:P1036} that directly links to the DDC or to a GND page containing a DDC tag. 
An example is the article about the \textit{Pythagorean theorem} (\url{https://en.wikipedia.org/wiki/Pythagorean_theorem}) which is linked to the GND page 4176546-1 (\url{https://d-nb.info/gnd/4176546-1}) referring to the DDC tag 516 (\textit{geometry}). 
Using such information, we obtain a corpus for a subset of 98 classes of the \nth{2} and for a subset of 641 classes of the 3rd DDC level. 
Since Wikipedia exists for many languages, such corpora can be created for each of them.
For preprocessing the input data of text2ddc, we use \texttt{TextImager} \cite{Hemati:Uslu:Mehler:2016} and \texttt{fastSense} \cite{Uslu:et:al:2018:a} for disambiguating this data on the sense level. 
The resulting information is used to train a neural network for classifying any piece of text (down to the word level) into DDC classes (see \url{https://textimager.hucompute.org/DDC/}).
To this end, text2ddc uses a very efficient classifier, that is, \texttt{fastText} \cite{Joulin:Grave:Bojanowski:Mikolov:2016},
a bag-of-words model to train a neural network with a single hidden layer.
To optimize \texttt{fastText}, we optimize the following hyperparameters:
learning rate: \numprint{0.2}; update rate: \numprint{150}; minimal number of word occurrences: \numprint{5}; number of epochs: \numprint{10000}.
In this way, we increase the $F$-score to \numprint{87.4}\% for the \nth{2} and to \numprint{78.1}\% for the 3rd level of the DDC.

\subsection*{Color codes and \nth{2} class members of the DDC}

\begin{figure}[t]
	\includegraphics[width=1\textwidth]{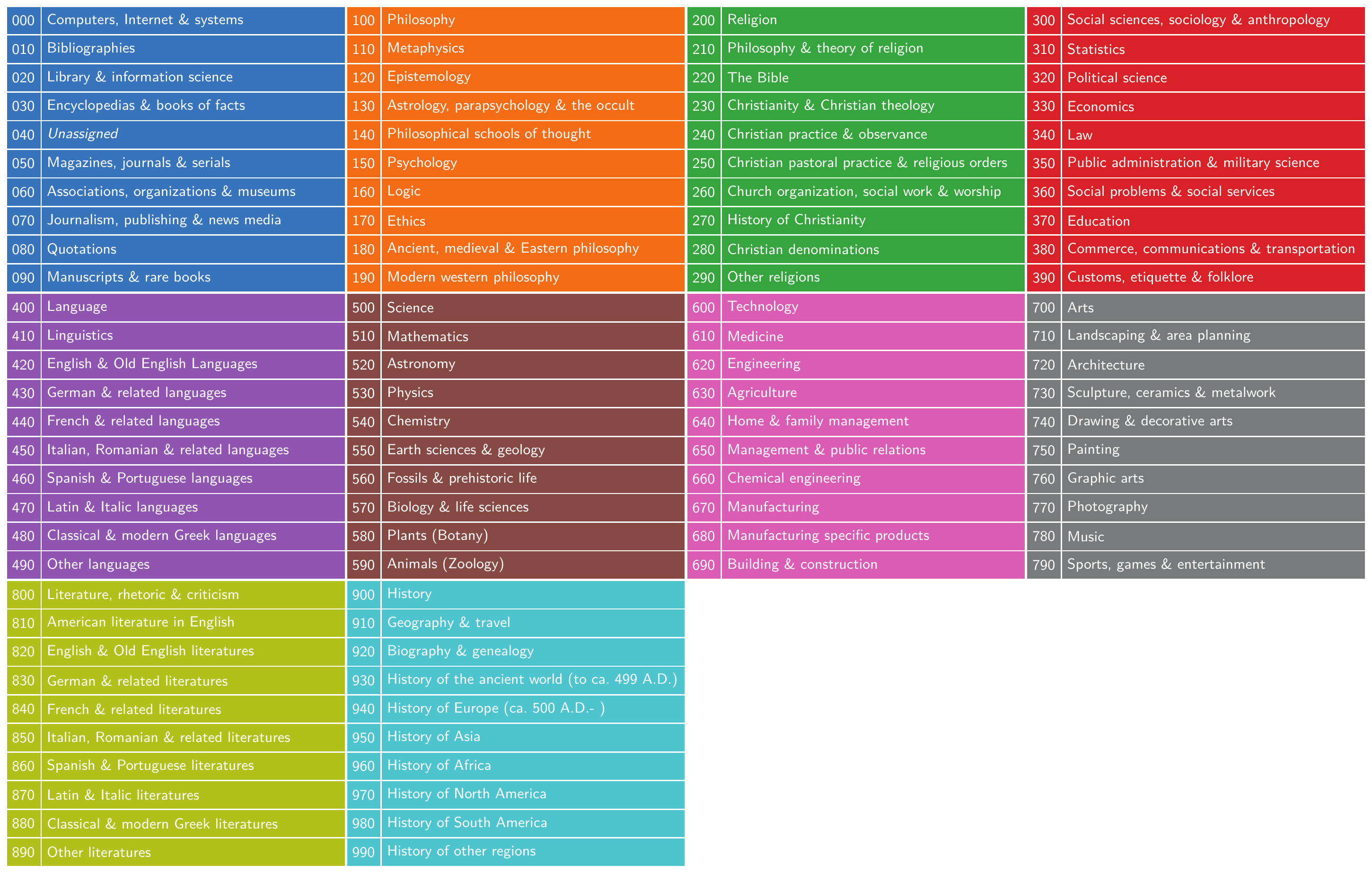}
	\caption{Color codes of the classes of the \nth{2} level of the DDC.}\label{fig:DDC - First Level}
\end{figure}

Table \ref{fig:DDC - First Level} shows the colors and labels of the classes of the \nth{2} level of the DDC.

\end{document}